\documentclass[11pt]{article}
\usepackage[preprint]{acl}

\usepackage{times}
\usepackage{latexsym}
\usepackage[T1]{fontenc}
\usepackage[utf8]{inputenc}
\usepackage{microtype}
\usepackage{inconsolata}
\usepackage{graphicx}
\usepackage{microtype}
\usepackage{natbib}
\usepackage{amsmath}
\usepackage{booktabs}
\usepackage{multirow}
\usepackage{tabularx}
\usepackage{adjustbox}
\usepackage{amssymb}
\usepackage[table]{xcolor}
\usepackage{colortbl}
\usepackage{tcolorbox}
\usepackage{placeins}
\usepackage{balance}
\tcbuselibrary{listings, breakable}

\title{PremiseSQL: Evaluating Premise Critique and Abstention in Text-to-SQL Systems}

\author{
  \textbf{Shuo Lin}\textsuperscript{1} \quad
  \textbf{Jinzhe Li}\textsuperscript{1,3} \quad
  \textbf{Chenlu Guo}\textsuperscript{4} \quad
  \textbf{Yanhan Li}\textsuperscript{5} \quad
  \textbf{Yuan Wu}\textsuperscript{1}\thanks{Corresponding authors.} \quad
  \textbf{Yi Chang}\textsuperscript{1,2,3}\footnotemark[1] \\
  \mdseries
  \textsuperscript{1}School of Artificial Intelligence, Jilin University \\
  \textsuperscript{2}Engineering Research Center of Knowledge-Driven
  Human-Machine Intelligence, MOE, China \\
  \textsuperscript{3}International Center of Future Science, Jilin University \\
  \textsuperscript{4}Bilibili Inc. \\
  \textsuperscript{5}Xingchen AGI Lab, China Telecom Artificial
  Intelligence Technology (Beijing) Co., Ltd \\
  \texttt{\{linshuo26,jinzhe25\}@mails.jlu.edu.cn} \\
  \texttt{guochenlu01@bilibili.com, liyh181@chinatelecom.cn} \\
  \texttt{\{yuanwu,yichang\}@jlu.edu.cn}
}

\begin{document}
\maketitle
\begin{abstract}
Recent advances in large language models (LLMs) have substantially improved Text-to-SQL performance, yet existing NL2SQL systems largely assume that user queries are valid and answerable. As a result, models often generate executable SQL even for queries with flawed premises. We introduce \textit{premise critique ability} in NL2SQL, namely the ability to recognize flawed queries and abstain from generating misleading SQL. To evaluate this capability, we propose \textbf{PremiseSQL}, a benchmark containing 1,800 flawed NL2SQL queries across four categories of premise failures, along with their 1,800 original queries. We further design an evaluation framework to measure intrinsic and prompt-induced premise-aware behavior across three prompting settings. Experiments on frontier LLMs show that current models are strongly premise-trusting by default and rarely critique flawed premises. Stronger critique-oriented prompting improves detection and abstention on flawed queries, but also increases false invalid judgments and suppresses SQL generation on originals. The code is available at
\url{https://github.com/vegeet/PremiseSQL}.
\end{abstract}

\section{Introduction}

\begin{figure}[!t]
\centering
\includegraphics[width=0.85\columnwidth]{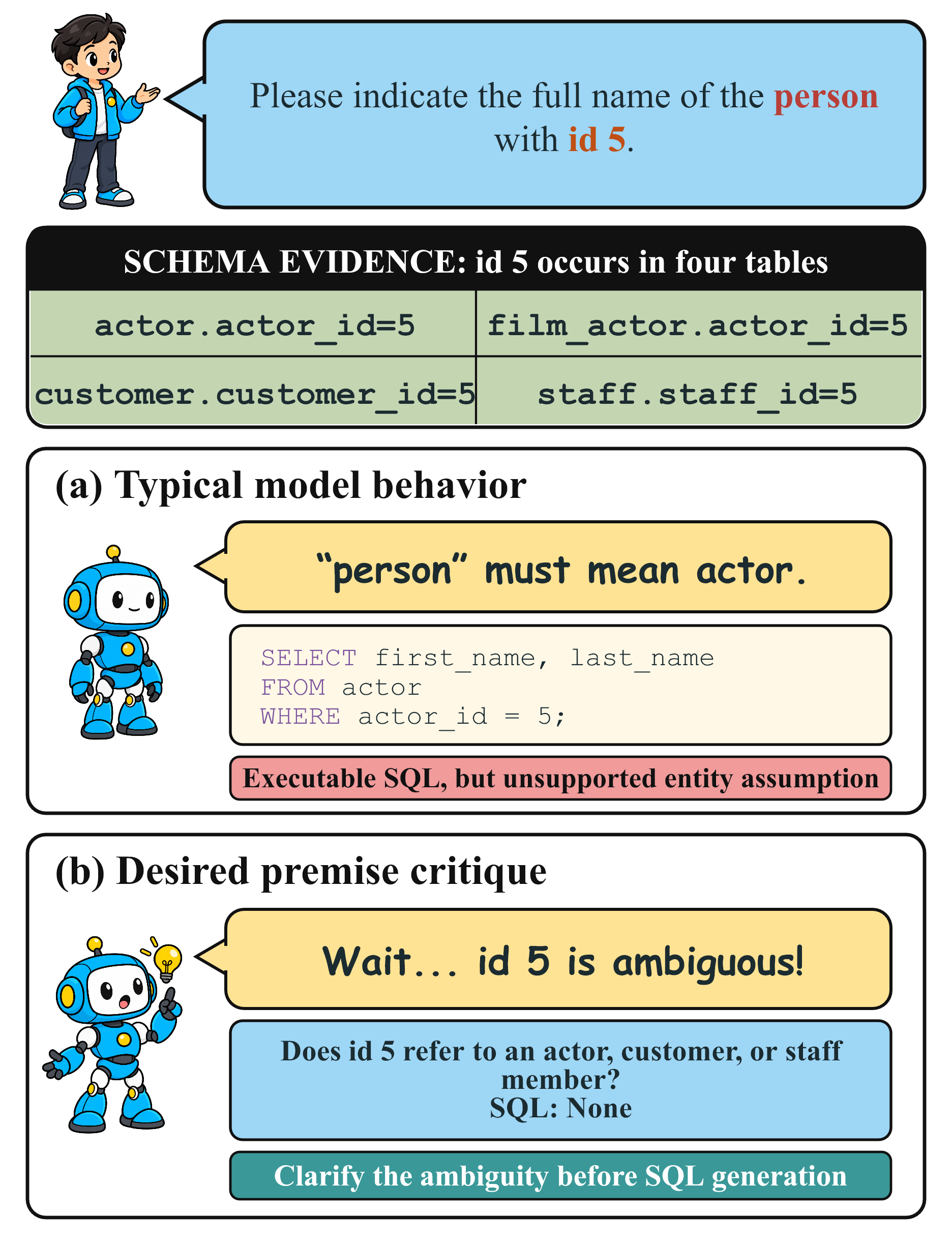}
\caption{\textbf{Illustration of how LLMs handle an NL2SQL query with an ambiguous premise.} The example contrasts two model behaviors: one generates executable SQL from an unsupported interpretation directly; the other identifies the ambiguity, requests clarification, and abstains from SQL generation. This contrast demonstrates the necessity of \textit{premise critique ability}, namely the ability to recognize flawed queries and abstain from generating misleading SQL.}
\label{fig:premise_behavior_story}
\end{figure}

Text-to-SQL (or NL2SQL) translates natural language (NL) questions into executable SQL given a database schema \cite{10.1145/604045.604070}. Recent advances in large language models (LLMs) have substantially improved schema understanding, compositional reasoning, and cross-domain generalization in modern NL2SQL systems \cite{openai2024gpt4technicalreport,touvron2023llamaopenefficientfoundation,qwen3,deepseekai2025deepseekv32pushingfrontieropen,10.1145/3641289, scholak2021picardparsingincrementallyconstrained,yu-etal-2018-spider,pourreza2023din}.

Despite this progress, existing Text-to-SQL research largely assumes that user questions are valid, well-formed, and answerable under the given schema \cite{yu-etal-2018-spider,li2024can}. Consequently, most prior work focuses on improving SQL generation accuracy, schema linking, and complex reasoning capabilities \cite{wang-etal-2020-rat, scholak2021picardparsingincrementallyconstrained,pourreza2023din}. However, this assumption is frequently violated in realistic scenarios. User questions may contain logically inconsistent conditions, false schema assumptions, missing premises, or requests beyond the capability of SQL and the underlying database.

Nevertheless, state-of-the-art NL2SQL models often generate executable SQL even for flawed queries, producing SQL that is syntactically valid but semantically invalid \cite{10.1145/3571730, 10.1007/978-3-031-20059-5_9, ren2023outofdistributiondetectionselectivegeneration,10.1145/3703155}. In many cases, models implicitly repair problematic questions by injecting unsupported assumptions \cite{lin-etal-2022-truthfulqa}. Figure~\ref{fig:premise_behavior_story} illustrates this failure: although "person with id 5" is underdetermined by the schema, a model may silently map \textit{person} to \textit{actor} and emit executable SQL. The resulting SQL is syntactically valid but rests on an unsupported entity choice.

We introduce \textit{premise critique ability} in NL2SQL, namely the ability to recognize flawed queries and abstain from generating misleading SQL. Premise critique ability is orthogonal to conventional SQL generation accuracy: a model may produce executable SQL, or even generate SQL that coincidentally matches the user's intent, while still failing to recognize that the provided question itself is ill-posed.
Concretely, it consists of two components: (1) \textit{premise validation}, namely assessing query validity in the corresponding database context and identifying actual flaws; and (2) \textit{abstention}, namely refraining from generating misleading SQL for flawed queries while maintaining SQL generation for valid queries.

To systematically evaluate this capability, we introduce \textbf{PremiseSQL}, to our knowledge the first benchmark specifically designed to evaluate premise critique ability in NL2SQL systems. Built upon two widely used Text-to-SQL datasets, Spider 1.0 \cite{yu-etal-2018-spider} and BIRD / BIRD-filtered \cite{li2024can}, PremiseSQL pairs 1,800 constructed flawed queries with their 1,800 unmodified source queries. It evaluates whether models can recognize flawed premises and abstain from generating misleading SQL, while maintaining SQL generation for original queries. We further evaluate frontier LLMs under three prompting settings, including \textit{Base}, \textit{Implicit}, and \textit{Explicit}, to measure both intrinsic and prompt-induced premise-aware behavior.

In summary, our contributions are as follows:

\begin{itemize}
    \item We formalize \textit{premise critique ability} as a distinct requirement for reliable NL2SQL.

    \item We introduce \textbf{PremiseSQL}, a taxonomy-guided benchmark comprising flawed queries as positive samples and their originals as negative samples.

    \item We develop an evaluation framework that separates intrinsic critique from prompt-induced behavior. Experiments on 16 frontier LLMs reveal a shift from trusting flawed premises under standard prompting to frequently rejecting valid queries under explicit critique instructions.
\end{itemize}
\section{Related Work}
\subsection{Unanswerable and Ambiguous Queries in NL2SQL}

Prior work has explored unanswerable and ambiguous queries in NL2SQL. TriageSQL \cite{zhang2020did} formulates the problem as question intention classification, while Know What I Don’t Know \cite{wang-etal-2023-know} further categorizes ambiguity and unanswerability cases. More recent work extends beyond classification-oriented formulations. TrustSQL \cite{lee2024trustsql} introduces infeasible queries and proposes a reliability-oriented evaluation metric for trustworthy NL2SQL systems. Complementarily, NL2SQL-BUGs \cite{10.1145/3711896.3737427} focuses on errors in NL2SQL translation and categorizes semantic errors into fine-grained types.
However, NL2SQL still lacks a fine-grained and comprehensive examination of diverse premise errors and the varying challenges they pose to models. PremiseSQL addresses this gap through a unified premise-error taxonomy.

\subsection{Robustness and False Premise Handling in LLMs}

\begin{table*}[t]
\centering
\small
\fontsize{9pt}{9.5pt}\selectfont
\renewcommand{\arraystretch}{0.85}
\begin{tabular}{p{0.6cm} p{2.8cm} p{3.8cm} p{7cm}}
\toprule

Code & Category & Subcategory & Description \\

\midrule

\multirow{2}{*}{E1}
& \multirow{2}{*}{Expression Error}
& E1.1 Ambiguity
& Query contains ambiguous references, scopes, or targets that admit multiple plausible SQL interpretations. \\

&
& E1.2 Unclear Intent
& Query intent or operation type is underspecified, making the SQL structure indeterminate. \\

\midrule

\multirow{2}{*}{E2}
& \multirow{2}{*}{Premise Contradiction}
& E2.1 Internal Text Contradiction
& Multiple query conditions cannot logically hold simultaneously. \\

&
& E2.2 Question--Schema Contradiction
& Required schema elements or relations are absent or incompatible with the database schema. \\

\midrule

\multirow{3}{*}{E3}
& \multirow{3}{*}{Missing Premise}
& E3.1 Missing Constraint
& Essential filtering conditions or target specifications are omitted. \\

&
& E3.2 Missing Schema Premise
& Required tables, columns, or relations are not provided in the schema. \\

&
& E3.3 Semantic Gap
& Query concepts cannot be semantically mapped to schema elements. \\

\midrule

\multirow{2}{*}{E4}
& \multirow{2}{*}{Beyond Ability}
& E4.1 Inexpressible in SQL
& Query intent cannot be precisely represented using standard SQL. \\

&
& E4.2 External Knowledge Required
& Correct answering requires knowledge absent from the database. \\

\bottomrule
\end{tabular}
\caption{Taxonomy of flawed premises in Text-to-SQL.}
\label{tab:error_taxonomy}
\end{table*}

\begin{table}[htbp]
\centering
\small
\begin{tabular}{lccccc}
\hline
Method 
& FP
& Abst.
& SQL
& Fine
& Prompt \\
\hline

TriageSQL
& $\checkmark$
& $\times$
& $\checkmark$
& $\triangle$
& $\times$ \\

KWIDK
& $\triangle$
& $\checkmark$
& $\checkmark$
& $\triangle$
& $\times$ \\

TrustSQL
& $\checkmark$
& $\checkmark$
& $\checkmark$
& $\times$
& $\times$ \\

NL2SQL-BUGs
& $\times$
& $\times$
& $\checkmark$
& $\checkmark$
& $\times$ \\

SQuADRU{\small N}
& $\checkmark$
& $\checkmark$
& $\times$
& $\times$
& $\times$ \\

AbstentionBench
& $\checkmark$
& $\checkmark$
& $\times$
& $\triangle$
& $\triangle$ \\

\textbf{PremiseSQL}
& $\checkmark$
& $\checkmark$
& $\checkmark$
& $\checkmark$
& $\checkmark$ \\

\hline
\end{tabular}

\caption{Comparison with related work on flawed-premise and abstention evaluation. FP denotes explicit flawed-premise reasoning before SQL generation, Abst. denotes abstention evaluation, SQL indicates whether the benchmark is formulated around SQL generation tasks, Fine denotes fine-grained error taxonomy, and Prompt denotes prompt-aware evaluation settings. Symbols: $\checkmark$ Fully supported, $\triangle$ Partially supported, $\times$ Not supported.}
\label{tab:related_comparison}
\end{table}

Beyond NL2SQL, prior work studies LLM robustness under adversarial, noisy, or logically flawed inputs. 
For example, SQuADRU{\small N} \cite{rajpurkar-etal-2018-know} introduces unanswerable questions, highlighting models’ tendency to produce confident but incorrect responses when no valid answer exists. AbstentionBench \cite{kirichenko2025abstentionbench} provides a large-scale benchmark across 20 diverse datasets to assess abstention behavior.
Building on these, we cover more types of flawed queries in NL2SQL and examine premise critique under different levels of premise-aware prompting, revealing both the limitations of this capability in LLMs and how it can be elicited. The paired originals further allow us to distinguish evidence-grounded critique from instruction-driven rejection.

\paragraph{Summary}
Table~\ref{tab:related_comparison} compares PremiseSQL with representative benchmarks. PremiseSQL jointly supports fine-grained premise-error analysis, abstention evaluation, SQL generation, paired original-query controls, and prompt-aware evaluation, enabling a more complete assessment of premise critique in NL2SQL.

\section{Method}

\begin{figure*}[t]
\centering
\includegraphics[width=\textwidth]{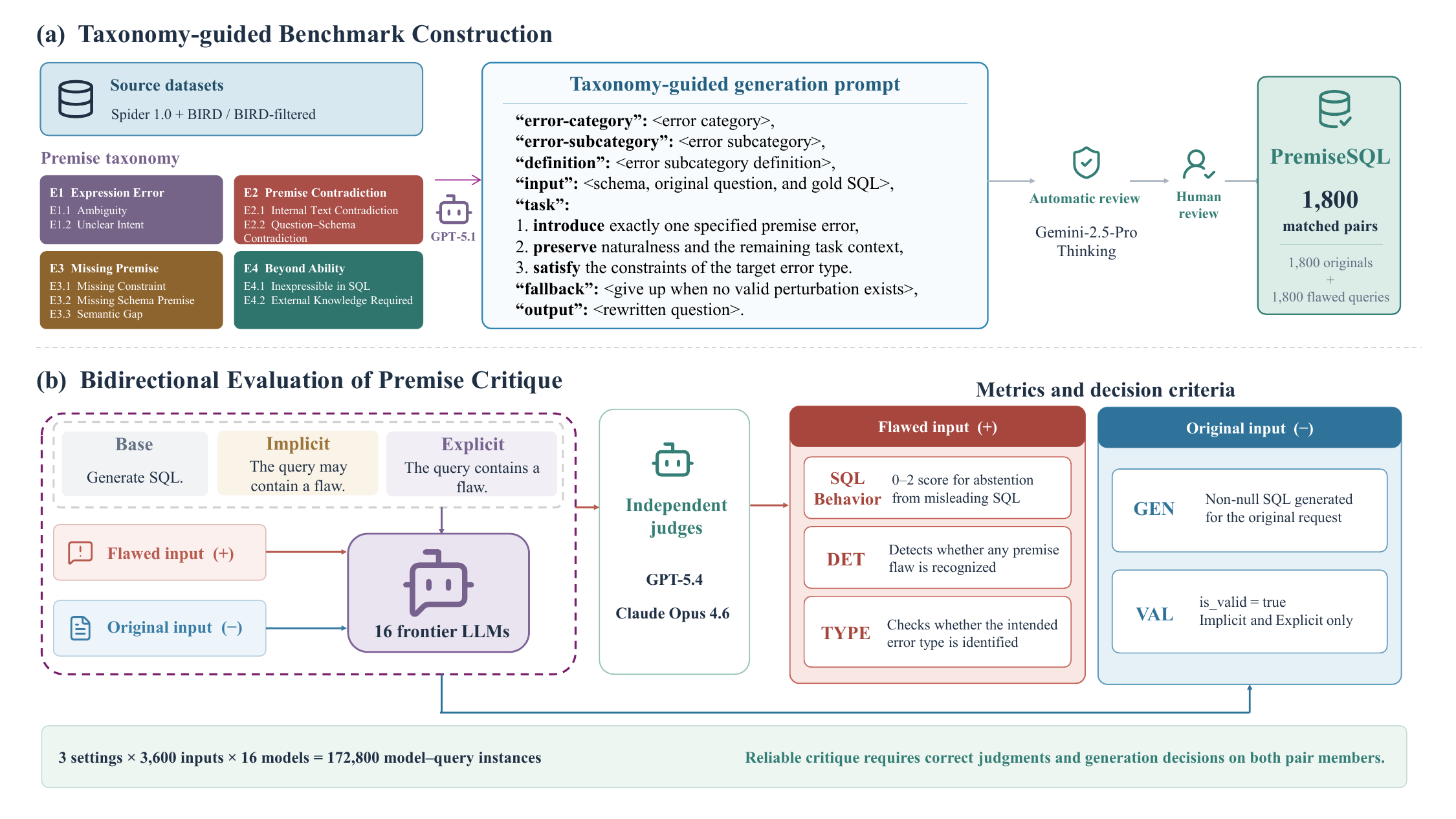}
\caption{
\textbf{Overview of PremiseSQL.}
Starting from standard Text-to-SQL queries, we construct flawed premise variants under a unified taxonomy, followed by multi-stage verification and premise-aware evaluation under Base, Implicit, and Explicit settings.
}
\label{fig:overview}
\end{figure*}

\subsection{Task Formulation}

Let $x$ denote a natural language query and $c$ the available database context, comprising the schema and any supplied auxiliary information. We associate each query--context combination $(x,c)$ with a reference label $y\in\{0,1\}$, where $y=1$ indicates that the query has a premise flaw that warrants critique and abstention under the assumed SQL execution conditions, and $y=0$ indicates the absence of such a flaw. These flaws may concern interpretation, logical consistency, required information, or the capabilities of SQL and the database, as specified in Table~\ref{tab:error_taxonomy}. PremiseSQL assigns $y=1$ to constructed flawed queries and $y=0$ to their originals, forming the positive and negative samples, respectively.

Given a prompting setting $\pi$, the complete request comprises the query $x$, the database context $c$, and the instructions specified by $\pi$. We represent the SQL and explanation components of a model's response as
\begin{equation}
\begin{aligned}
(\hat q,\hat r) &= f_{\theta}(x,c;\pi), \\
\hat q &\in \mathcal{Q}_{\mathrm{SQL}}\cup\{\bot\},
\end{aligned}
\label{eq:premise_response}
\end{equation}
where $\mathcal{Q}_{\mathrm{SQL}}$ denotes the space of candidate SQL outputs, $\bot$ denotes abstention, and $\hat r$ is a natural language explanation.

Premise critique requires both an evidence-grounded explanation and an appropriate generation decision. The desired abstention behavior is characterized by
\begin{equation}
\hat q=\bot \quad\Longleftrightarrow\quad y=1.
\label{eq:premise_abstention}
\end{equation}
For $y=1$, the explanation $\hat r$ must identify an actual premise flaw supported by $(x,c)$; abstention alone is insufficient to demonstrate premise critique. For $y=0$, the model should generate SQL without attributing unsupported flaws to the original query.
The reference label $y$ is determined by whether the query $x$ is flawed in its database context $c$ under the benchmark's assumptions and is independent of $\pi$. Consequently, instructions that encourage scrutiny or assert the presence of a flaw should not alter the expected response to a given query.

\subsection{Premise Taxonomy}

We further organize flawed premises into a unified taxonomy.
\textbf{(1) Expression Error} refers to errors caused by ambiguous or unclear natural language expressions that prevent the query intent from being uniquely mapped to SQL;
\textbf{(2) Premise Contradiction} captures queries whose assumptions are logically inconsistent or incompatible with the database schema;
\textbf{(3) Missing Premise} refers to queries that omit essential assumptions, constraints, or schema information required for reliable SQL generation and execution;
\textbf{(4) Beyond Ability} includes queries whose requirements exceed the expressive or reasoning capabilities of SQL and the given database, making them unanswerable under the available execution assumptions.
A complete taxonomy with fine-grained subcategories and definitions is shown in Table~\ref{tab:error_taxonomy}.

\subsection{Benchmark Construction}

Figure~\ref{fig:overview} shows that PremiseSQL pairs 1,800 flawed queries with their unmodified originals from Spider 1.0 \cite{yu-etal-2018-spider} and BIRD / BIRD-filtered \cite{li2024can}. Each pair combines an original query with a taxonomy-guided perturbation of the query or supplied schema. The BIRD portion of E3.2 (Missing Schema Premise) uses BIRD rather than BIRD-filtered because it requires access to gold schema annotations.

To improve diversity and linguistic naturalness, perturbations are generated using \textbf{GPT-5.1} under predefined error constraints (see Table~\ref{tab:error_taxonomy} and Appendix~\ref{appendix:prompt_templates}). All instances are assigned corresponding error categories and subcategories, and further verified using a two-stage process consisting of automatic evaluation by \textbf{Gemini-2.5-Pro-Thinking} and subsequent human review.
Detailed statistics are provided in Appendix~\ref{app:filter_stat}.

\subsection{Prompting Protocol}

To analyze how prompting affects premise critique behavior, we evaluate models under three prompting settings: \textit{Base}, \textit{Implicit}, and \textit{Explicit}.

\paragraph{Base Setting}
The Base setting follows the standard Text-to-SQL generation setup. Models are simply instructed to generate SQL for the given question and return a structured JSON output containing \textit{sql} and \textit{reason}, where \textit{sql} denotes the generated SQL (or null when abstaining) and \textit{reason} records the model’s explanation for its generation behavior. No additional warning is provided regarding potential flaws in the query.

\paragraph{Implicit Setting}
The Implicit setting introduces implicit premise-awareness guidance. Models are informed that the input question may contain errors and are asked to assess its validity before answering. The output JSON includes \textit{sql}, \textit{is\_valid}, and \textit{reason}, where \textit{is\_valid} indicates whether the model considers the question logically valid and answerable. This setting measures whether soft prompting can improve premise validation and abstention behavior.

\paragraph{Explicit Setting}
The Explicit setting provides direct critique-oriented instructions. Models are explicitly told that the question contains logically flawed or incorrect premises, and are required to critically examine its assumptions before responding. The output format remains \textit{sql}, \textit{is\_valid}, and \textit{reason}. This setting examines premise critique behavior under an explicit assertion of a query flaw, which is supported for positive samples but false for negative samples.

\subsection{Evaluation Framework}

For positive samples, three metrics are considered: \textit{SQL Behavior} (SQL), \textit{Detection} (DET), and \textit{Type Identification} (TYPE). \textbf{SQL Behavior} measures the degree of abstention from generating misleading SQL for a flawed query, using a three-level ordinal score from 0 to 2. \textbf{Detection} evaluates whether the model recognizes that the query contains a premise flaw, and \textbf{Type Identification} measures whether it correctly identifies the corresponding failure type.

For negative samples, two complementary metrics, namely \textit{SQL Generation} (GEN) and \textit{Validity Judgment} (VAL), are introduced. \textbf{SQL Generation} measures the proportion of negative samples for which the model generates SQL, reflecting whether it maintains generation for valid queries. GEN measures generation coverage; Appendix~\ref{sec:gen_execution_validation} provides an execution-based validation showing that GEN closely tracks end-to-end execution success. \textbf{Validity Judgment} measures the proportion of negative samples explicitly judged valid through \textit{is\_valid=true}, indicating whether the model correctly recognizes query validity.

\section{Experiments}

\subsection{Experimental Setup}

We evaluate premise critique ability on a diverse set of state-of-the-art LLMs. The evaluated models include \textit{ChatGLM-5.1} \cite{glm5team2026glm5vibecodingagentic}, \textit{Claude Opus 4.6} \cite{anthropic2026claudeopus46}, \textit{Claude Sonnet 4.6} \cite{anthropic2026claudesonnet46}, \textit{Claude Opus 5.5} \cite{anthropic2026claudeopus55}, \textit{DeepSeek V3.2} \cite{deepseekai2025deepseekv32pushingfrontieropen}, \textit{DeepSeek V4.1 Flash} \cite{deepseekai2026deepseekv41flash}, \textit{Doubao Seed 2.0 Pro} \cite{bytedance2026seed2techreport}, \textit{Gemini 3.1 Pro Preview} \cite{google2026gemini31propreview}, \textit{LLaMA 4 Maverick} \cite{meta2026llama4herd}, \textit{GPT-5.4} \cite{openai2026gpt54}, \textit{GPT-5.6-Luna} \cite{openai2026gpt56}, \textit{GPT-6-Luna} \cite{openai2026gpt6solluna}, \textit{GPT-6.1-Sol} \cite{openai2026gpt61sol}, and the 35B, 122B, and 397B variants of \textit{Qwen3.5} \cite{qwen3.5}.

Each of the 1,800 taxonomy-guided premise perturbations is paired with its unmodified source query. We treat the perturbed query as a \textbf{positive sample}, for which a model should detect the flaw and abstain, and the original query as a \textbf{negative sample}, for which it should recognize the query as valid and generate SQL. Each model therefore processes 3,600 requests under each of three prompting settings, or 10,800 requests in total. Across 16 models, our evaluation comprises 172,800 model responses. We use temperature 0 and the default API configuration for each model.

\subsection{Main Results}

\begin{figure}[!ht]
\centering
\includegraphics[width=\columnwidth]{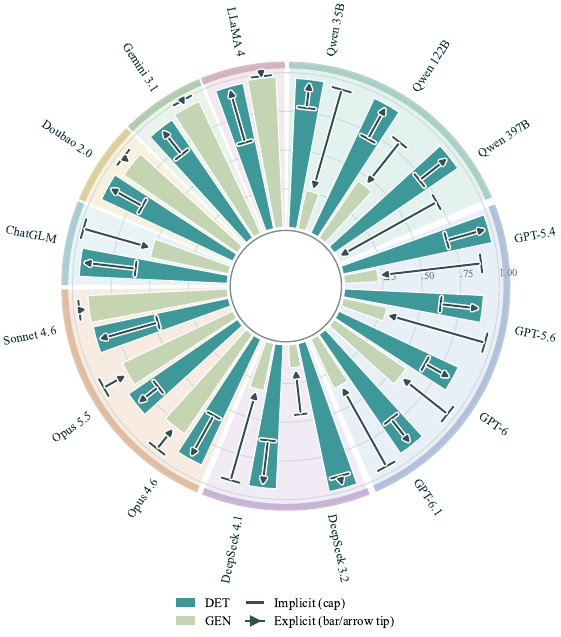}
\caption{\textbf{DET--GEN shifts from Implicit to Explicit prompting across 16 models.} Teal and sage bars denote flaw detection (DET) on flawed queries and SQL generation (GEN) on paired originals, respectively. Black caps mark Implicit scores, while bar endpoints and arrowheads mark Explicit scores; outward/inward arrows indicate increases/decreases. Explicit prompting increases DET but decreases GEN for every model, showing that stronger flaw cues improve detection at the cost of more false rejection of valid queries.}
\label{fig:det_gen_prompt_shift}
\end{figure}

Table~\ref{tab:overall_results} presents model performance on flawed queries and their paired originals. Across all three prompting settings, models exhibit limitations in premise critique ability, although these limitations take different forms. Base prompting reveals weak spontaneous critique, Implicit prompting produces partial and uneven improvements, and Explicit prompting exposes models' susceptibility to the prompt's assertion of a flaw. Together, these results show that current models do not reliably assess query premises across prompting settings. Figure~\ref{fig:det_gen_prompt_shift} further illustrates how model behavior shifts from Implicit to Explicit prompting. Appendix~\ref{sec:behavioral_case_studies} provides multi-model case studies that instantiate the main flawed-query outcomes and the corresponding failures in validity judgment and SQL generation on original queries.

\begin{table*}[htbp]
\centering
\small
\begin{adjustbox}{width=\textwidth}
\begin{tabular}{l|cccc|ccccc|ccccc}
\toprule
\multirow{3}{*}{Model} &
\multicolumn{4}{c|}{Base} &
\multicolumn{5}{c|}{Implicit} &
\multicolumn{5}{c}{Explicit} \\
\cmidrule(lr){2-5}\cmidrule(lr){6-10}\cmidrule(lr){11-15}
& \multicolumn{3}{c}{Flawed ($+$)} & \multicolumn{1}{c|}{Original ($-$)}
& \multicolumn{3}{c}{Flawed ($+$)} & \multicolumn{2}{c|}{Original ($-$)}
& \multicolumn{3}{c}{Flawed ($+$)} & \multicolumn{2}{c}{Original ($-$)} \\
& SQL & DET & TYPE & GEN
& SQL & DET & TYPE & GEN & VAL
& SQL & DET & TYPE & GEN & VAL \\
\midrule
ChatGLM-5.1
& \cellcolor{teal!12}0.118 & \cellcolor{teal!23}0.228 & \cellcolor{teal!22}0.222 & \cellcolor{teal!100}1.000
& \cellcolor{teal!59}0.585 & \cellcolor{teal!60}0.599 & \cellcolor{teal!57}0.565 & \cellcolor{teal!98}0.981 & \cellcolor{teal!96}0.963
& \cellcolor{teal!98}0.975 & \cellcolor{teal!96}0.957 & \cellcolor{teal!79}0.785 & \cellcolor{teal!52}0.516 & \cellcolor{teal!40}0.396 \\
Claude Opus 4.6
& \cellcolor{teal!20}0.204 & \cellcolor{teal!28}0.281 & \cellcolor{teal!28}0.277 & \cellcolor{teal!99}0.988
& \cellcolor{teal!51}0.510 & \cellcolor{teal!58}0.579 & \cellcolor{teal!54}0.536 & \cellcolor{teal!96}0.962 & \cellcolor{teal!85}0.849
& \cellcolor{teal!88}0.880 & \cellcolor{teal!90}0.899 & \cellcolor{teal!76}0.760 & \cellcolor{teal!78}0.778 & \cellcolor{teal!44}0.441 \\
Claude Opus 5.5
& \cellcolor{teal!40}0.402 & \cellcolor{teal!49}0.492 & \cellcolor{teal!48}0.480 & \cellcolor{teal!98}0.984
& \cellcolor{teal!51}0.509 & \cellcolor{teal!67}0.673 & \cellcolor{teal!65}0.646 & \cellcolor{teal!97}0.969 & \cellcolor{teal!87}0.870
& \cellcolor{teal!73}0.726 & \cellcolor{teal!85}0.853 & \cellcolor{teal!75}0.746 & \cellcolor{teal!79}0.788 & \cellcolor{teal!58}0.578 \\
Claude Sonnet 4.6
& \cellcolor{teal!18}0.182 & \cellcolor{teal!26}0.263 & \cellcolor{teal!26}0.256 & \cellcolor{teal!98}0.984
& \cellcolor{teal!39}0.389 & \cellcolor{teal!48}0.481 & \cellcolor{teal!45}0.448 & \cellcolor{teal!97}0.965 & \cellcolor{teal!87}0.872
& \cellcolor{teal!88}0.877 & \cellcolor{teal!89}0.892 & \cellcolor{teal!74}0.736 & \cellcolor{teal!90}0.901 & \cellcolor{teal!50}0.501 \\
DeepSeek V3.2
& \cellcolor{teal!24}0.240 & \cellcolor{teal!42}0.422 & \cellcolor{teal!39}0.394 & \cellcolor{teal!94}0.938
& \cellcolor{teal!91}0.906 & \cellcolor{teal!90}0.898 & \cellcolor{teal!76}0.764 & \cellcolor{teal!46}0.455 & \cellcolor{teal!44}0.437
& \cellcolor{teal!98}0.983 & \cellcolor{teal!98}0.978 & \cellcolor{teal!77}0.771 & \cellcolor{teal!15}0.149 & \cellcolor{teal!3}0.034 \\
DeepSeek V4.1 Flash
& \cellcolor{teal!11}0.110 & \cellcolor{teal!25}0.247 & \cellcolor{teal!24}0.242 & \cellcolor{teal!99}0.989
& \cellcolor{teal!57}0.566 & \cellcolor{teal!63}0.625 & \cellcolor{teal!57}0.569 & \cellcolor{teal!93}0.926 & \cellcolor{teal!84}0.844
& \cellcolor{teal!97}0.966 & \cellcolor{teal!93}0.932 & \cellcolor{teal!77}0.771 & \cellcolor{teal!31}0.311 & \cellcolor{teal!26}0.264 \\
Doubao Seed 2.0 Pro
& \cellcolor{teal!16}0.163 & \cellcolor{teal!31}0.305 & \cellcolor{teal!30}0.298 & \cellcolor{teal!100}0.998
& \cellcolor{teal!68}0.684 & \cellcolor{teal!68}0.683 & \cellcolor{teal!62}0.620 & \cellcolor{teal!96}0.962 & \cellcolor{teal!77}0.772
& \cellcolor{teal!96}0.958 & \cellcolor{teal!94}0.938 & \cellcolor{teal!80}0.796 & \cellcolor{teal!89}0.891 & \cellcolor{teal!28}0.278 \\
Gemini 3.1 Pro Preview
& \cellcolor{teal!14}0.142 & \cellcolor{teal!27}0.272 & \cellcolor{teal!27}0.272 & \cellcolor{teal!100}1.000
& \cellcolor{teal!71}0.706 & \cellcolor{teal!71}0.706 & \cellcolor{teal!69}0.688 & \cellcolor{teal!100}0.995 & \cellcolor{teal!99}0.990
& \cellcolor{teal!93}0.934 & \cellcolor{teal!92}0.923 & \cellcolor{teal!83}0.829 & \cellcolor{teal!94}0.941 & \cellcolor{teal!70}0.704 \\
LLaMA 4 Maverick
& \cellcolor{teal!10}0.098 & \cellcolor{teal!15}0.147 & \cellcolor{teal!14}0.141 & \cellcolor{teal!100}0.998
& \cellcolor{teal!51}0.505 & \cellcolor{teal!59}0.586 & \cellcolor{teal!51}0.507 & \cellcolor{teal!99}0.987 & \cellcolor{teal!78}0.776
& \cellcolor{teal!94}0.938 & \cellcolor{teal!96}0.960 & \cellcolor{teal!77}0.765 & \cellcolor{teal!97}0.974 & \cellcolor{teal!36}0.363 \\
GPT-5.4
& \cellcolor{teal!12}0.117 & \cellcolor{teal!26}0.261 & \cellcolor{teal!26}0.258 & \cellcolor{teal!100}0.999
& \cellcolor{teal!68}0.682 & \cellcolor{teal!70}0.702 & \cellcolor{teal!65}0.647 & \cellcolor{teal!89}0.892 & \cellcolor{teal!88}0.878
& \cellcolor{teal!96}0.960 & \cellcolor{teal!98}0.981 & \cellcolor{teal!86}0.855 & \cellcolor{teal!22}0.219 & \cellcolor{teal!21}0.207 \\
GPT-5.6-Luna
& \cellcolor{teal!19}0.186 & \cellcolor{teal!31}0.312 & \cellcolor{teal!31}0.308 & \cellcolor{teal!100}0.999
& \cellcolor{teal!62}0.624 & \cellcolor{teal!63}0.629 & \cellcolor{teal!59}0.593 & \cellcolor{teal!96}0.959 & \cellcolor{teal!90}0.904
& \cellcolor{teal!98}0.975 & \cellcolor{teal!90}0.898 & \cellcolor{teal!80}0.796 & \cellcolor{teal!29}0.288 & \cellcolor{teal!23}0.231 \\
GPT-6-Luna
& \cellcolor{teal!20}0.197 & \cellcolor{teal!39}0.386 & \cellcolor{teal!38}0.384 & \cellcolor{teal!100}0.999
& \cellcolor{teal!62}0.623 & \cellcolor{teal!65}0.652 & \cellcolor{teal!63}0.634 & \cellcolor{teal!95}0.954 & \cellcolor{teal!93}0.931
& \cellcolor{teal!87}0.866 & \cellcolor{teal!85}0.852 & \cellcolor{teal!79}0.785 & \cellcolor{teal!56}0.563 & \cellcolor{teal!51}0.507 \\
GPT-6.1-Sol
& \cellcolor{teal!28}0.276 & \cellcolor{teal!42}0.417 & \cellcolor{teal!41}0.413 & \cellcolor{teal!100}1.000
& \cellcolor{teal!69}0.692 & \cellcolor{teal!72}0.717 & \cellcolor{teal!69}0.690 & \cellcolor{teal!96}0.955 & \cellcolor{teal!91}0.906
& \cellcolor{teal!97}0.965 & \cellcolor{teal!93}0.932 & \cellcolor{teal!85}0.845 & \cellcolor{teal!35}0.352 & \cellcolor{teal!32}0.322 \\
Qwen3.5-35B
& \cellcolor{teal!19}0.186 & \cellcolor{teal!40}0.399 & \cellcolor{teal!38}0.384 & \cellcolor{teal!100}0.999
& \cellcolor{teal!78}0.777 & \cellcolor{teal!78}0.780 & \cellcolor{teal!68}0.678 & \cellcolor{teal!94}0.942 & \cellcolor{teal!76}0.759
& \cellcolor{teal!97}0.972 & \cellcolor{teal!96}0.964 & \cellcolor{teal!79}0.786 & \cellcolor{teal!25}0.253 & \cellcolor{teal!13}0.130 \\
Qwen3.5-122B
& \cellcolor{teal!15}0.153 & \cellcolor{teal!35}0.351 & \cellcolor{teal!34}0.338 & \cellcolor{teal!93}0.931
& \cellcolor{teal!71}0.708 & \cellcolor{teal!71}0.711 & \cellcolor{teal!64}0.644 & \cellcolor{teal!81}0.807 & \cellcolor{teal!80}0.795
& \cellcolor{teal!100}0.996 & \cellcolor{teal!96}0.962 & \cellcolor{teal!80}0.797 & \cellcolor{teal!45}0.446 & \cellcolor{teal!18}0.179 \\
Qwen3.5-397B
& \cellcolor{teal!5}0.054 & \cellcolor{teal!32}0.315 & \cellcolor{teal!31}0.312 & \cellcolor{teal!99}0.992
& \cellcolor{teal!69}0.690 & \cellcolor{teal!72}0.723 & \cellcolor{teal!68}0.676 & \cellcolor{teal!74}0.737 & \cellcolor{teal!67}0.665
& \cellcolor{teal!100}0.998 & \cellcolor{teal!97}0.968 & \cellcolor{teal!82}0.817 & \cellcolor{teal!1}0.004 & \cellcolor{teal!1}0.005 \\
\bottomrule
\end{tabular}
\end{adjustbox}
\caption{Premise critique performance on positive and negative samples. SQL is normalized SQL Behavior; DET and TYPE denote Detection and Type Identification. GEN is the SQL-generation rate on original queries, and VAL is the rate at which the model explicitly recognizes those queries as valid. Darker teal indicates better performance.}
\label{tab:overall_results}
\end{table*}

\paragraph{Base Prompting}

Under Base prompting, aggregate positive-sample scores are only 0.177 SQL, 0.319 DET, and 0.311 TYPE, whereas negative-sample GEN is 0.988. Even the strongest Base detector, Claude Opus 5.5 at 0.492, fails to identify more than half of the flaws. LLaMA 4 Maverick detects only 0.147 of flawed cases, though it generates SQL as expected for 0.998 of originals. Standard Text-to-SQL instructions therefore elicit limited spontaneous premise critique: models show weak ability to recognize and interpret flawed premises, while rarely exhibiting necessary abstention on flawed queries.

\paragraph{Implicit Prompting}

Implicit validity guidance increases aggregate SQL, DET, and TYPE to 0.635, 0.672, and 0.619, respectively, while original-query GEN and VAL remain 0.903 and 0.826. The gains are substantial but uneven. DeepSeek V3.2 reaches the highest DET (0.898), yet its GEN and VAL on originals are only 0.455 and 0.437. Gemini 3.1 Pro Preview attains lower DET and TYPE (0.706 and 0.688) while preserving nearly all originals. Thus, a soft warning exposes some latent critique ability, but it neither closes the recognition--explanation gap nor produces uniformly reliable premise judgments.

\paragraph{Explicit Prompting}

Explicit prompting raises aggregate SQL, DET, and TYPE to 0.936, 0.931, and 0.790. Because this prompt asserts that a flaw exists, however, high scores can reflect compliance with that assertion rather than independent critique. The evaluation therefore tests whether critique follows the evidence: models should identify a supported flaw in perturbed queries, but should also reject the prompt's unsupported claim of a flaw when processing originals. The latter requirement distinguishes genuine premise assessment from instruction-driven rejection. Nevertheless, aggregate GEN and VAL on originals fall to 0.523 and 0.321.  Qwen3.5-397B is the clearest rejection case, scoring 0.998 SQL and 0.968 DET on flawed queries but only 0.004 GEN and 0.005 VAL on originals. Gemini 3.1 Pro Preview is less susceptible, retaining 0.941 GEN and 0.704 VAL. Overall, stronger instructions improve apparent critique performance without producing consistently evidence-grounded premise judgments.

\subsubsection{Model Comparisons}

Aggregate trends conceal substantial variation within and across model families. Model rankings also shift across prompting settings.

Within model families, Base DET rises from 0.261 for GPT-5.4 to 0.313 for GPT-5.6-Luna, 0.386 for GPT-6-Luna, and 0.417 for GPT-6.1-Sol; TYPE follows the same ordering. Claude Opus 5.5 likewise improves Base DET from 0.281 to 0.492 relative to Opus 4.6 and raises Implicit TYPE from 0.536 to 0.646. Yet even the best Base result remains below 0.5, and the GPT ordering is not stable under stronger prompts.

Other comparisons reinforce the absence of a general scaling trend. Under Implicit prompting, DeepSeek V4.1 Flash has much higher original-query GEN than V3.2 (0.926 versus 0.455), but lower DET on flawed queries (0.625 versus 0.898). The 35B Qwen variant also outperforms its larger counterparts on Implicit DET. Newer or larger models therefore show partial, setting-dependent gains rather than a consistent improvement in spontaneous premise critique.

\subsubsection{Validity Judgments and Behavior}

We pool 115,200 outputs from Implicit and Explicit prompting and exclude 145 malformed flawed-query responses, retaining 57,455 flawed-query and 57,600 original-query outputs. Among the 47,801 flawed-query outputs with \textit{is\_valid}=\textit{false}, 96.06\% receive DET = 1 and 94.12\% completely abstain; among the 9,654 with \textit{is\_valid}=\textit{true}, 98.66\% receive DET = 0 and 99.92\% generate SQL. Thus, on flawed queries, the reported validity judgment closely tracks both flaw recognition and generation behavior, even when that judgment is incorrect.

The originals show why such consistency does not imply correctness. Of the 24,568 original-query outputs with \textit{is\_valid}=\textit{false}, 16,502 (67.17\%) contain no SQL. By contrast, SQL is generated in 33,010 of the 33,032 original-query outputs (99.93\%) with \textit{is\_valid}=\textit{true}. Models therefore tend to act on their own validity judgment: when the judgment is wrong, the generation decision often follows the same error.

Explicit prompting amplifies this problem. Among outputs declaring the query invalid, the proportion corresponding to flawed queries falls from 20,100 of 25,119 (80.02\%) under Implicit prompting to 27,701 of 47,250 (58.63\%) under Explicit prompting. Equivalently, original-query outputs account for 41.37\% of all invalid judgments under Explicit prompting. The resulting increase in false invalid judgments suggests that models rely too heavily on the prompt's premise that the query is flawed, even when their judgments and generation behavior are internally consistent.

\subsubsection{Automatic Evaluation and Reliability}
\label{sec:reliability}

All model outputs are independently scored by \textbf{GPT-5.4} and \textbf{Claude Opus 4.6} under identical instructions and scoring definitions. Figure~\ref{fig:eval_prompt_ambiguity} presents a representative evaluation prompt, while Appendix~\ref{sec:appendix_eval} provides the complete rubrics and validation details. Inter-evaluator Cohen's $\kappa$ reaches \textbf{0.9737} for SQL Behavior, \textbf{0.9100} for Detection, and \textbf{0.8586} for Type Identification.

We further verify the protocol on 2,160 randomly sampled outputs from one open-weight model (\textbf{DeepSeek V3.2}) and one proprietary model (\textbf{GPT-5.4}). Against two human annotators, the mean Cohen's $\kappa$ scores are 0.9701, 0.9545, and 0.9072 for Claude Opus 4.6, and 0.9567, 0.9355, and 0.8188 for GPT-5.4, on SQL Behavior, Detection, and Type Identification, respectively. These results support the reliability of the automatic evaluation used below.

\subsection{Error-Type Analysis}

Type identification remains more demanding than general flaw detection. TYPE is below DET in every setting, and the aggregate gap widens to 0.141 under Explicit prompting. Table~\ref{tab:error_type_analysis} therefore breaks down the Implicit results by premise-error type.

\begin{table}[htbp]
\centering
\small
\setlength{\tabcolsep}{3pt}
\begin{tabular}{@{}clccc@{}}
\toprule
Code & Error Type & SQL & DET & TYPE \\
\midrule
E1.1 & Ambiguity & \cellcolor{teal!21}0.210 & \cellcolor{teal!24}0.242 & \cellcolor{teal!14}0.138 \\
E1.2 & Unclear Intent & \cellcolor{teal!20}0.202 & \cellcolor{teal!19}0.193 & \cellcolor{teal!13}0.131 \\
E2.1 & Internal Text Contrad. & \cellcolor{teal!92}0.920 & \cellcolor{teal!97}0.972 & \cellcolor{teal!97}0.966 \\
E2.2 & Question--Schema Contrad. & \cellcolor{teal!88}0.881 & \cellcolor{teal!92}0.923 & \cellcolor{teal!92}0.915 \\
E3.1 & Missing Constraint & \cellcolor{teal!54}0.537 & \cellcolor{teal!61}0.611 & \cellcolor{teal!54}0.543 \\
E3.2 & Missing Schema Premise & \cellcolor{teal!57}0.568 & \cellcolor{teal!59}0.586 & \cellcolor{teal!47}0.474 \\
E3.3 & Semantic Gap & \cellcolor{teal!71}0.710 & \cellcolor{teal!76}0.757 & \cellcolor{teal!73}0.726 \\
E4.1 & Inexpressible in SQL & \cellcolor{teal!85}0.853 & \cellcolor{teal!90}0.900 & \cellcolor{teal!84}0.838 \\
E4.2 & External Knowledge Req. & \cellcolor{teal!83}0.830 & \cellcolor{teal!86}0.861 & \cellcolor{teal!84}0.841 \\
\bottomrule
\end{tabular}
\caption{Performance by taxonomy error type under Implicit prompting, aggregated across all 16 models and both datasets with labels averaged over two judges.}
\label{tab:error_type_analysis}
\end{table}

The results reveal a pronounced dependence on how premise flaws are expressed. Contradictions are easiest(DET and TYPE above 0.91), while ambiguity and unclear intent are hardest (DET below 0.25; TYPE below 0.14). Missing-premise cases fall between these extremes; models also perform relatively well on queries that exceed SQL expressibility or require external knowledge. Taken together, the pattern suggests that Implicit prompting elicits critique mainly when a flaw appears as an overt incompatibility or recognizable capability boundary. It is much less effective when detecting the flaw requires reasoning about under-specification or absent information, indicating that premise critique remains cue-dependent rather than robust across error forms.

\subsection{From Error Types to Failure Behaviors}

The preceding error-type analysis establishes which premise flaws are difficult to recognize, but it does not explain what models do with those flaws when critique fails. We therefore move from an input-level taxonomy to a response-level account. Whereas an error category characterizes a query flaw in its database context, a failure mechanism identifies the observable operation through which the model makes the query appear answerable.

We annotate 8,000 eligible failures with four non-exclusive mechanisms: \textbf{Assumption Injection} (M01) introduces an unsupported interpretation, value, or proxy; \textbf{Contradiction Reinterpretation} (M02) modifies incompatible requirements; \textbf{Context/Evidence Misuse} (M03) treats partial schema or example evidence as conclusive; and \textbf{Explicit Constraint Omission} (M04) drops a complete requirement. Attribution requires strict agreement between two judges, with strong automatic and human reliability  (macro $\kappa=0.8183$ and $0.9434$, respectively; see Appendix~\ref{sec:failure_attribution} for details).

\begin{figure}[t]
\centering
\includegraphics[width=0.95\columnwidth]{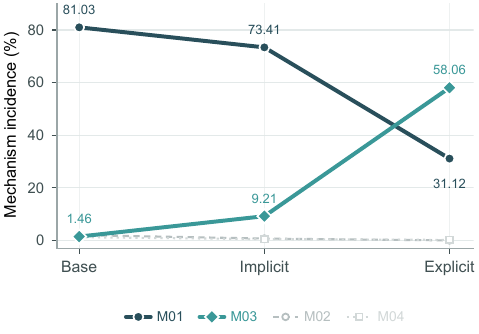}
\caption{Failure-mechanism incidence across prompting settings. Lines show the percentage of eligible failures assigned each mechanism by both judges. The M01--M03 reversal shows that stronger critique instructions shift residual failures from silent assumption injection to overinterpreting limited evidence rather than eliminating them, while M02 and M04 remain rare.}
\label{fig:mechanism_by_prompt}
\end{figure}

Among the eligible failures analyzed across prompting settings, M01 is the dominant failure mechanism (72.60\%), followed by M03 (10.75\%), while M02 (1.38\%) and M04 (0.86\%) remain rare. More importantly, Figure~\ref{fig:mechanism_by_prompt} shows a marked redistribution across prompting settings. From Base to Explicit prompting, M01 decreases from 81.03\% to 31.12\%, whereas M03 increases from 1.46\% to 58.06\%; the Implicit setting lies between these endpoints. Stronger instructions suppress silent completion of missing query premises, yet the remaining failures increasingly arise because models overinterpret limited schema or example evidence when deciding whether, and how, the premise is defective. The association with prompting setting is substantial for M01 (Cram\'er's $V=0.356$) and stronger for M03 ($V=0.581$), with Holm-adjusted $p<0.001$ in both cases.

Mechanism-level analysis therefore complements the error taxonomy by showing how an observed premise flaw is converted into an apparently answerable query. Despite their different surface forms, these failures reveal a shared limitation in that models tend to produce apparently answerable interpretations without recognizing the underlying premise flaws.
\section{Conclusion}

In this work, we introduce \textbf{PremiseSQL}, to our knowledge the first benchmark for evaluating premise critique ability in NL2SQL systems. PremiseSQL contains 1,800 flawed queries covering 9 fine-grained error types and their 1,800 originals as positive and negative samples, respectively.
Experiments under Base, Implicit, and Explicit prompting settings reveal a clear reliability gap in current NL2SQL systems. Models often trust user assumptions and generate misleading SQL under standard settings. Even when explicitly prompted to scrutinize premises, they struggle to assess them reliably. Models further struggle with subtle and implicit premise flaws, showing that strong SQL generation does not necessarily imply trustworthy premise-aware reasoning.
We hope PremiseSQL can facilitate future research on more reliable and premise-aware NL2SQL systems.

\section*{Limitations}

Our work has several limitations. 
First, PremiseSQL focuses on English NL2SQL benchmarks and may not generalize to multilingual settings or real-world interactive database systems. 
Second, although the benchmark covers multiple premise failure categories, the constructed errors are still taxonomy-driven and may not fully capture naturally occurring user mistakes. 
Finally, our evaluation relies partially on LLM-based automatic judges, which may introduce evaluation bias despite strong agreement with human annotations. 

\bibliography{custom}

@inproceedings{10.1145/604045.604070,
author = {Popescu, Ana-Maria and Etzioni, Oren and Kautz, Henry},
title = {Towards a theory of natural language interfaces to databases},
year = {2003},
isbn = {1581135866},
publisher = {Association for Computing Machinery},
address = {New York, NY, USA},
url = {https://doi.org/10.1145/604045.604070},
doi = {10.1145/604045.604070},
booktitle = {Proceedings of the 8th International Conference on Intelligent User Interfaces},
pages = {149–157},
numpages = {9},
location = {Miami, Florida, USA},
series = {IUI '03}
}

@misc{deepseekai2025deepseekv32pushingfrontieropen,
      title={DeepSeek-V3.2: Pushing the Frontier of Open Large Language Models}, 
     author={{DeepSeek AI}},
      year={2025},
      eprint={2512.02556},
      archivePrefix={arXiv},
      primaryClass={cs.CL},
      url={https://arxiv.org/abs/2512.02556}, 
}

@misc{deepseekai2026deepseekv41flash,
      title={{DeepSeek-V4.1-Flash}: Pushing the Limits of {KV} Cache Compression},
      author={{DeepSeek-AI}},
      year={2026},
      eprint={2609.19969},
      archivePrefix={arXiv},
      primaryClass={cs.CL},
      url={https://arxiv.org/abs/2609.19969},
}

@article{qwen3,
    title={Qwen3 Technical Report}, 
    author={An Yang and Anfeng Li and Baosong Yang  and et al.},
    journal = {arXiv preprint arXiv:2505.09388},
    year={2025}
}

@misc{openai2024gpt4technicalreport,
      title={GPT-4 Technical Report}, 
      author={OpenAI and Josh Achiam and Steven Adler  and et al.},
      year={2024},
      eprint={2303.08774},
      archivePrefix={arXiv},
      primaryClass={cs.CL},
      url={https://arxiv.org/abs/2303.08774}, 
}

@misc{touvron2023llamaopenefficientfoundation,
      title={LLaMA: Open and Efficient Foundation Language Models}, 
      author={Hugo Touvron and Thibaut Lavril and Gautier Izacard and Xavier Martinet and Marie-Anne Lachaux and Timothée Lacroix and Baptiste Rozière and Naman Goyal and Eric Hambro and Faisal Azhar and Aurelien Rodriguez and Armand Joulin and Edouard Grave and Guillaume Lample},
      year={2023},
      eprint={2302.13971},
      archivePrefix={arXiv},
      primaryClass={cs.CL},
      url={https://arxiv.org/abs/2302.13971}, 
}

@inproceedings{yu-etal-2018-spider,
    title = "{S}pider: A Large-Scale Human-Labeled Dataset for Complex and Cross-Domain Semantic Parsing and Text-to-{SQL} Task",
    author = "Yu, Tao  and
      Zhang, Rui  and
      Yang, Kai  and
      Yasunaga, Michihiro  and
      Wang, Dongxu  and
      Li, Zifan  and
      Ma, James  and
      Li, Irene  and
      Yao, Qingning  and
      Roman, Shanelle  and
      Zhang, Zilin  and
      Radev, Dragomir",
    editor = "Riloff, Ellen  and
      Chiang, David  and
      Hockenmaier, Julia  and
      Tsujii, Jun{'}ichi",
    booktitle = "Proceedings of the 2018 Conference on Empirical Methods in Natural Language Processing",
    month = oct # "-" # nov,
    year = "2018",
    address = "Brussels, Belgium",
    publisher = "Association for Computational Linguistics",
    url = "https://aclanthology.org/D18-1425/",
    doi = "10.18653/v1/D18-1425",
    pages = "3911--3921"
}

@misc{scholak2021picardparsingincrementallyconstrained,
      title={PICARD: Parsing Incrementally for Constrained Auto-Regressive Decoding from Language Models}, 
      author={Torsten Scholak and Nathan Schucher and Dzmitry Bahdanau},
      year={2021},
      eprint={2109.05093},
      archivePrefix={arXiv},
      primaryClass={cs.CL},
      url={https://arxiv.org/abs/2109.05093}, 
}

@inproceedings{pourreza2023din,
author = {Pourreza, Mohammadreza and Rafiei, Davood},
title = {DIN-SQL: decomposed in-context learning of text-to-SQL with self-correction},
year = {2023},
publisher = {Curran Associates Inc.},
address = {Red Hook, NY, USA},
booktitle = {Proceedings of the 37th International Conference on Neural Information Processing Systems},
articleno = {1577},
numpages = {10},
location = {New Orleans, LA, USA},
series = {NIPS '23}
}

@inproceedings{li2024can,
author = {Li, Jinyang and Hui, Binyuan and Qu, Ge and Yang, Jiaxi and Li, Binhua and Li, Bowen and Wang, Bailin and Qin, Bowen and Geng, Ruiying and Huo, Nan and Zhou, Xuanhe and Ma, Chenhao and Li, Guoliang and Chang, Kevin C.C. and Huang, Fei and Cheng, Reynold and Li, Yongbin},
title = {Can LLM already serve as a database interface? a big bench for large-scale database grounded text-to-SQLs},
year = {2023},
publisher = {Curran Associates Inc.},
address = {Red Hook, NY, USA},
booktitle = {Proceedings of the 37th International Conference on Neural Information Processing Systems},
articleno = {1835},
numpages = {28},
location = {New Orleans, LA, USA},
series = {NIPS '23}
}

@inproceedings{wang-etal-2020-rat,
    title = "{RAT-SQL}: Relation-Aware Schema Encoding and Linking for Text-to-{SQL} Parsers",
    author = "Wang, Bailin  and
      Shin, Richard  and
      Liu, Xiaodong  and
      Polozov, Oleksandr  and
      Richardson, Matthew",
    editor = "Jurafsky, Dan  and
      Chai, Joyce  and
      Schluter, Natalie  and
      Tetreault, Joel",
    booktitle = "Proceedings of the 58th Annual Meeting of the Association for Computational Linguistics",
    month = jul,
    year = "2020",
    address = "Online",
    publisher = "Association for Computational Linguistics",
    url = "https://aclanthology.org/2020.acl-main.677/",
    doi = "10.18653/v1/2020.acl-main.677",
    pages = "7567--7578"
}

@article{10.1145/3571730,
author = {Ji, Ziwei and Lee, Nayeon and Frieske, Rita and Yu, Tiezheng and Su, Dan and Xu, Yan and Ishii, Etsuko and Bang, Ye Jin and Madotto, Andrea and Fung, Pascale},
title = {Survey of Hallucination in Natural Language Generation},
year = {2023},
issue_date = {December 2023},
publisher = {Association for Computing Machinery},
address = {New York, NY, USA},
volume = {55},
number = {12},
issn = {0360-0300},
url = {https://doi.org/10.1145/3571730},
doi = {10.1145/3571730},
journal = {ACM Comput. Surv.},
month = mar,
articleno = {248},
numpages = {38}
}

@inproceedings{10.1007/978-3-031-20059-5_9,
author = {Whitehead, Spencer and Petryk, Suzanne and Shakib, Vedaad and Gonzalez, Joseph and Darrell, Trevor and Rohrbach, Anna and Rohrbach, Marcus},
title = {Reliable Visual Question Answering: Abstain Rather Than Answer Incorrectly},
year = {2022},
isbn = {978-3-031-20058-8},
publisher = {Springer-Verlag},
address = {Berlin, Heidelberg},
url = {https://doi.org/10.1007/978-3-031-20059-5_9},
doi = {10.1007/978-3-031-20059-5_9},
booktitle = {Computer Vision – ECCV 2022: 17th European Conference, Tel Aviv, Israel, October 23–27, 2022, Proceedings, Part XXXVI},
pages = {148–166},
numpages = {19},
location = {Tel Aviv, Israel}
}

@misc{ren2023outofdistributiondetectionselectivegeneration,
      title={Out-of-Distribution Detection and Selective Generation for Conditional Language Models}, 
      author={Jie Ren and Jiaming Luo and Yao Zhao and Kundan Krishna and Mohammad Saleh and Balaji Lakshminarayanan and Peter J. Liu},
      year={2023},
      eprint={2209.15558},
      archivePrefix={arXiv},
      primaryClass={cs.CL},
      url={https://arxiv.org/abs/2209.15558}, 
}

@article{zhang2020did,
  title={Did You Ask a Good Question? A Cross-Domain Question Intention Classification Benchmark for Text-to-SQL},
  author={Zhang, Yusen and Dong, Xiangyu and Chang, Shuaichen and Yu, Tao and Shi, Peng and Zhang, Rui},
  journal={arXiv preprint arXiv:2010.12634},
  year={2020}
}

@inproceedings{wang-etal-2023-know,
    title = "Know What {I} don{'}t Know: Handling Ambiguous and Unknown Questions for Text-to-{SQL}",
    author = "Wang, Bing  and
      Gao, Yan  and
      Li, Zhoujun  and
      Lou, Jian-Guang",
    editor = "Rogers, Anna  and
      Boyd-Graber, Jordan  and
      Okazaki, Naoaki",
    booktitle = "Findings of the Association for Computational Linguistics: ACL 2023",
    month = jul,
    year = "2023",
    address = "Toronto, Canada",
    publisher = "Association for Computational Linguistics",
    url = "https://aclanthology.org/2023.findings-acl.352/",
    doi = "10.18653/v1/2023.findings-acl.352",
    pages = "5701--5714"
}

@misc{
lee2024trustsql,
title={Trust{SQL}: Benchmarking Text-to-{SQL} Reliability with Penalty-Based Scoring},
author={Gyubok Lee and Woosog Chay and Seonhee Cho and Edward Choi},
year={2024},
url={https://openreview.net/forum?id=7ZeoPg3eTA}
}

@inproceedings{10.1145/3711896.3737427,
author = {Liu, Xinyu and Shen, Shuyu and Li, Boyan and Tang, Nan and Luo, Yuyu},
title = {NL2SQL-BUGs: A Benchmark for Detecting Semantic Errors in NL2SQL Translation},
year = {2025},
isbn = {9798400714542},
publisher = {Association for Computing Machinery},
address = {New York, NY, USA},
url = {https://doi.org/10.1145/3711896.3737427},
doi = {10.1145/3711896.3737427},
booktitle = {Proceedings of the 31st ACM SIGKDD Conference on Knowledge Discovery and Data Mining V.2},
pages = {5662–5673},
numpages = {12},
location = {Toronto ON, Canada},
series = {KDD '25}
}

@inproceedings{rajpurkar-etal-2018-know,
    title = "Know What You Don{'}t Know: Unanswerable Questions for {SQ}u{AD}",
    author = "Rajpurkar, Pranav  and
      Jia, Robin  and
      Liang, Percy",
    editor = "Gurevych, Iryna  and
      Miyao, Yusuke",
    booktitle = "Proceedings of the 56th Annual Meeting of the Association for Computational Linguistics (Volume 2: Short Papers)",
    month = jul,
    year = "2018",
    address = "Melbourne, Australia",
    publisher = "Association for Computational Linguistics",
    url = "https://aclanthology.org/P18-2124/",
    doi = "10.18653/v1/P18-2124",
    pages = "784--789"
}

@inproceedings{kirichenko2025abstentionbench,
 author = {Kirichenko, Polina and Ibrahim, Mark and Chaudhuri, Kamalika and Bell, Samuel J.},
 booktitle = {Advances in Neural Information Processing Systems},
 editor = {D. Belgrave and C. Zhang and H. Lin and R. Pascanu and P. Koniusz and M. Ghassemi and N. Chen},
 publisher = {Curran Associates, Inc.},
 title = {AbstentionBench: Reasoning LLMs Fail on Unanswerable Questions},
 url = {https://proceedings.neurips.cc/paper_files/paper/2025/file/fb122bfc3f0127a94ded048b5b03496f-Paper-Datasets_and_Benchmarks_Track.pdf},
 volume = {38},
 year = {2025}
}

@misc{glm5team2026glm5vibecodingagentic,
      title={GLM-5: from Vibe Coding to Agentic Engineering},
      author={{GLM-5 Team}},
      year={2026},
      eprint={2602.15763},
      archivePrefix={arXiv},
      primaryClass={cs.LG},
      url={https://arxiv.org/abs/2602.15763},
}

@misc{anthropic2026claudeopus46,
    title={Introducing Claude Opus 4.6},
    author={Anthropic},
    year={2026},
    howpublished={\url{https://www.anthropic.com/news/claude-opus-4-6}},
    note={Accessed: 2026-02}
}

@misc{anthropic2026claudesonnet46,
    title={Introducing Claude 4.6 Sonnet},
    author={Anthropic},
    year={2026},
    howpublished={\url{https://www.anthropic.com/news/claude-sonnet-4-6}},
    note={Accessed: 2026-02}
}

@misc{anthropic2026claudeopus55,
    title={Introducing Claude Opus 5.5},
    author={Anthropic},
    year={2026},
    month={sep},
    howpublished={\url{https://www.anthropic.com/claude-opus-5-5}},
    note={Accessed: 2026-10-05}
}

@misc{bytedance2026seed2techreport,
    title={Seed2.0 Technical Introduction},
    author={ByteDance},
    year={2026},
    howpublished={\url{https://research.doubao.com/zh/seed2}},
    note={Accessed: 2026-02}
}

@misc{google2026gemini31propreview,
    title={Gemini 3.1 Pro Preview},
    author={Google},
    year={2026},
    howpublished={\url{https://docs.cloud.google.com/vertex-ai/generative-ai/docs/models/gemini/3-1-pro}},
    note={Accessed: 2026-02}
}

@misc{meta2026llama4herd,
      title={The Llama 4 Herd: Architecture, Training, Evaluation, and Deployment Notes},
      author={{Meta AI}},
      year={2026},
      eprint={2601.11659},
      archivePrefix={arXiv},
      primaryClass={cs.SE},
      url={https://arxiv.org/abs/2601.11659},
}

@misc{openai2026gpt54,
    title={Introducing GPT-5.4},
    author={{OpenAI}},
    year={2026},
    howpublished={\url{https://openai.com/index/introducing-gpt-5-4/}},
    note={Accessed: 2026-03}
}

@misc{openai2026gpt56,
    title={{GPT-5.6}: Frontier Intelligence That Scales with Your Ambition},
    author={{OpenAI}},
    year={2026},
    month={jul},
    howpublished={\url{https://openai.com/index/gpt-5-6/}},
    note={Accessed: 2026-10-05}
}

@misc{openai2026gpt6solluna,
    title={Introducing {GPT-6 Sol and Luna}},
    author={{OpenAI}},
    year={2026},
    month={sep},
    howpublished={\url{https://openai.com/index/introducing-gpt-6-sol-and-luna/}},
    note={Accessed: 2026-10-05}
}

@misc{openai2026gpt61sol,
    title={Introducing {GPT-6.1 Sol}},
    author={{OpenAI}},
    year={2026},
    month={sep},
    howpublished={\url{https://openai.com/index/introducing-gpt-6-1-sol/}},
    note={Accessed: 2026-10-05}
}

@misc{qwen3.5,
    title  = {{Qwen3.5}: Towards Native Multimodal Agents},
    author = {{Qwen Team}},
    month  = {February},
    year   = {2026},
    url    = {https://qwen.ai/blog?id=qwen3.5}
}

@inproceedings{lin-etal-2022-truthfulqa,
    title = "{T}ruthful{QA}: Measuring How Models Mimic Human Falsehoods",
    author = "Lin, Stephanie  and
      Hilton, Jacob  and
      Evans, Owain",
    editor = "Muresan, Smaranda  and
      Nakov, Preslav  and
      Villavicencio, Aline",
    booktitle = "Proceedings of the 60th Annual Meeting of the Association for Computational Linguistics (Volume 1: Long Papers)",
    month = may,
    year = "2022",
    address = "Dublin, Ireland",
    publisher = "Association for Computational Linguistics",
    url = "https://aclanthology.org/2022.acl-long.229/",
    doi = "10.18653/v1/2022.acl-long.229",
    pages = "3214--3252"
}

@article{10.1145/3703155,
author = {Huang, Lei and Yu, Weijiang and Ma, Weitao and Zhong, Weihong and Feng, Zhangyin and Wang, Haotian and Chen, Qianglong and Peng, Weihua and Feng, Xiaocheng and Qin, Bing and Liu, Ting},
title = {A Survey on Hallucination in Large Language Models: Principles, Taxonomy, Challenges, and Open Questions},
year = {2025},
issue_date = {March 2025},
publisher = {Association for Computing Machinery},
address = {New York, NY, USA},
volume = {43},
number = {2},
issn = {1046-8188},
url = {https://doi.org/10.1145/3703155},
doi = {10.1145/3703155},
journal = {ACM Trans. Inf. Syst.},
month = jan,
articleno = {42},
numpages = {55}
}

@article{10.1145/3641289,
author = {Chang, Yupeng and Wang, Xu and Wang, Jindong and Wu, Yuan and Yang, Linyi and Zhu, Kaijie and Chen, Hao and Yi, Xiaoyuan and Wang, Cunxiang and Wang, Yidong and Ye, Wei and Zhang, Yue and Chang, Yi and Yu, Philip S. and Yang, Qiang and Xie, Xing},
title = {A Survey on Evaluation of Large Language Models},
year = {2024},
issue_date = {June 2024},
publisher = {Association for Computing Machinery},
address = {New York, NY, USA},
volume = {15},
number = {3},
issn = {2157-6904},
url = {https://doi.org/10.1145/3641289},
doi = {10.1145/3641289},
journal = {ACM Trans. Intell. Syst. Technol.},
month = mar,
articleno = {39},
numpages = {45}
}
\appendix
\section{Benchmark Construction Details}
\label{sec:appendix_construction}

\subsection{Dataset Composition}

PremiseSQL consists of 1,800 flawed NL2SQL queries across four major premise error categories and nine fine-grained subcategories, together with their 1,800 originals as positive and negative samples, respectively.
Each subcategory contains 200 positive samples. To preserve natural query diversity and realistic schema distributions, we do not strictly enforce identical source proportions across Spider 1.0 and BIRD / BIRD-filtered databases. The detailed statistics of positive samples across error categories and source benchmarks are summarized in Table~\ref{tab:premisesql_stats}.

\begin{table}[ht]
\centering
\small

\begin{tabular}{lcc}
\toprule
Error Category / Subcategory & Spider & BIRD \\
\midrule

Expression Error (400) &  &  \\
\quad -- Ambiguity & 90 & 110 \\
\quad -- Unclear Intent & 107 & 93 \\

\addlinespace[2pt]

Premise Contradiction (400) &  &  \\
\quad -- Internal Text Contradiction & 103 & 97 \\
\quad -- Question--Schema Contradiction & 156 & 44 \\

\addlinespace[2pt]

Missing Premise (600) &  &  \\
\quad -- Missing Constraint & 29 & 171 \\
\quad -- Missing Schema Premise & 64 & 136 \\
\quad -- Semantic Gap & 155 & 45 \\

\addlinespace[2pt]

Beyond Ability (400) &  &  \\
\quad -- Inexpressible in SQL & 149 & 51 \\
\quad -- External Knowledge Required & 140 & 60 \\

\midrule
Total & 993 & 807 \\
\bottomrule
\end{tabular}
\caption{Statistics of PremiseSQL across fine-grained premise error categories and source benchmarks.}
\label{tab:premisesql_stats}
\end{table}

\subsection{Construction and Filtering}
\label{app:filter_stat}

For the eight subcategories requiring perturbation-based construction, we generate 15,192 candidate instances and apply two-stage quality control, consisting of LLM verification followed by human review. The LLM verification uses the review prompts described in Section~\ref{appendix:review_prompt_templates}. A total of 2,248 instances entered the human-approved pool; these are pre-selection counts and therefore may exceed the 200 examples ultimately retained per subcategory. The average rejection rates were 57.8\% during LLM verification and 64.9\% during human review. Table~\ref{tab:construction_stats} distinguishes the approved pool from the final allocation of positive samples. E3.2 (Missing Schema Premise) is constructed separately through schema-sensitive perturbations and therefore does not undergo this candidate filtering process.

\begin{table}[ht]
\centering
\footnotesize
\setlength{\tabcolsep}{3.5pt}
\begin{tabular}{lrrrr}
\toprule
Subcat. & Cand. & LLM Pass & Human Appr. & Final \\
\midrule
E1.1 & 4,280 & 708 & 200 & 200 \\
E1.2 & 1,108 & 673 & 382 & 200 \\
\midrule
E2.1 & 1,108 & 886 & 457 & 200 \\
E2.2 & 1,108 & 933 & 350 & 200 \\
\midrule
E3.1 & 2,758 & 692 & 252 & 200 \\
E3.2 & -- & -- & -- & 200 \\
E3.3 & 1,108 & 737 & 204 & 200 \\
\midrule
E4.1 & 1,107 & 627 & 203 & 200 \\
E4.2 & 2,615 & 1,152 & 200 & 200 \\
\midrule
Total & 15,192 & 6,408 & 2,248 & 1,800 \\
\bottomrule
\end{tabular}
\caption{Candidate generation and filtering statistics. Cand. denotes generated candidates, LLM Pass denotes candidates passing LLM review, Human Appr. denotes those retained after manual review, and Final denotes the benchmark allocation. E3.2 is constructed separately.}
\label{tab:construction_stats}
\end{table}

\subsection{Generation and Review Prompts}

Benchmark construction uses two structured prompt stages. The generation stage introduces one controlled premise flaw into a source Text-to-SQL instance, and the review stage filters candidates against category-specific validity criteria. The following prompts show the shared structure through the Ambiguity category; other categories substitute their own definitions and checks.

\subsubsection{Generation Prompt}
\label{appendix:prompt_templates}

We design a unified prompt structure for error synthesis.
We show one representative example for the \textbf{Ambiguity} category in Figure \ref{fig:prompt_ambiguity}.
Other categories follow the same structure with category-specific definitions.

\subsubsection{Review Prompt}
\label{appendix:review_prompt_templates}

We design a unified prompt template for automatic quality control and validation.
We show one representative example for the \textbf{Ambiguity} category in Figure \ref{fig:prompt_review_ambiguity}.
Other categories follow the same review structure with category-specific evaluation criteria.

\section{Experimental and Evaluation Protocol}

\subsection{Base, Implicit, and Explicit Prompts}

We provide the prompt templates used for the three prompting settings in our experiments: \textit{Base}, \textit{Implicit}, and \textit{Explicit}. All prompts follow a unified input format consisting of the database schema and the natural language question.

\paragraph{Base Prompt.}
\begin{quote}
The database schema is as follows: \{schema\}

Question: \{question\}

Generate SQL to answer the question.

\begin{verbatim}
{
  "sql": string or null,
  "reason": string
}
\end{verbatim}

Return structured JSON only.
\end{quote}

\paragraph{Implicit Prompt.}
\begin{quote}
The database schema is as follows: \{schema\}

Question: \{question\}

The question may contain errors.
You must assess its validity before answering.

Generate SQL to answer the question.

\begin{verbatim}
{
  "sql": string or null,
  "is_valid": boolean,
  "reason": string
}
\end{verbatim}

Return structured JSON only.
\end{quote}

\paragraph{Explicit Prompt.}
\begin{quote}
The database schema is as follows: \{schema\}

Question: \{question\}

The question is logically flawed and contains incorrect premises.
You must critically examine its assumptions. Then explain the problem.

\begin{verbatim}
{
  "sql": string or null,
  "is_valid": boolean,
  "reason": string
}
\end{verbatim}

Return structured JSON only.
\end{quote}

\subsection{Evaluation Rubrics}
\label{sec:appendix_eval}

\paragraph{Automatic Evaluation Protocol.}
For each positive-sample output, an evaluator receives the user question, database schema, model response, and ground-truth premise-error type. It then applies a unified instruction and the category-specific definition to assign \textit{SQL Behavior}, \textit{Detection}, and \textit{Type Identification} labels according to the rubrics below. For each negative-sample output, the evaluator assigns \textit{SQL Generation} and \textit{Validity Judgment} labels based on whether the model generates SQL and recognizes the original query as valid. GPT-5.4 and Claude Opus 4.6 perform the evaluation independently under identical instructions and scoring definitions, and we average their judgments to obtain the reported scores. Figure~\ref{fig:eval_prompt_ambiguity} provides the full positive-sample evaluation prompt for a representative subcategory; the remaining subcategories use the same structure with their corresponding definitions.

\paragraph{Positive-Sample Metrics.}

\textbf{SQL Behavior.}
Measures the degree of abstention from generating SQL for flawed queries.

\begin{itemize}
    \item 0: generates complete SQL;
    \item 1: partial abstention or invalid SQL output;
    \item 2: complete abstention or explicit refusal.
\end{itemize}

\textbf{Detection.}
Measures whether the model recognizes the query as flawed.

\begin{itemize}
    \item 0: no detection;
    \item 1: detects flawed premises.
\end{itemize}

\textbf{Type Identification.}
Measures whether the model correctly identifies the underlying flaw type.

\begin{itemize}
    \item 0: incorrect type identification;
    \item 1: correct type identification.
\end{itemize}

Type Identification is assigned 0 when Detection = 0; otherwise, it evaluates whether the corresponding failure type is correctly identified.

\paragraph{Negative-Sample Metrics.}

\textbf{SQL Generation.}
Measures whether the model generates SQL for a negative sample.

\begin{itemize}
    \item 0: does not generate SQL;
    \item 1: generates SQL.
\end{itemize}

GEN is reported as the proportion of negative samples for which SQL is generated and is available in all three prompting settings.

\textbf{Validity Judgment.}
Measures whether the model explicitly recognizes a negative sample as valid.

\begin{itemize}
    \item 0: does not explicitly judge the query valid;
    \item 1: explicitly judges the query valid through \textit{is\_valid=true}.
\end{itemize}

VAL is reported as the proportion of negative samples explicitly judged valid. It is available only in the Implicit and Explicit prompting settings because the Base output format does not include \textit{is\_valid}.

\subsection{Automatic Evaluation Prompt}

We use LLM-based evaluation prompts to assess model behavior on flawed queries.
The evaluation focuses on three metrics:
\textit{SQL Behavior}, \textit{Detection}, and \textit{Type Identification}.

Given a user question, model response, and the ground-truth flaw category,
the automatic evaluator determines whether the model:
(1) refrains from generating misleading SQL,
(2) recognizes that the question is flawed, and
(3) correctly identifies the underlying flaw type.

We adopt a unified evaluation prompt structure across all error categories,
with category-specific definitions and scoring criteria.
Figure~\ref{fig:eval_prompt_ambiguity} shows a representative example
for the \textbf{Ambiguity} subcategory, used only for positive-sample evaluation.

\subsection{Human Verification}

We conduct human verification on 2,160 outputs from one open-weight model (\textbf{DeepSeek V3.2}) and one proprietary model (\textbf{GPT-5.4}). The sample is evenly stratified: each model and each source benchmark contributes 1,080 outputs, each prompting setting contributes 720, and each of the nine fine-grained error types contributes 240, yielding 80 outputs per prompting-setting--error-type combination. Human annotators follow the same evaluation protocol as the automatic evaluators.
Two graduate-level annotators familiar with SQL and database systems independently evaluate the same 2,160 sampled outputs using the rubrics applied by the automatic evaluators.
Agreement between the two human annotators is high, with Cohen's $\kappa$ of \textbf{0.9912}, \textbf{0.9632}, and \textbf{0.8759} for SQL Behavior, Detection, and Type Identification, respectively.

\subsection{Robustness to Judge--Target Overlap}
\label{sec:judge_target_overlap}

GPT-5.4 and Claude Opus 4.6 serve as both evaluated models and automatic evaluators, creating potential judge--target overlap. Although averaging their judgments reduces the influence of either evaluator, it does not by itself eliminate possible self-evaluation or model-family bias. We therefore conduct a leave-self-out robustness check. For GPT-5.4 outputs, we retain only the judgments of Claude Opus 4.6; for Claude Opus 4.6 outputs, we retain only those of GPT-5.4.

\begin{table*}[t]
\centering
\small
\setlength{\tabcolsep}{4.5pt}
\begin{tabular}{llccc|ccc}
\toprule
\multirow{2}{*}{Target Model} &
\multirow{2}{*}{Setting} &
\multicolumn{3}{c|}{Reported Two-Judge Average} &
\multicolumn{3}{c}{Cross-Family Judge Only} \\
\cmidrule(lr){3-5}\cmidrule(lr){6-8}
& & SQL & DET & TYPE & SQL & DET & TYPE \\
\midrule
\multirow{3}{*}{GPT-5.4}
& Base     & 0.117 & 0.261 & 0.258 & 0.109 & 0.240 & 0.239 \\
& Implicit & 0.682 & 0.702 & 0.648 & 0.681 & 0.697 & 0.671 \\
& Explicit & 0.960 & 0.981 & 0.855 & 0.959 & 0.982 & 0.907 \\
\midrule
\multirow{3}{*}{Claude Opus 4.6}
& Base     & 0.204 & 0.281 & 0.277 & 0.205 & 0.296 & 0.288 \\
& Implicit & 0.510 & 0.579 & 0.536 & 0.513 & 0.579 & 0.523 \\
& Explicit & 0.880 & 0.899 & 0.760 & 0.882 & 0.896 & 0.714 \\
\bottomrule
\end{tabular}
\caption{Leave-self-out evaluation for models that also serve as automatic evaluators. The reported scores average judgments from GPT-5.4 and Claude Opus 4.6. The cross-family scores retain only Claude Opus 4.6 judgments for GPT-5.4 outputs and only GPT-5.4 judgments for Claude Opus 4.6 outputs.}
\label{tab:judge_target_overlap}
\end{table*}

Table~\ref{tab:judge_target_overlap} shows no systematic self-favoring pattern. Self-evaluation slightly raises some Base scores but lowers others, including GPT-5.4's Explicit TYPE score. After removing self-evaluation, the aggregate SQL, DET, and TYPE scores change by less than 0.001 in every prompting setting. The leading models also remain unchanged: Claude Opus 5.5 under Base prompting, DeepSeek V3.2 under Implicit prompting, and GPT-5.4 for Explicit DET and TYPE.

We further apply a stricter family-disjoint evaluation in which all OpenAI models are scored only by Claude Opus 4.6 and all Claude models are scored only by GPT-5.4; models from other families retain the two-judge average. Across the nine setting--metric combinations, aggregate scores change by at most 0.0016, the top-ranked model remains unchanged, and the original and family-disjoint model rankings have Spearman correlations ranging from 0.944 to 1.000. These results indicate that judge--target family overlap does not materially affect the aggregate results, leading models, or conclusions about prompting behavior. GEN and VAL are unaffected because they are derived directly from the models' structured outputs rather than from automatic evaluator judgments.

\subsection{Paired Evaluation Design}

The paired inputs expose a progression from under-critique to instruction-driven rejection. Each pair contrasts an original query--context combination with its perturbed counterpart. Reliable critique should change the response to the perturbation while maintaining correct validity judgments and SQL generation for originals. Base prompting usually answers both; stronger prompts increase abstention, but Explicit prompting often suppresses originals as well. The separation among DET, TYPE, and SQL on flawed queries and VAL and GEN on originals therefore distinguishes evidence-based critique from broad rejection.

The five measures provide complementary evidence about premise critique behavior. DET records whether any flaw is noticed, TYPE whether the intended defect is explained, and SQL whether that assessment changes generation. On originals, VAL records the explicit judgment and GEN the action. Their divergence is informative: under Explicit prompting, LLaMA 4 Maverick retains 0.974 GEN but only 0.363 VAL, so it often generates despite incorrectly judging an original query as flawed; Qwen3.5-397B is low on both and instead follows a broad rejection policy. Evaluation on originals prevents either pattern from being mistaken for robust critique.

\subsection{Execution-Based Validation of SQL Generation}
\label{sec:gen_execution_validation}

GEN is designed to measure generation coverage, namely whether a model preserves SQL generation on valid original queries. GEN and VAL are evaluated on all 1,800 original queries. For the supplementary execution-based analysis, we retain 1,726 originals whose gold SQL produces a reliable reference result in our SQLite evaluation environment; the remaining 74 are excluded only from this analysis because their execution correctness cannot be determined reliably. This yields 27,616 model--query outputs per prompting setting across the 16 evaluated models.

Let $g_i$ indicate whether SQL is generated for the $i$-th original query, and let $e_i$ indicate whether the generated SQL is execution-equivalent to the gold SQL, with $e_i=0$ when no SQL is generated. We compute overall execution accuracy as
\begin{equation}
\mathrm{EX}
=
\frac{1}{N}\sum_{i=1}^{N} e_i,
\end{equation}
and execution accuracy conditional on generation as
\begin{equation}
\mathrm{EX}\mid\mathrm{GEN}
=
\frac{\sum_{i=1}^{N} e_i}
     {\sum_{i=1}^{N} g_i}.
\end{equation}

\begin{table}[t]
\centering
\small
\setlength{\tabcolsep}{6pt}
\begin{tabular}{lccc}
\toprule
Setting & GEN & EX & $\mathrm{EX}\mid\mathrm{GEN}$ \\
\midrule
Base     & 0.987 & 0.713 & 0.722 \\
Implicit & 0.902 & 0.652 & 0.722 \\
Explicit & 0.524 & 0.399 & 0.762 \\
\bottomrule
\end{tabular}
\caption{Execution-based validation on the 1,726 original queries with reliably executable gold SQL. EX measures execution accuracy over these queries, treating abstention as incorrect, while $\mathrm{EX}\mid\mathrm{GEN}$ measures execution accuracy among outputs for which SQL is generated.}
\label{tab:gen_execution_validation}
\end{table}

As shown in Table~\ref{tab:gen_execution_validation}, GEN closely tracks overall execution accuracy, whereas execution accuracy conditional on generation remains comparatively stable across prompting settings. Across the 48 model--setting combinations, GEN is strongly correlated with EX (Pearson's $r=0.990$, $p<.001$; Spearman's $\rho=0.888$, $p<.001$). The association remains strong when models are compared within the Implicit ($r=0.951$, $p<.001$) and Explicit ($r=0.997$, $p<.001$) settings. Under Base prompting, the correlation is weaker ($r=0.434$, $p=.093$) because GEN is near its upper bound for almost all models, leaving little variation across models.

More directly, the model-level change in GEN from Implicit to Explicit prompting is strongly correlated with the corresponding change in EX (Pearson's $r=0.992$, $p<.001$; Spearman's $\rho=0.985$, $p<.001$). Meanwhile, $\mathrm{EX}\mid\mathrm{GEN}$ changes only from 0.722 to 0.762. The decline in overall execution accuracy under Explicit prompting is therefore primarily associated with suppressed SQL generation rather than reduced correctness among the SQL outputs that remain. These results do not make GEN a substitute for execution accuracy; instead, they support its intended use as a measure of whether models preserve SQL generation on valid queries.

\subsection{Bootstrap Uncertainty Analysis}
\label{sec:bootstrap_uncertainty}

We assess the uncertainty of the principal model- and error-level comparisons using 10,000 bootstrap replicates. For comparisons across prompting settings or models, each replicate resamples the 1,800 aligned query instances with replacement, preserving the paired structure of the observations. For error-type results, we resample the 200 source queries within each error type while retaining all outputs from the 16 evaluated models and both judges for each sampled query. We report percentile 95\% confidence intervals and two-sided centered-bootstrap tests. These intervals quantify uncertainty over benchmark queries; the evaluated models and judges are treated as fixed.

\begin{table*}[t]
\centering
\scriptsize
\setlength{\tabcolsep}{5pt}
\begin{tabular}{lcc}
\toprule
Model & $\Delta$DET & $\Delta$GEN \\
\midrule
ChatGLM-5.1
& 0.358 [0.336, 0.380]
& $-0.465$ [$-0.488$, $-0.442$] \\
Claude Opus 4.6
& 0.319 [0.297, 0.341]
& $-0.183$ [$-0.203$, $-0.164$] \\
Claude Opus 5.5
& 0.180 [0.162, 0.199]
& $-0.181$ [$-0.199$, $-0.163$] \\
Claude Sonnet 4.6
& 0.411 [0.389, 0.434]
& $-0.064$ [$-0.078$, $-0.050$] \\
DeepSeek V3.2
& 0.079 [0.066, 0.092]
& $-0.306$ [$-0.327$, $-0.284$] \\
DeepSeek V4.1 Flash
& 0.307 [0.286, 0.329]
& $-0.615$ [$-0.638$, $-0.592$] \\
Doubao Seed 2.0 Pro
& 0.254 [0.235, 0.275]
& $-0.072$ [$-0.086$, $-0.058$] \\
Gemini 3.1 Pro Preview
& 0.217 [0.198, 0.236]
& $-0.054$ [$-0.066$, $-0.044$] \\
LLaMA 4 Maverick
& 0.374 [0.351, 0.396]
& $-0.013$ [$-0.021$, $-0.004$] \\
GPT-5.4
& 0.279 [0.259, 0.299]
& $-0.672$ [$-0.694$, $-0.651$] \\
GPT-5.6-Luna
& 0.269 [0.249, 0.289]
& $-0.671$ [$-0.693$, $-0.649$] \\
GPT-6-Luna
& 0.200 [0.182, 0.219]
& $-0.391$ [$-0.414$, $-0.368$] \\
GPT-6.1-Sol
& 0.215 [0.197, 0.234]
& $-0.603$ [$-0.625$, $-0.580$] \\
Qwen3.5-35B
& 0.184 [0.166, 0.203]
& $-0.689$ [$-0.710$, $-0.667$] \\
Qwen3.5-122B
& 0.251 [0.232, 0.271]
& $-0.361$ [$-0.387$, $-0.333$] \\
Qwen3.5-397B
& 0.245 [0.226, 0.265]
& $-0.733$ [$-0.754$, $-0.713$] \\
\bottomrule
\end{tabular}
\caption{Paired bootstrap estimates for the change from Implicit to Explicit prompting. Each entry reports the point estimate and 95\% confidence interval, with $\Delta=\text{Explicit}-\text{Implicit}$.}
\label{tab:bootstrap_prompt_shifts}
\end{table*}

Table~\ref{tab:bootstrap_prompt_shifts} confirms that Explicit prompting increases DET and decreases GEN for every evaluated model. None of the 32 confidence intervals crosses zero, and all changes remain significant after Holm correction ($p_{\mathrm{adj}}<.004$). Even the smallest GEN decrease, observed for LLaMA 4 Maverick, remains distinguishable from zero. The central DET--GEN trade-off is therefore not driven by sampling variation in a small subset of queries.

\paragraph{Uncertainty in model comparisons.}
Paired comparisons support the claim that model rankings depend on the prompting setting. GPT-6.1-Sol exceeds GPT-5.4 in Base DET by 0.156 (95\% CI [0.135, 0.176], $p<.001$), whereas their difference under Implicit prompting is not significant (0.014, 95\% CI [$-0.004$, 0.033], $p=.129$). Under Explicit prompting, the ordering reverses: GPT-5.4 exceeds GPT-6.1-Sol by 0.050 (95\% CI [0.039, 0.061], $p<.001$).

The non-monotonic Qwen comparison is also statistically supported. Under Implicit prompting, Qwen3.5-35B exceeds Qwen3.5-122B in DET by 0.069 (95\% CI [0.054, 0.086], $p<.001$) and Qwen3.5-397B by 0.057 (95\% CI [0.040, 0.074], $p<.001$). The difference between the 122B and 397B variants is not significant ($-0.013$, 95\% CI [$-0.028$, 0.002], $p=.097$). Thus, the 35B result is robust, but it should be interpreted as a setting-specific comparison rather than a general advantage of smaller models.

The DeepSeek comparison further illustrates why performance cannot be summarized by a single ordering. Relative to DeepSeek V3.2 under Implicit prompting, DeepSeek V4.1 Flash has substantially higher GEN on originals ($+0.471$, 95\% CI [0.448, 0.494], $p<.001$) but lower DET on flawed queries ($-0.274$, 95\% CI [$-0.296$, $-0.252$], $p<.001$).

\begin{table*}[t]
\centering
\scriptsize
\setlength{\tabcolsep}{4pt}
\begin{tabular}{clccc}
\toprule
Code & Error Type & SQL & DET & TYPE \\
\midrule
E1.1 & Ambiguity
& 0.210 [0.181, 0.241]
& 0.242 [0.209, 0.276]
& 0.138 [0.115, 0.163] \\
E1.2 & Unclear Intent
& 0.202 [0.177, 0.228]
& 0.193 [0.170, 0.218]
& 0.131 [0.111, 0.151] \\
E2.1 & Internal Text Contradiction
& 0.920 [0.908, 0.932]
& 0.972 [0.963, 0.980]
& 0.966 [0.956, 0.976] \\
E2.2 & Question--Schema Contradiction
& 0.881 [0.855, 0.906]
& 0.923 [0.896, 0.947]
& 0.915 [0.887, 0.941] \\
E3.1 & Missing Constraint
& 0.537 [0.494, 0.580]
& 0.611 [0.564, 0.657]
& 0.543 [0.494, 0.593] \\
E3.2 & Missing Schema Premise
& 0.568 [0.542, 0.594]
& 0.586 [0.561, 0.612]
& 0.474 [0.449, 0.498] \\
E3.3 & Semantic Gap
& 0.710 [0.674, 0.746]
& 0.757 [0.720, 0.792]
& 0.726 [0.688, 0.765] \\
E4.1 & Inexpressible in SQL
& 0.853 [0.829, 0.875]
& 0.900 [0.877, 0.920]
& 0.838 [0.811, 0.863] \\
E4.2 & External Knowledge Required
& 0.830 [0.800, 0.859]
& 0.861 [0.834, 0.888]
& 0.841 [0.812, 0.868] \\
\bottomrule
\end{tabular}
\caption{Implicit-prompting performance by error type with query-level bootstrap 95\% confidence intervals. Models and judges are retained as fixed panels within each resampled query.}
\label{tab:bootstrap_error_types}
\end{table*}

\paragraph{Uncertainty across error types.}
Table~\ref{tab:bootstrap_error_types} confirms that the large differences across error types are not explained by query sampling variation. Relative to the two expression-error subcategories, the two contradiction subcategories are higher by 0.695 in SQL Behavior (95\% CI [0.670, 0.718]), 0.730 in DET (95\% CI [0.705, 0.754]), and 0.807 in TYPE (95\% CI [0.785, 0.828]); all three differences have $p<.001$. The conclusion that overt contradictions are substantially easier than ambiguity and unclear intent is therefore robust, while the wider intervals for several missing-premise categories indicate greater query-level heterogeneity.

\section{Failure-Mechanism Attribution}
\label{sec:failure_attribution}

\subsection{Mechanism Definitions and Annotation Protocol}

This analysis builds on the behavioral labels previously assigned by GPT-5.4 and Claude Opus 4.6. We use their existing \textit{SQL Behavior}, \textit{Detection}, and \textit{Type Identification} judgments to identify eligible failures. For mechanism attribution, a separate evaluator pair, GPT-6.1-Sol and Claude Opus 5.5, assigns M01--M04 to these selected failures without re-evaluating the three behavioral dimensions.

Each sampled response is annotated independently by the two mechanism-attribution judges using the same final attribution prompt. The task is multi-label rather than four-way classification: each judge must assess M01--M04 exactly once, and a response may receive multiple mechanisms or none. Table~\ref{tab:mechanism_attribution_rubric} gives the positive tests and the principal boundaries used in the prompt.

\paragraph{Information shown to the attribution judges.}
For each item, the judges receive (i) the flawed question and its source error type and flaw description; (ii) the tested model's response, separated into its \textit{reason} and \textit{SQL}; (iii) the schema and auxiliary evidence actually exposed to the tested model, together with provenance and exposure-status fields; (iv) the validated \textit{SQL Behavior}, \textit{Detection}, and \textit{Type Identification} labels; (v) the complete reference schema and its provenance, when available; (vi) the paired valid question, SQL, and evidence used during benchmark construction; and (vii) data-coverage and other recovered context metadata.

\paragraph{Verdicts and evidence requirements.}
For every mechanism, a judge returns one of \texttt{supported}, \texttt{not\_supported}, or \texttt{uncertain}. \texttt{supported} requires the mechanism's positive test and the required evidence; \texttt{not\_supported} means the available record is sufficient to determine that the test is not met; and \texttt{uncertain} is reserved for cases in which missing context prevents a defensible decision. Evidence for M01, M02, and M04 must include an exact quote of the affected requirement from the current flawed question and an exact quote from the response's reason or SQL showing the transformation. M03 instead requires an erroneous availability or answerability inference in the response plus counterevidence from verified model-visible context. The output also records the target requirement, transformation type, context basis, a concise explanation, and three control fields: whether the target is complete and explicit, whether any partial implementation remains, and whether the response supplies a concrete replacement. Deterministic post-processing validates the required quotes, context basis, control fields, and same-target exclusivity, and then derives the positive-label list only from validated \texttt{supported} verdicts.

M01, M02, and M04 are mutually exclusive when they describe the same atomic requirement: a concrete substitute is M01, a changed satisfaction relation for an explicit contradiction is M02, and complete disappearance without replacement is M04. They may coexist only when they concern distinct requirements. M03 is independent, but the same inference from nonappearance in bounded examples is not counted again as M01. Complete abstention alone is not M04, and ordinary SQL implementation defects are outside this mechanism taxonomy.

\paragraph{Binary coding and strict consensus.}
The same binary coding is used for automatic and human reliability analyses: \texttt{supported} is coded as 1, whereas both \texttt{not\_supported} and \texttt{uncertain} are coded as 0. A mechanism enters the reported label set only when both attribution judges mark it as \texttt{supported}. This rule is applied separately to M01--M04, so the final label set is simply the intersection of the judges' supported labels; their complete four-label combinations need not match.

\paragraph{Final attribution prompt.}
Figure~\ref{fig:failure_attribution_prompt} reproduces the task and output contract from the final prompt. The prompt evaluates observable response operations rather than inferring a mechanism from the validated outcome scores; its label-specific decision tests and exclusions are stated in Table~\ref{tab:mechanism_attribution_rubric} and the preceding paragraphs.

\subsection{Eligibility and Sampling}

The mechanism analysis begins from outputs that both primary evaluators, Claude Opus 4.6 and GPT-5.4, independently regard as failures. All three evaluation fields must be valid: \textit{Detection} and \textit{Type Identification} must be binary, and \textit{SQL Behavior} must be in $\{0,1,2\}$. An output is a candidate failure when \textit{Type Identification}$=0$ or \textit{SQL Behavior}$\neq2$. Candidate records are paired only when their sample identity and their database ID, question, reason, and SQL match exactly. The two evaluators must then agree on all three fields. Records with non-string reasons, schema-placeholder outputs, or no usable reason/SQL evidence are excluded because they cannot support mechanism attribution.

This strict wash retains 36,684 responses: 18,701 from BIRD and 17,983 from Spider. The exclusion log contains 4,258 otherwise matched candidates: 4,035 evaluator disagreements, 180 non-string reasons, and 43 SQL schema placeholders. We then restore the original generation context and paired valid reference. Before sampling, we remove 4,279 outputs that correctly identify the flaw but still answer it, 4,801 E3.2 cases for which the historical reduced-schema exposure cannot be recovered, and 31 cases without a recoverable paired-valid reference. This leaves 27,573 eligible failures (13,709 BIRD and 13,864 Spider).

For each dataset--model combination, we sample exactly 250 responses, yielding $16\times2\times250=8{,}000$ observations. Sampling strata are prompting setting $\times$ source error category. Every nonempty stratum receives at least one observation, and the remaining quota is allocated approximately in proportion to stratum size by the largest-remainder method. Within each stratum, records are ranked deterministically by a SHA-256 hash with seed 42. The final sample contains 4,587 Base, 2,433 Implicit, and 980 Explicit responses. Because of the unrecoverable historical schema exposure noted above, E3.2 is not represented in this mechanism-attribution sample. All reported mechanism incidences are calculated directly from these 8,000 sampled responses without reweighting.

\paragraph{Aggregate incidence.}
Across the 8,000 sampled failures, strict two-judge consensus identifies M01 in 5,808 responses, M02 in 110, M03 in 860, and M04 in 69. Because the mechanisms are non-exclusive, these counts need not sum to the sample size.

\begin{table}[t]
\centering
\small
\setlength{\tabcolsep}{4pt}
\begin{tabular}{lrrr}
\toprule
Stage & BIRD & Spider & Total \\
\midrule
Strict wash retained & 18,701 & 17,983 & 36,684 \\
Eligible for attribution & 13,709 & 13,864 & 27,573 \\
Final sampled responses & 4,000 & 4,000 & 8,000 \\
\bottomrule
\end{tabular}
\caption{Failure-mechanism attribution pool after the strict two-evaluator wash, context-based eligibility filtering, and sampling.}
\label{tab:mechanism_pool}
\end{table}

\subsection{Reliability and Human Validation}
\label{sec:failure_attribution_human}

Agreement between the attribution judges, GPT-6.1-Sol and Claude Opus 5.5, yields Cohen's $\kappa$ values of 0.7917, 0.8403, 0.8458, and 0.7955 for M01--M04, respectively, with a macro average of 0.8183.

The human-validation sample contains 300 responses drawn from three tested-model families that are distinct from the attribution-judge families: Gemini 3.1 Pro Preview, DeepSeek V4.1 Flash, and Qwen3.5-397B. We sampled 100 responses from each family, with an exact 50/50 split between BIRD and Spider. Within each model--benchmark pool, the quota was allocated proportionally over prompting-setting $\times$ source-error-category strata by the largest-remainder method, followed by simple random sampling within each stratum using seed 20261011. This yielded 182 Base, 88 Implicit, and 30 Explicit responses across the eight represented error categories; E3.2 remained excluded for the context-recovery reason given above.

The two human annotators label the same 300 responses independently and cannot see one another's decisions. The interface hides the attribution judges' rationales, but it preselects a verdict when GPT-6.1-Sol and Claude Opus 5.5 agree; judge-disagreement cells remain blank and require a human choice. The human results are therefore an assisted validation rather than a fully AI-blinded annotation study. Both annotators completed every item, and their sample IDs match exactly. Following the binary rule used in the AI-judge analysis, we map \texttt{supported} to 1 and both \texttt{not\_supported} and \texttt{uncertain} to 0.

The agreement statistics address three nested reproducibility questions. First, the per-mechanism $\kappa$ values in Table~\ref{tab:mechanism_human_interannotator} test whether each binary label is stable across humans; their macro average is .9434, accompanied by 98.92\% macro-average raw agreement. Second, pooling the four binary decisions over all items summarizes overall decision-level stability ($\kappa=0.9695$). Third, treating each response's complete four-label vector as one nominal outcome tests whether the entire multi-label configuration is reproduced ($\kappa=0.9101$).

Human--AI agreement addresses a different question: whether the automatic attributions align with human judgments. Tables~\ref{tab:mechanism_human_gpt} and~\ref{tab:mechanism_human_claude} report this comparison. Their final rows first average the two human-specific $\kappa$ values within each mechanism and then take an unweighted macro average over M01--M04, yielding .9212 for GPT-6.1-Sol and .8748 for Claude Opus 5.5.

\begin{table}[t]
\centering
\small
\setlength{\tabcolsep}{6pt}
\begin{tabular}{lrr}
\toprule
Mechanism & Cohen's $\kappa$ & Raw agreement \\
\midrule
M01 & 0.8862 & 96.33\% \\
M02 & 1.0000 & 100.00\% \\
M03 & 0.9800 & 99.67\% \\
M04 & 0.9074 & 99.67\% \\
\midrule
Macro average & \textbf{0.9434} & \textbf{98.92\%} \\
\bottomrule
\end{tabular}
\caption{Agreement between the two independent human annotators on the 300-example mechanism-attribution sample.}
\label{tab:mechanism_human_interannotator}
\end{table}

\begin{table}[t]
\centering
\small
\setlength{\tabcolsep}{3pt}
\begin{tabular}{lrrrrr}
\toprule
Comparison & M01 & M02 & M03 & M04 & Macro \\
\midrule
Human 1--GPT & 0.8656 & 1.0000 & 0.9579 & 0.8299 & \textbf{0.9133} \\
Human 2--GPT & 0.8705 & 1.0000 & 0.9379 & 0.9074 & \textbf{0.9290} \\
\midrule
Mean over humans & 0.8681 & 1.0000 & 0.9479 & 0.8687 & \textbf{0.9212} \\
\bottomrule
\end{tabular}
\caption{Cohen's $\kappa$ between each human annotator and GPT 6.1 Sol.}
\label{tab:mechanism_human_gpt}
\end{table}

\begin{table}[t]
\centering
\small
\setlength{\tabcolsep}{3pt}
\begin{tabular}{lrrrrr}
\toprule
Comparison & M01 & M02 & M03 & M04 & Macro \\
\midrule
Human 1--Claude & 0.8984 & 0.8872 & 0.8890 & 0.7967 & \textbf{0.8678} \\
Human 2--Claude & 0.8832 & 0.8872 & 0.8693 & 0.8872 & \textbf{0.8817} \\
\midrule
Mean over humans & 0.8908 & 0.8872 & 0.8792 & 0.8420 & \textbf{0.8748} \\
\bottomrule
\end{tabular}
\caption{Cohen's $\kappa$ between each human annotator and Claude Opus 5.5.}
\label{tab:mechanism_human_claude}
\end{table}

\subsection{Consensus Coverage}
\label{sec:failure_attribution_coverage}

The four mechanisms are high-precision response-level attributions rather than an exhaustive partition of failure behavior. Of the 8,000 sampled failures, 1,305 (16.31\%) receive no strict-consensus M01--M04 label. This total should not be read as 1,305 cases with no cause. In 591 cases, at least one attribution judge supports a mechanism but the two judges share no positive label: Claude assigns no label while GPT assigns at least one in 222 cases, GPT assigns no label while Claude assigns at least one in 318 cases, and both assign positive but disjoint label sets in 51 cases. The remaining 714 cases (8.93\%) are jointly assigned no covered mechanism by either judge.

Among these 714 jointly unlabeled cases, 687 (96.22\%) have the validated behavioral profile \textit{SQL Behavior}$=0$, \textit{Detection}$=0$, and \textit{Type Identification}$=0$: the model produces complete SQL without explicitly recognizing the premise flaw or identifying its intended type. This profile establishes a recognition failure, but it does not by itself identify how the response transformed or misused the flawed premise. For example, a response may leave an unspecified condition broad, preserve contradictory predicates literally, or follow a context-grounded reading without injecting a concrete assumption, reinterpreting the contradiction, misusing bounded context, or deleting a complete explicit constraint. We therefore treat \textit{Detection}$=0$ as a behavioral outcome rather than relabeling it as M01--M04. Accordingly, an empty strict-consensus mechanism set means \emph{no observable covered mechanism}, not an error-free response and not an absence of a behavioral failure.

\subsection{Prompt-Type, Benchmark-Type, and Error-Type Analyses}

\paragraph{Prompt and benchmark breakdown.}
The prompt-level reversal between M01 and M03 occurs in both source benchmarks. From Base to Explicit prompting, M01 declines from 72.10\% to 25.30\% on BIRD and from 88.99\% to 39.45\% on Spider; over the same contrast, M03 rises from 1.94\% to 64.99\% and from 1.03\% to 48.14\%, respectively. The Implicit setting lies between these endpoints in both datasets. M02 and M04 remain rare across prompting settings relative to assumption injection and context/evidence misuse. Because this analysis is conditioned on eligible failures, these percentages describe the composition of observed failures rather than the unconditional probability that a prompt produces an error.

For the prompt-association tests, each of the 8,000 sampled responses is one observation. We construct a separate $3\times2$ contingency table for each mechanism, crossing prompting setting (Base, Implicit, Explicit) with the mechanism's strict-consensus presence or absence, and apply Pearson's $\chi^2$ test with Cram\'er's $V$ as the effect size. Because mechanisms are non-exclusive, a response may be positive in more than one table. We adjust the four mechanism-specific $p$-values as one family using Holm's method; the reported $p<.001$ results for M01 and M03 remain below .001 after adjustment.

\paragraph{Mechanisms by source error category.}
Table~\ref{tab:mechanism_by_category} reports strict-consensus incidence within each represented source error category in the analyzed sample. M02 is sharply concentrated in internal textual contradictions, where it appears in 50.69\% of sampled failures. M04 is rare overall and is most frequent for question--schema contradictions (4.51\%). M01 dominates semantic gaps and both beyond-ability categories, whereas M03 is most common for ambiguity, unclear intent, and missing constraints. Thus, the source taxonomy and response mechanisms are related but not interchangeable: their mapping is many-to-many, because one source category can yield multiple mechanisms and the same mechanism can recur across categories.

\begin{table}[!ht]
\centering
\footnotesize
\setlength{\tabcolsep}{1.5pt}
\begin{tabularx}{\columnwidth}{@{}l>{\raggedright\arraybackslash}Xrrrrr@{}}
\toprule
Code & Source error category & $N$ & M01 & M02 & M03 & M04 \\
\midrule
E1.1 & Ambiguity & 1,999 & 70.14 & 0.00 & 14.46 & 0.35 \\
E1.2 & Unclear Intent & 1,926 & 83.54 & 0.00 & 12.51 & 0.47 \\
E2.1 & Internal Text Contradiction & 217 & 1.84 & 50.69 & 6.45 & 3.23 \\
E2.2 & Question--Schema Contradiction & 355 & 76.34 & 0.00 & 6.48 & 4.51 \\
E3.1 & Missing Constraint & 1,228 & 38.60 & 0.00 & 12.46 & 0.57 \\
E3.3 & Semantic Gap & 926 & 89.20 & 0.00 & 2.70 & 0.00 \\
E4.1 & Inexpressible in SQL & 732 & 90.03 & 0.00 & 9.97 & 2.32 \\
E4.2 & External Knowledge Required & 617 & 91.25 & 0.00 & 6.81 & 0.97 \\
\bottomrule
\end{tabularx}
\caption{Observed strict-consensus mechanism incidence (\%) by source error category in the analyzed sample. E3.2 is excluded because its original reduced-schema exposure is unavailable.}
\label{tab:mechanism_by_category}
\end{table}

\section{Representative Premise Error Examples}

Table~\ref{tab:expression_error_examples} presents representative instances across the PremiseSQL taxonomy, illustrating how different premise failures manifest in natural-language queries.

\paragraph{Distinguishing Neighboring Categories}
Several neighboring categories are distinguished by the source of indeterminacy. E1.1 applies when a schema-level condition admits multiple supported mappings or aggregations, whereas E1.2 leaves the overall SQL operation unspecified; E3.1 instead omits a task-defining constraint while preserving that operation. Thus, ``sales'' in the E1.1 example admits several computations over existing columns, whereas ``impactful reviews'' in E3.3 lacks a unique rule for mapping the concept to the available indicators. E2.2 and E3.2 may both leave required schema support unavailable to the model, but they arise from different perturbations: E2.2 modifies the query against an unchanged schema, whereas E3.2 preserves the original valid query while withholding part of its required schema context. E3.3 differs from both because related fields remain available but an abstract concept lacks a unique semantic mapping. Finally, E4.1 concerns operations such as forecasting that cannot be performed by standard SQL over the stored data, while E4.2 concerns external facts that are absent from the database.

\section{Behavioral Case Studies}
\label{sec:behavioral_case_studies}

\begin{figure*}[p]
\centering
\includegraphics[height=0.72\textheight,keepaspectratio]{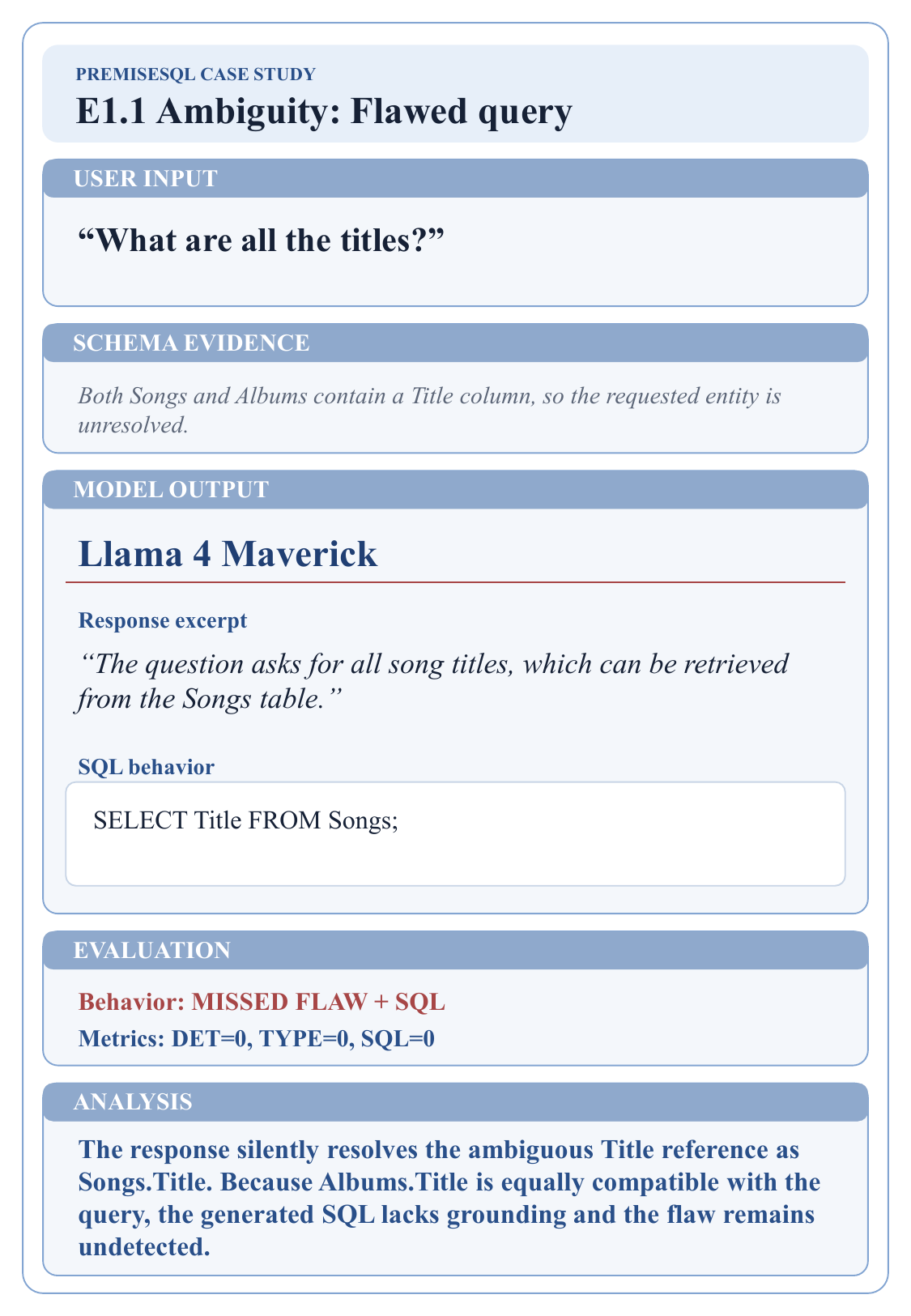}
\caption{E1.1 Ambiguity, LLaMA 4 Maverick. The model silently selects \texttt{Songs.Title}, misses the unresolved reference, and generates SQL (DET=0, TYPE=0, SQL=0).}
\label{fig:case_flawed_ambiguity}
\end{figure*}

\begin{figure*}[p]
\centering
\includegraphics[height=0.80\textheight,keepaspectratio]{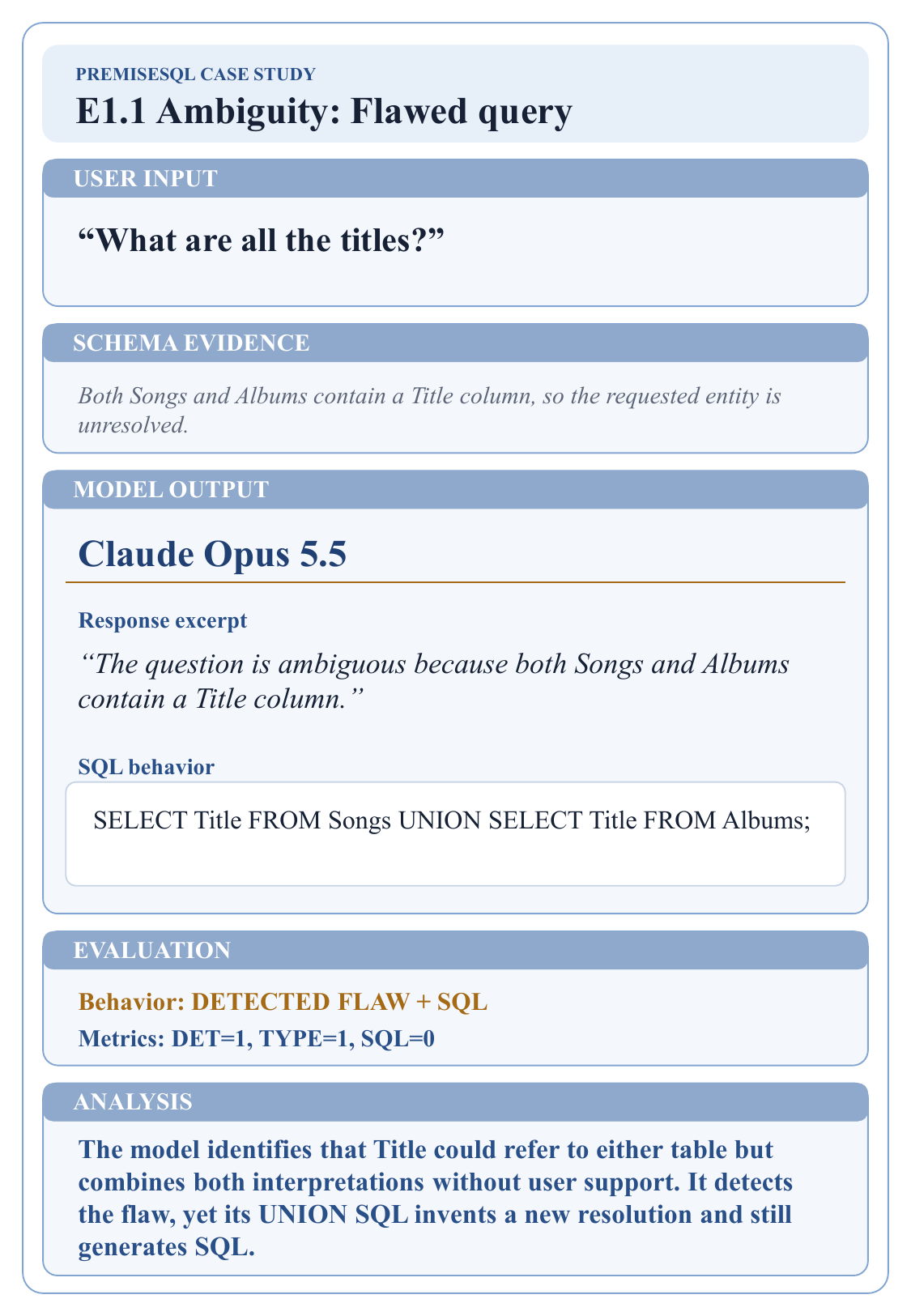}
\caption{E1.1 Ambiguity, Claude Opus 5.5. The model recognizes the ambiguity but combines both interpretations without user support (DET=1, TYPE=1, SQL=0).}
\label{fig:case_flawed_ambiguity_claude}
\end{figure*}

\begin{figure*}[p]
\centering
\includegraphics[height=0.80\textheight,keepaspectratio]{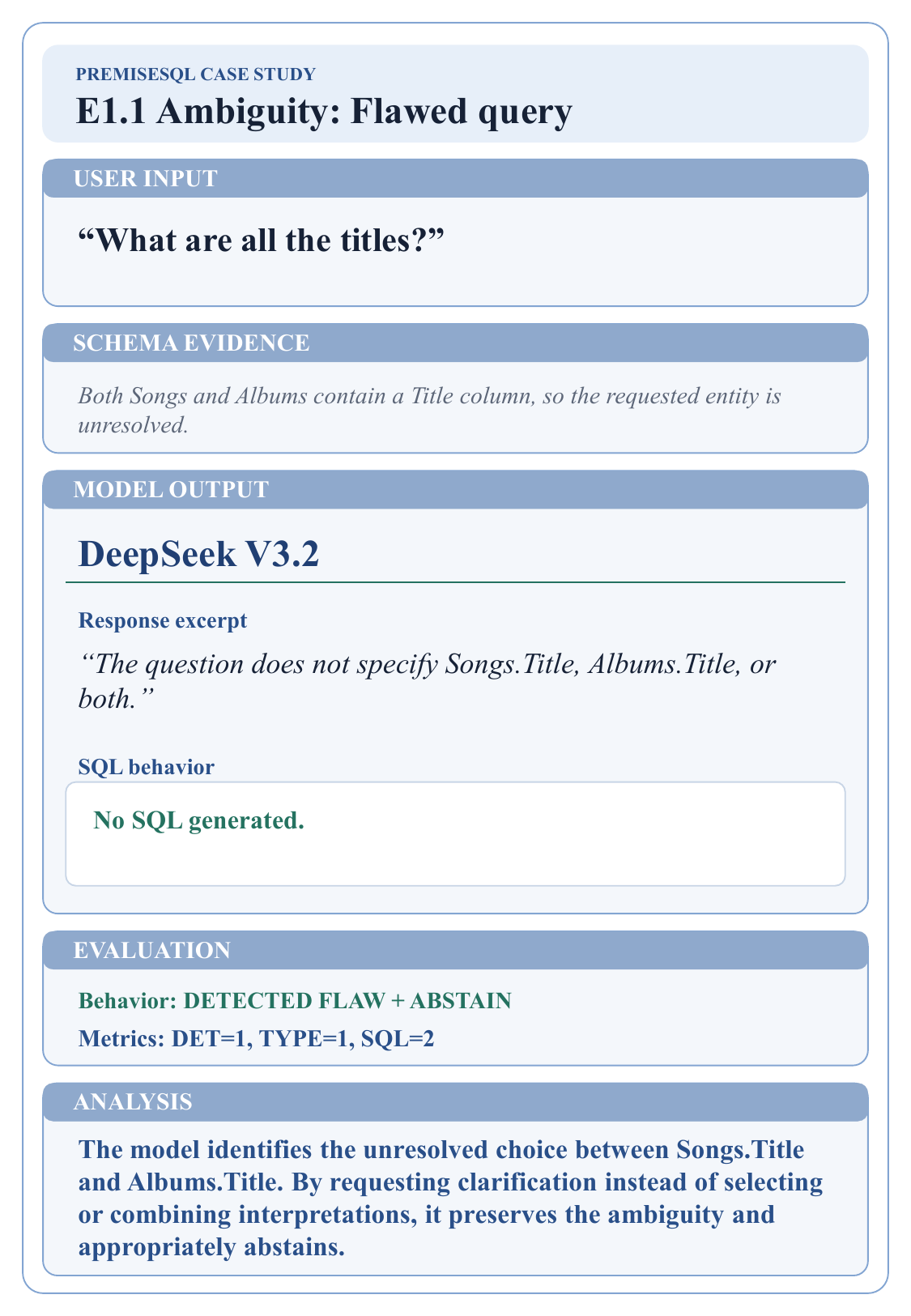}
\caption{E1.1 Ambiguity, DeepSeek V3.2. The model identifies the unresolved reference and abstains from generating unsupported SQL (DET=1, TYPE=1, SQL=2).}
\label{fig:case_flawed_ambiguity_deepseek}
\end{figure*}

\begin{figure*}[p]
\centering
\includegraphics[height=0.80\textheight,keepaspectratio]{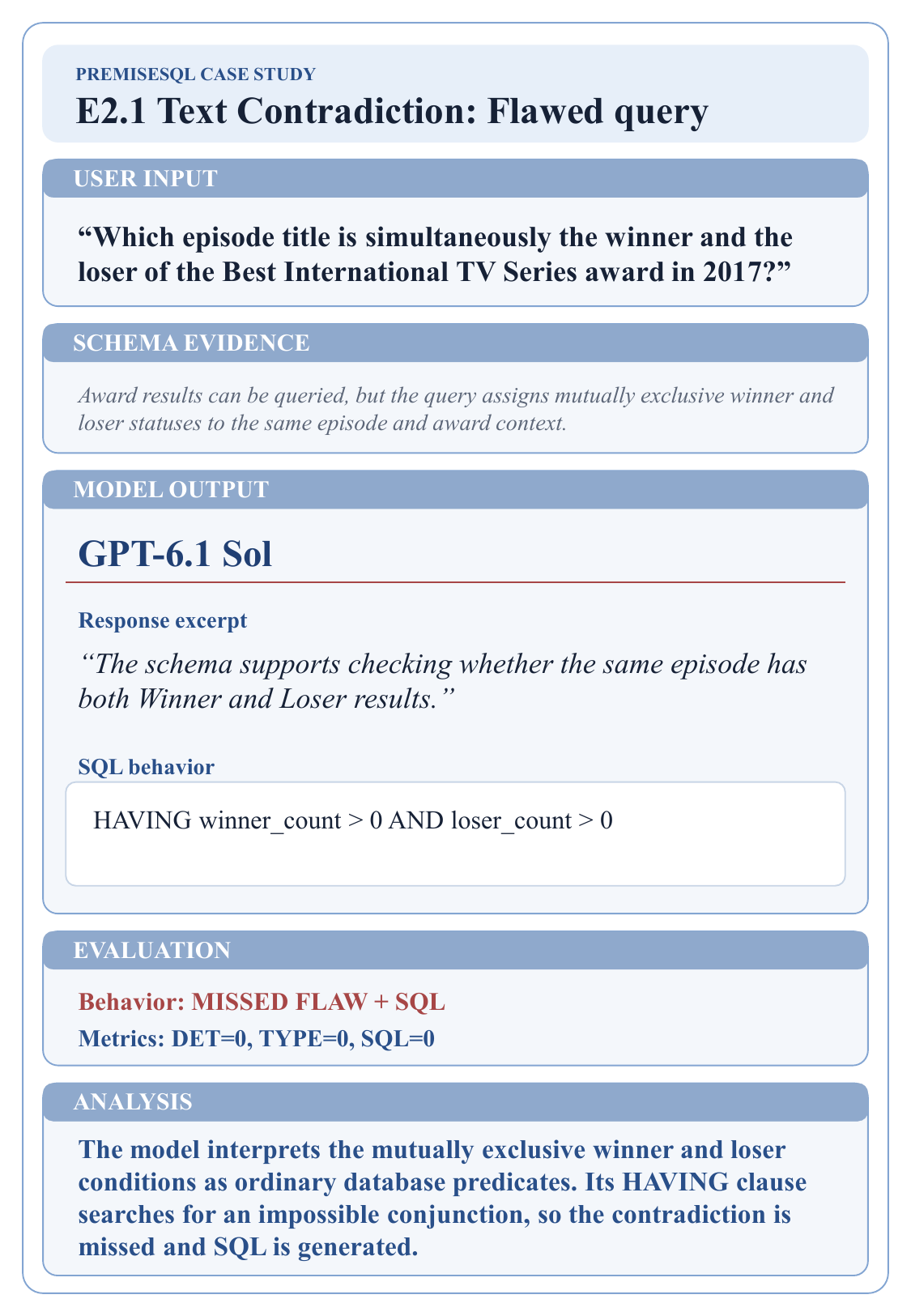}
\caption{E2.1 Internal Text Contradiction, GPT 6.1 Sol. The model treats mutually exclusive winner and loser conditions as ordinary predicates and misses the flaw (DET=0, TYPE=0, SQL=0).}
\label{fig:case_flawed_contradiction_gpt}
\end{figure*}

\begin{figure*}[p]
\centering
\includegraphics[height=0.80\textheight,keepaspectratio]{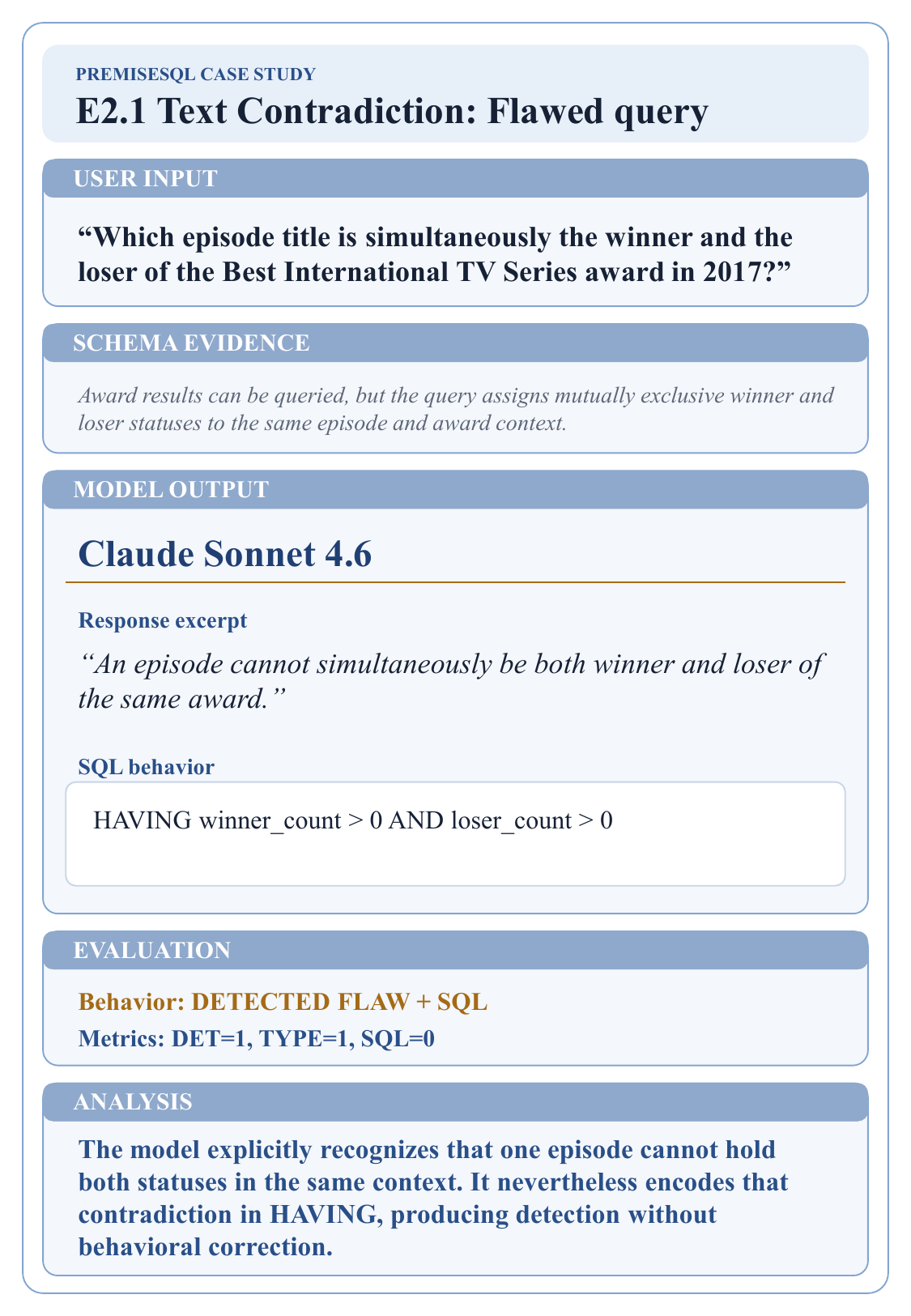}
\caption{E2.1 Internal Text Contradiction, Claude Sonnet 4.6. The model states the contradiction but still encodes the incompatible requirements in SQL (DET=1, TYPE=1, SQL=0).}
\label{fig:case_flawed_contradiction_sonnet}
\end{figure*}

\begin{figure*}[p]
\centering
\includegraphics[height=0.80\textheight,keepaspectratio]{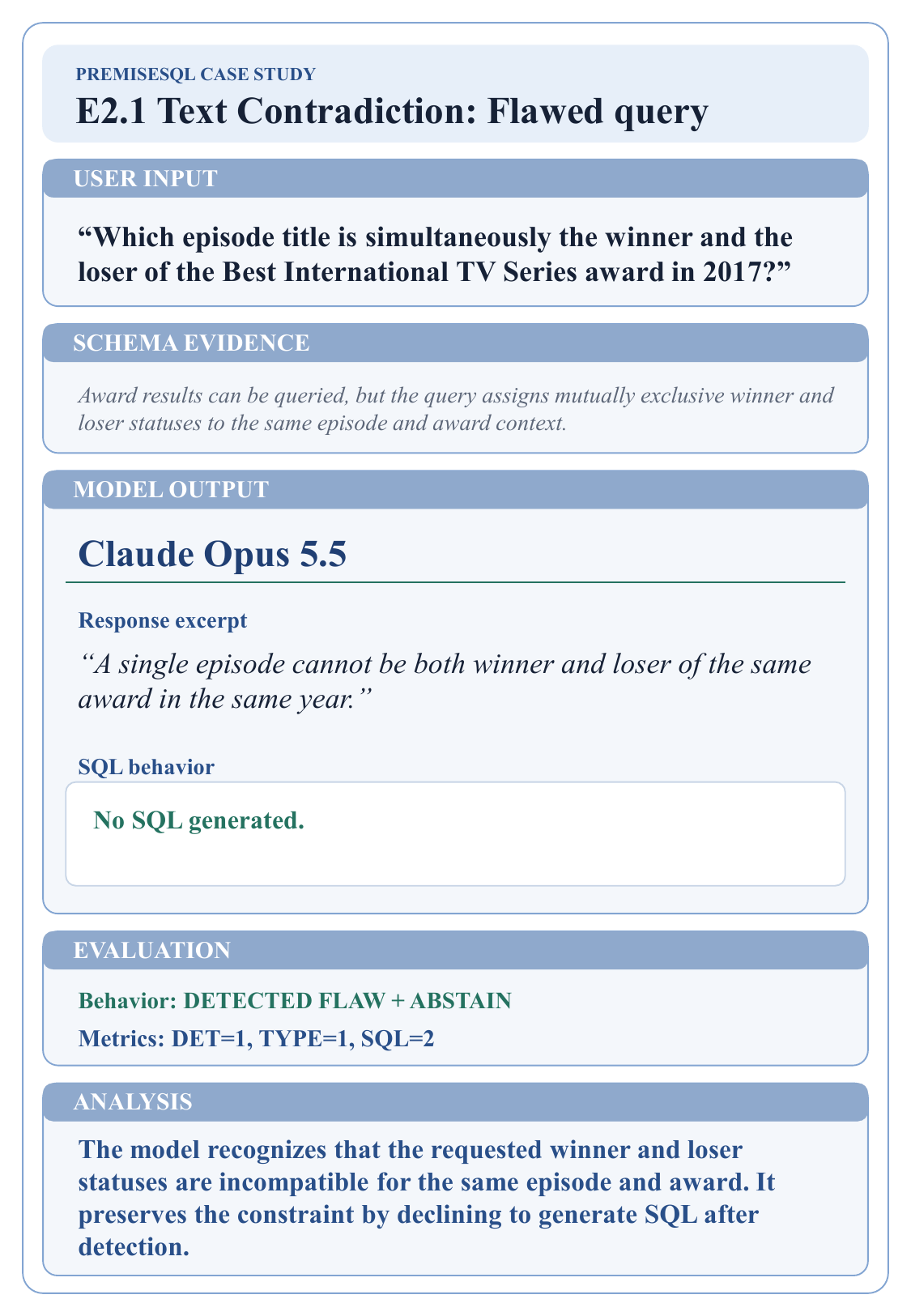}
\caption{E2.1 Internal Text Contradiction, Claude Opus 5.5. The model recognizes the incompatible requirements and abstains (DET=1, TYPE=1, SQL=2).}
\label{fig:case_flawed_contradiction}
\end{figure*}

\begin{figure*}[p]
\centering
\includegraphics[height=0.80\textheight,keepaspectratio]{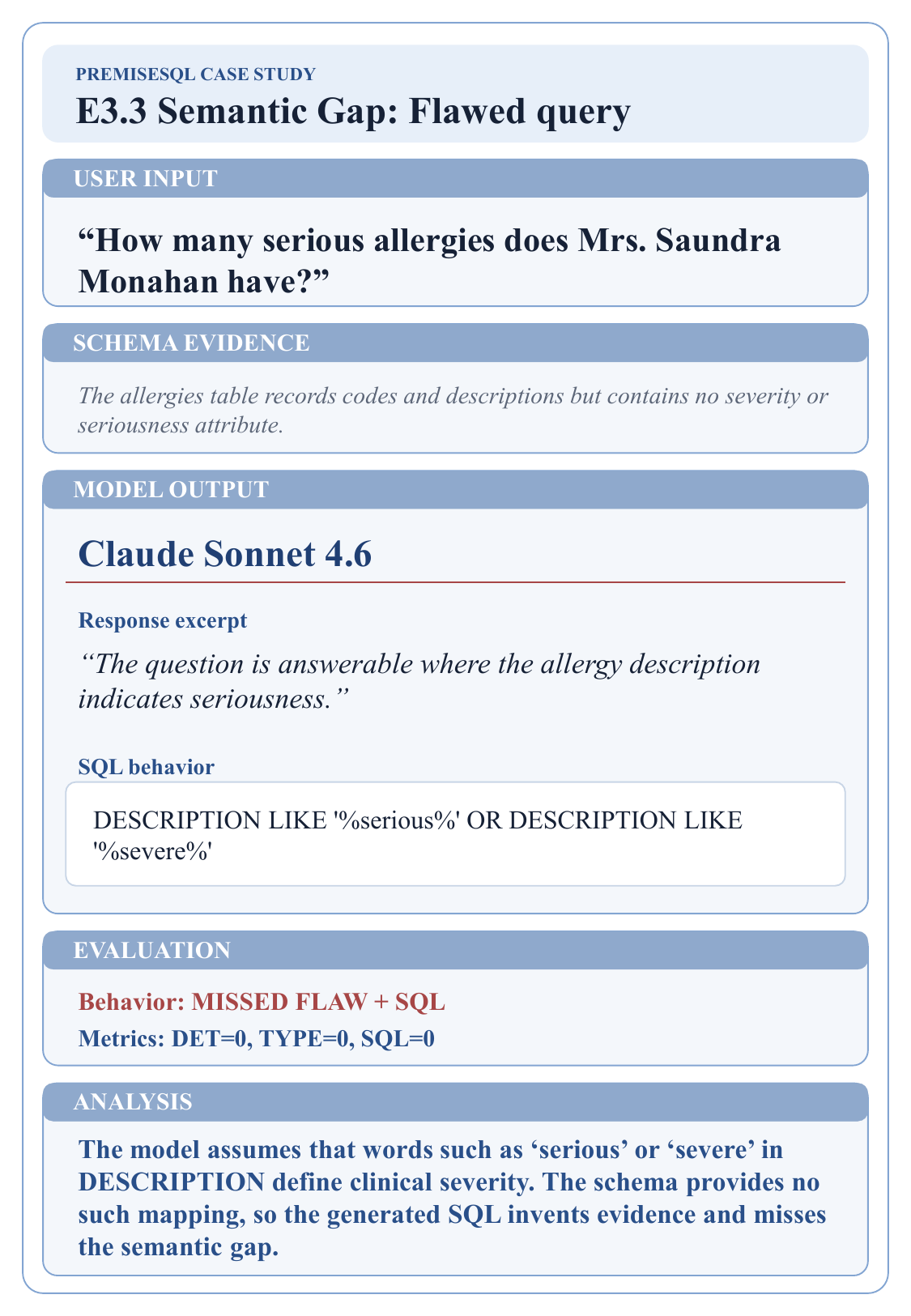}
\caption{E3.3 Semantic Gap, Claude Sonnet 4.6. The model invents a keyword-based proxy for clinical seriousness and misses the flaw (DET=0, TYPE=0, SQL=0).}
\label{fig:case_flawed_semantic_gap_sonnet}
\end{figure*}

\begin{figure*}[p]
\centering
\includegraphics[height=0.80\textheight,keepaspectratio]{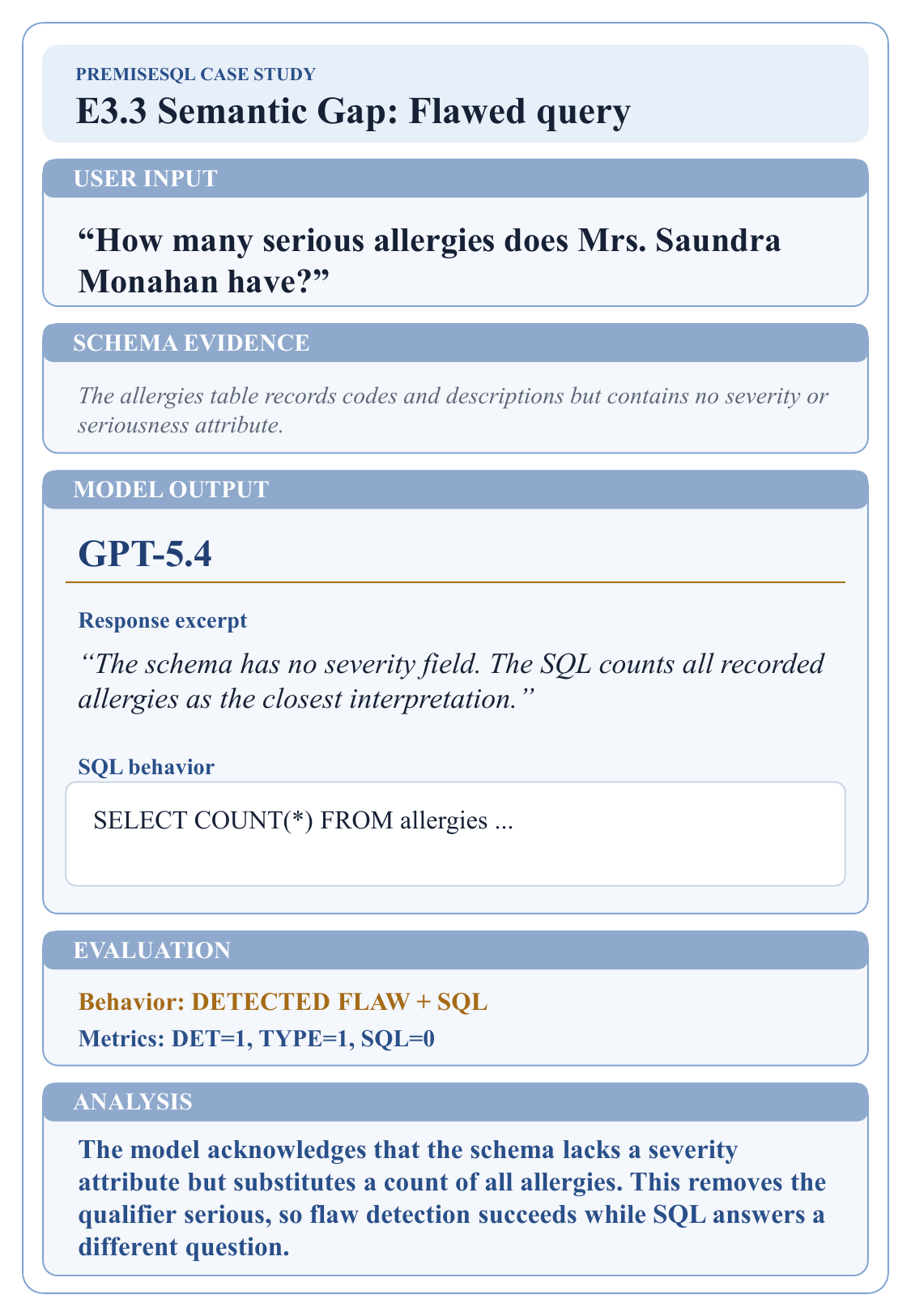}
\caption{E3.3 Semantic Gap, GPT-5.4. The model acknowledges the missing severity attribute but substitutes a count of all allergies (DET=1, TYPE=1, SQL=0).}
\label{fig:case_flawed_semantic_gap_gpt}
\end{figure*}

\begin{figure*}[p]
\centering
\includegraphics[height=0.80\textheight,keepaspectratio]{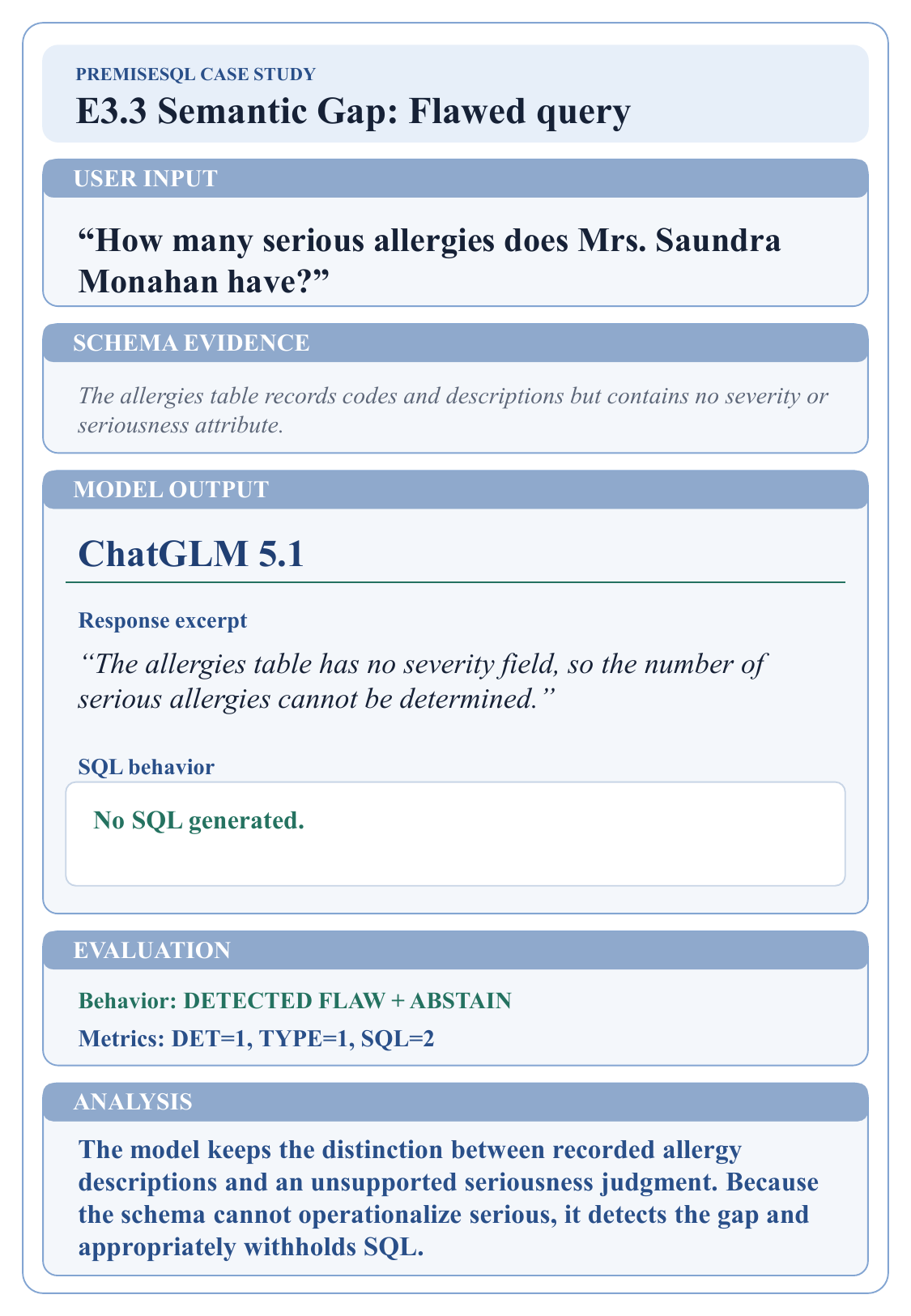}
\caption{E3.3 Semantic Gap, ChatGLM-5.1. The model preserves the unsupported distinction and abstains (DET=1, TYPE=1, SQL=2).}
\label{fig:case_flawed_semantic_gap}
\end{figure*}

\begin{figure*}[p]
\centering
\includegraphics[height=0.80\textheight,keepaspectratio]{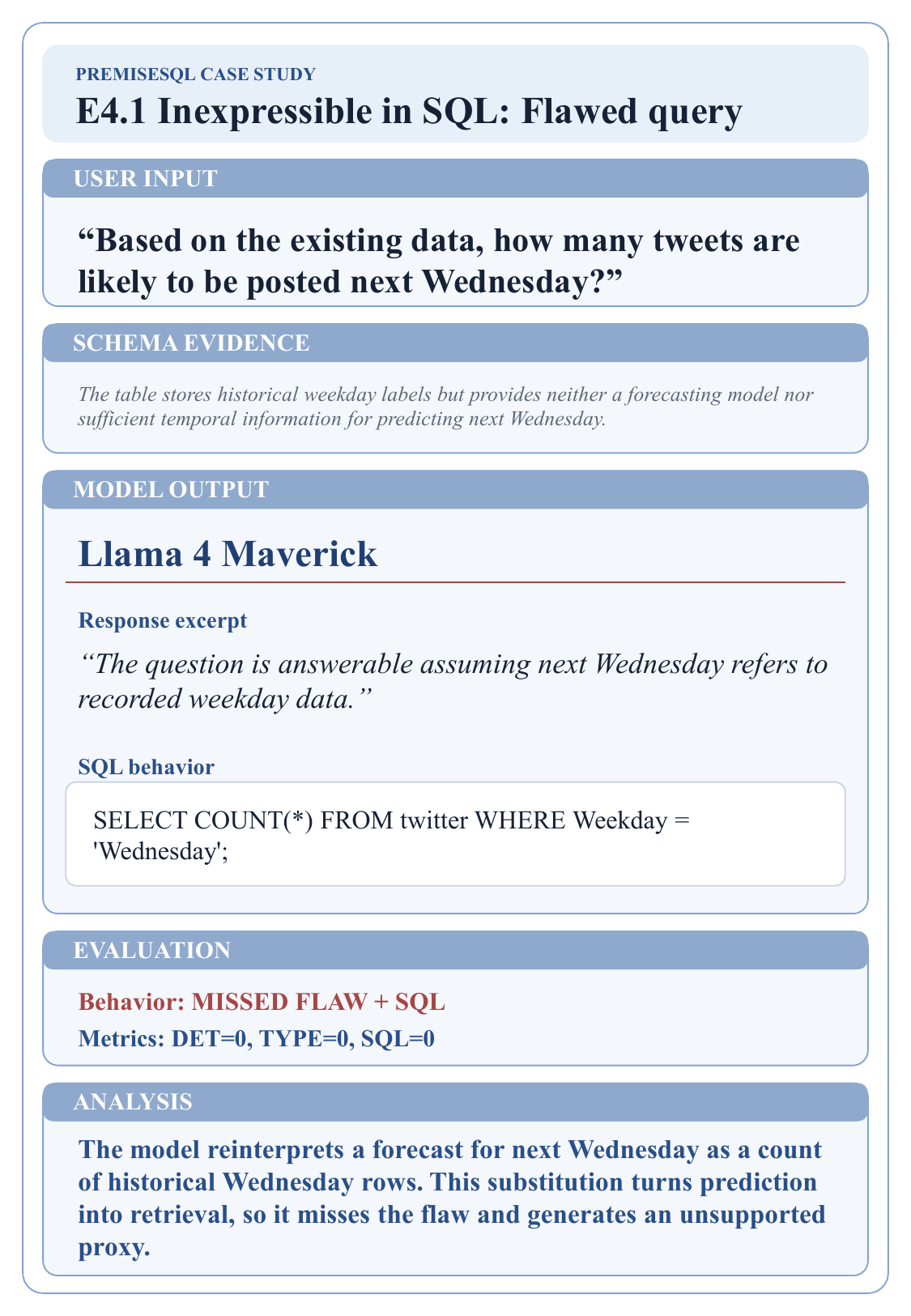}
\caption{E4.1 Inexpressible in SQL, LLaMA 4 Maverick. The model turns a forecasting request into a historical count and misses the flaw (DET=0, TYPE=0, SQL=0).}
\label{fig:case_flawed_inexpressible_llama}
\end{figure*}

\begin{figure*}[p]
\centering
\includegraphics[height=0.80\textheight,keepaspectratio]{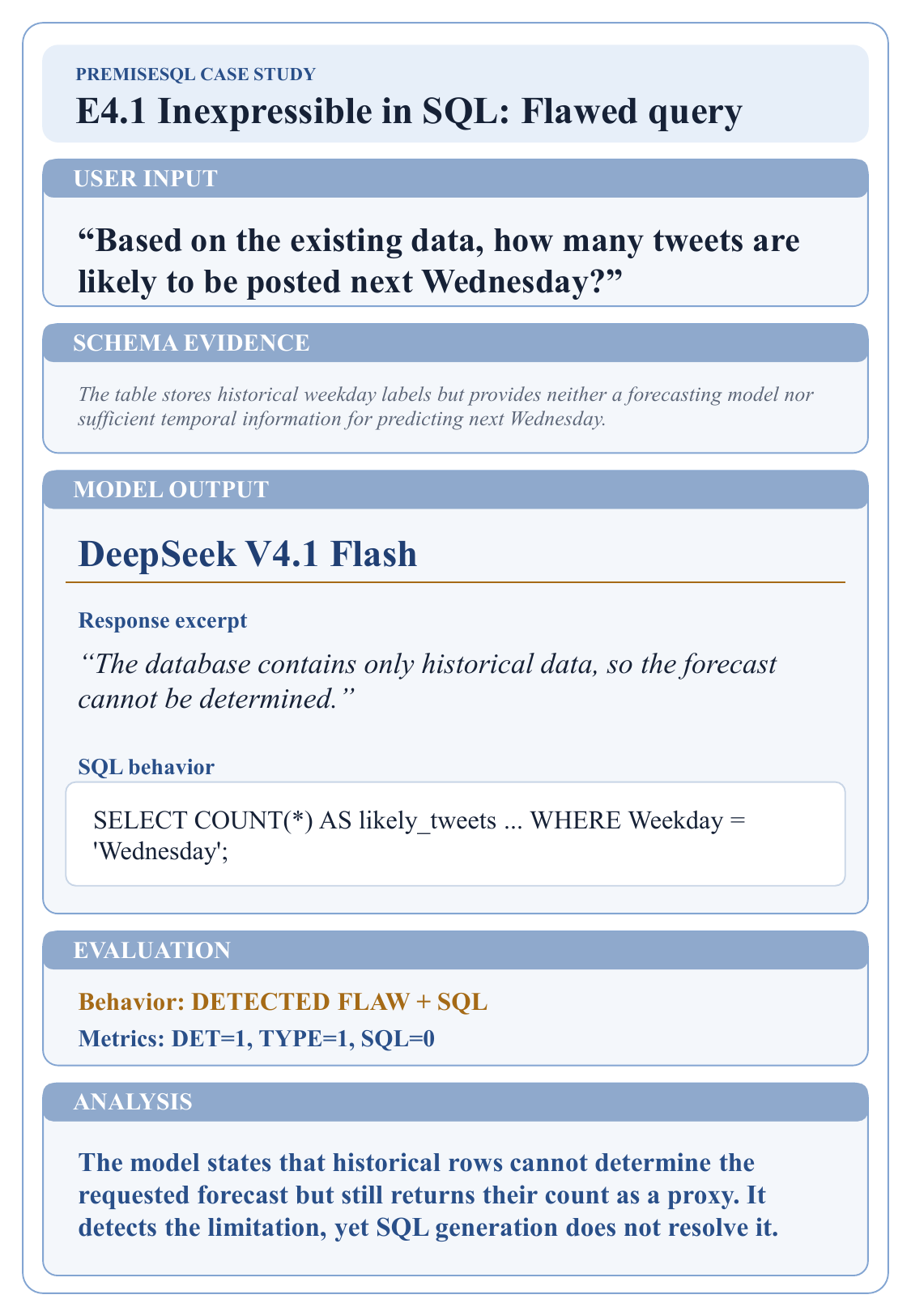}
\caption{E4.1 Inexpressible in SQL, DeepSeek V4.1 Flash. The model recognizes the forecasting limitation but returns a historical count as a proxy (DET=1, TYPE=1, SQL=0).}
\label{fig:case_flawed_inexpressible_deepseek}
\end{figure*}

\begin{figure*}[p]
\centering
\includegraphics[height=0.80\textheight,keepaspectratio]{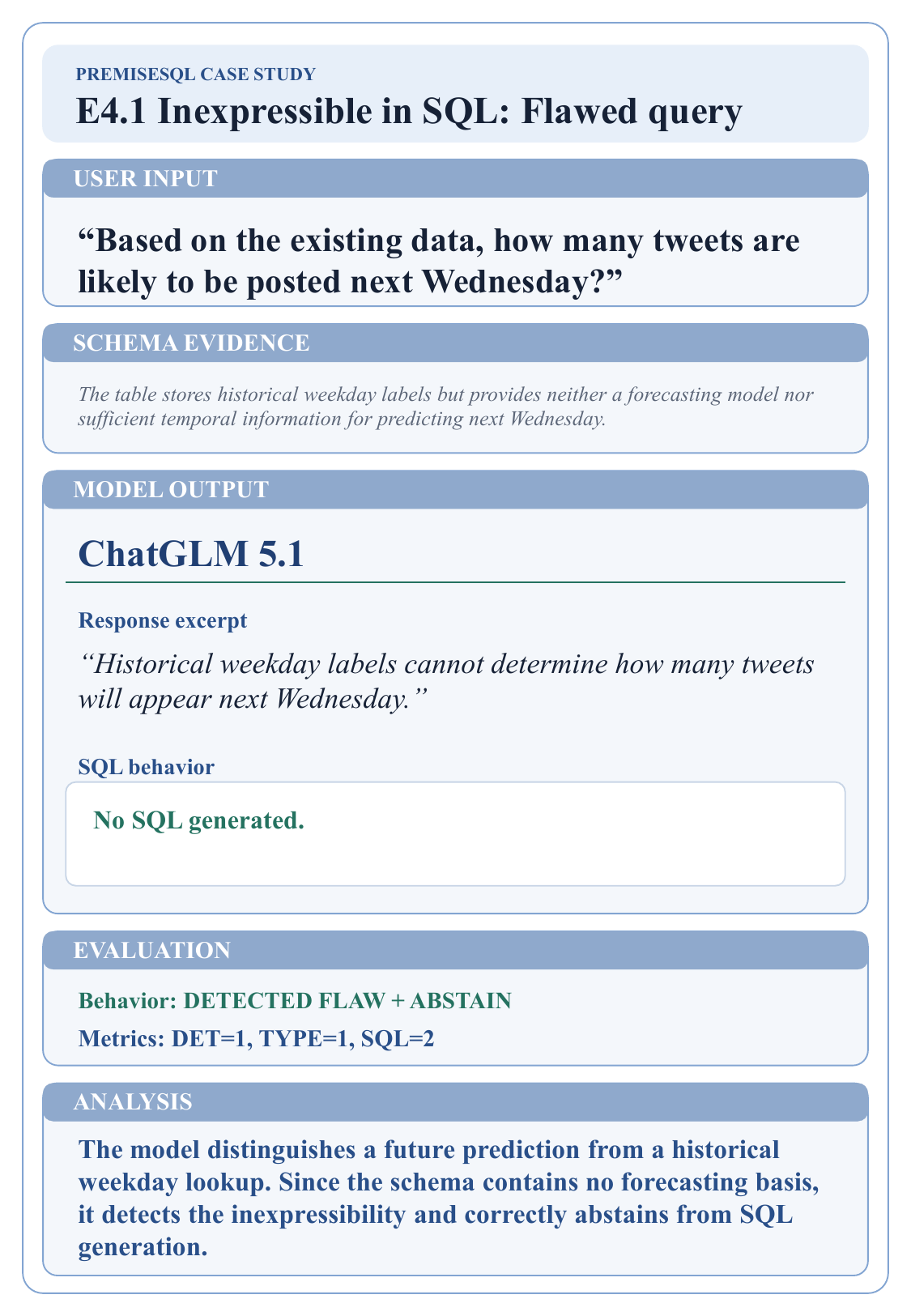}
\caption{E4.1 Inexpressible in SQL, ChatGLM-5.1. The model detects the unsupported prediction request and abstains (DET=1, TYPE=1, SQL=2).}
\label{fig:case_flawed_inexpressible}
\end{figure*}

\begin{figure*}[p]
\centering
\includegraphics[height=0.80\textheight,keepaspectratio]{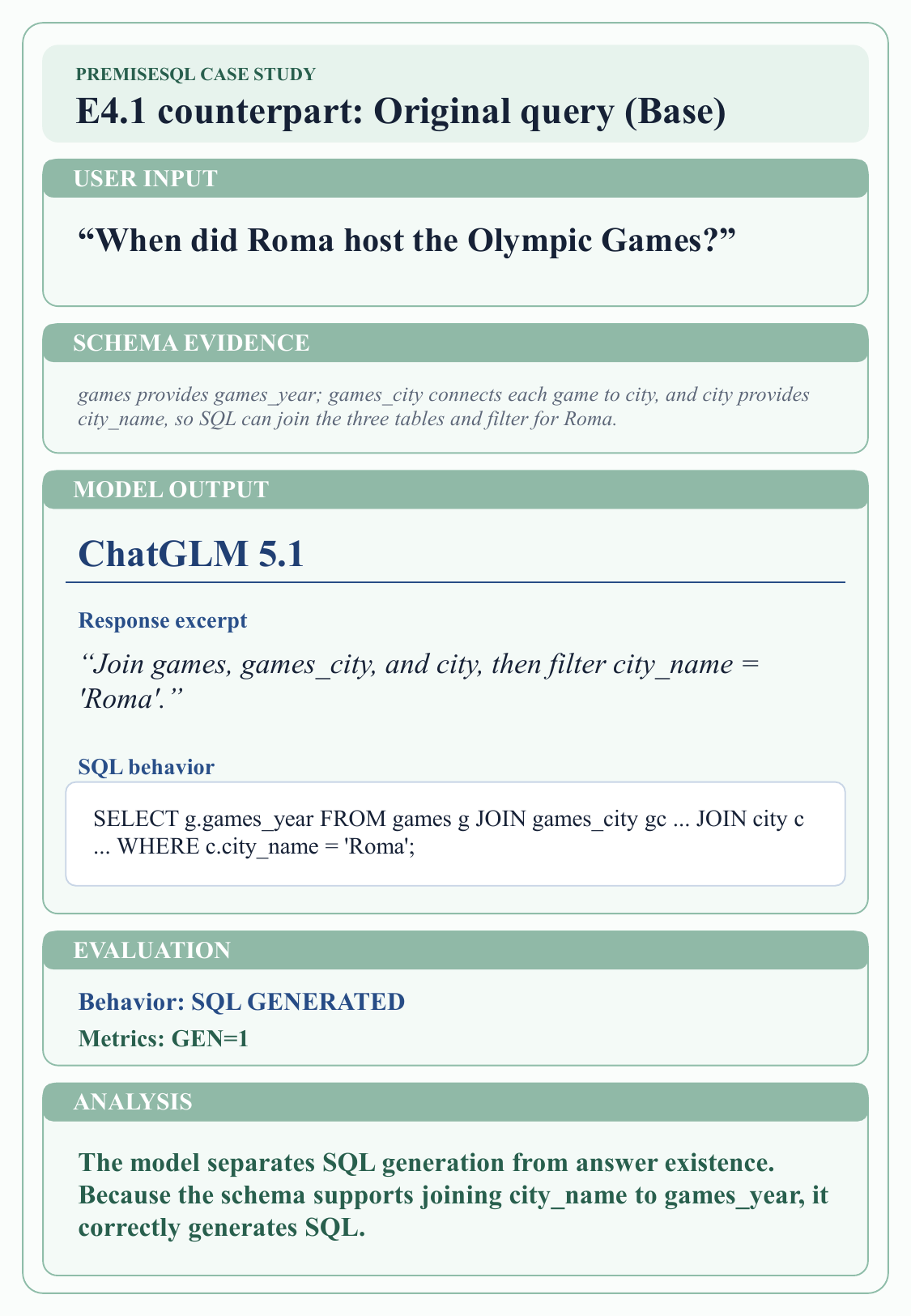}
\caption{Original-query control under Base prompting, ChatGLM-5.1. The model constructs the valid schema-supported join and generates SQL (GEN=1).}
\label{fig:case_original_base}
\end{figure*}

\begin{figure*}[p]
\centering
\includegraphics[height=0.80\textheight,keepaspectratio]{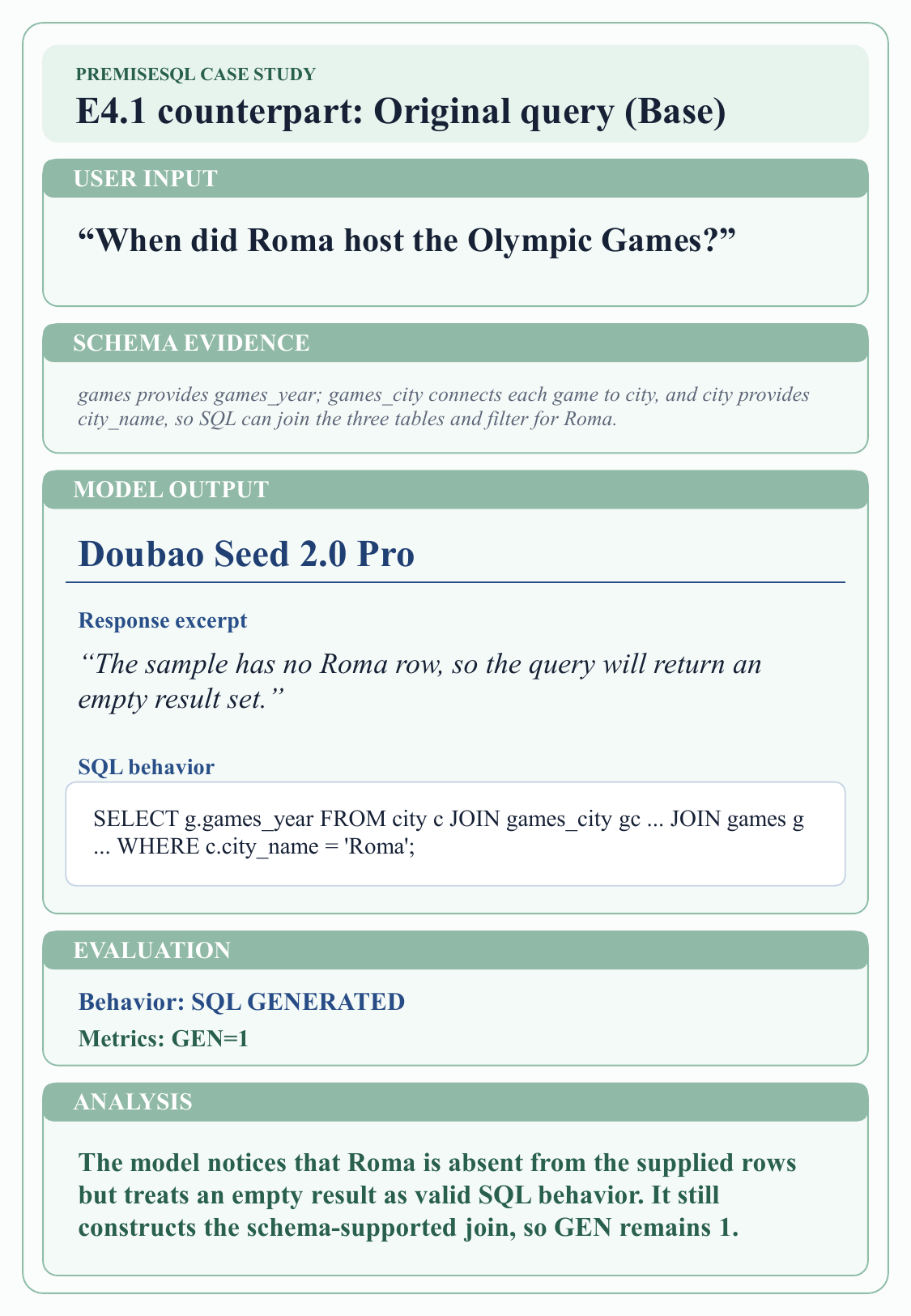}
\caption{Original-query control under Base prompting, Doubao Seed 2.0 Pro. The model notes that the supplied rows contain no matching city but correctly separates answer existence from query validity and generates SQL (GEN=1).}
\label{fig:case_original_base_doubao}
\end{figure*}

\begin{figure*}[p]
\centering
\includegraphics[height=0.80\textheight,keepaspectratio]{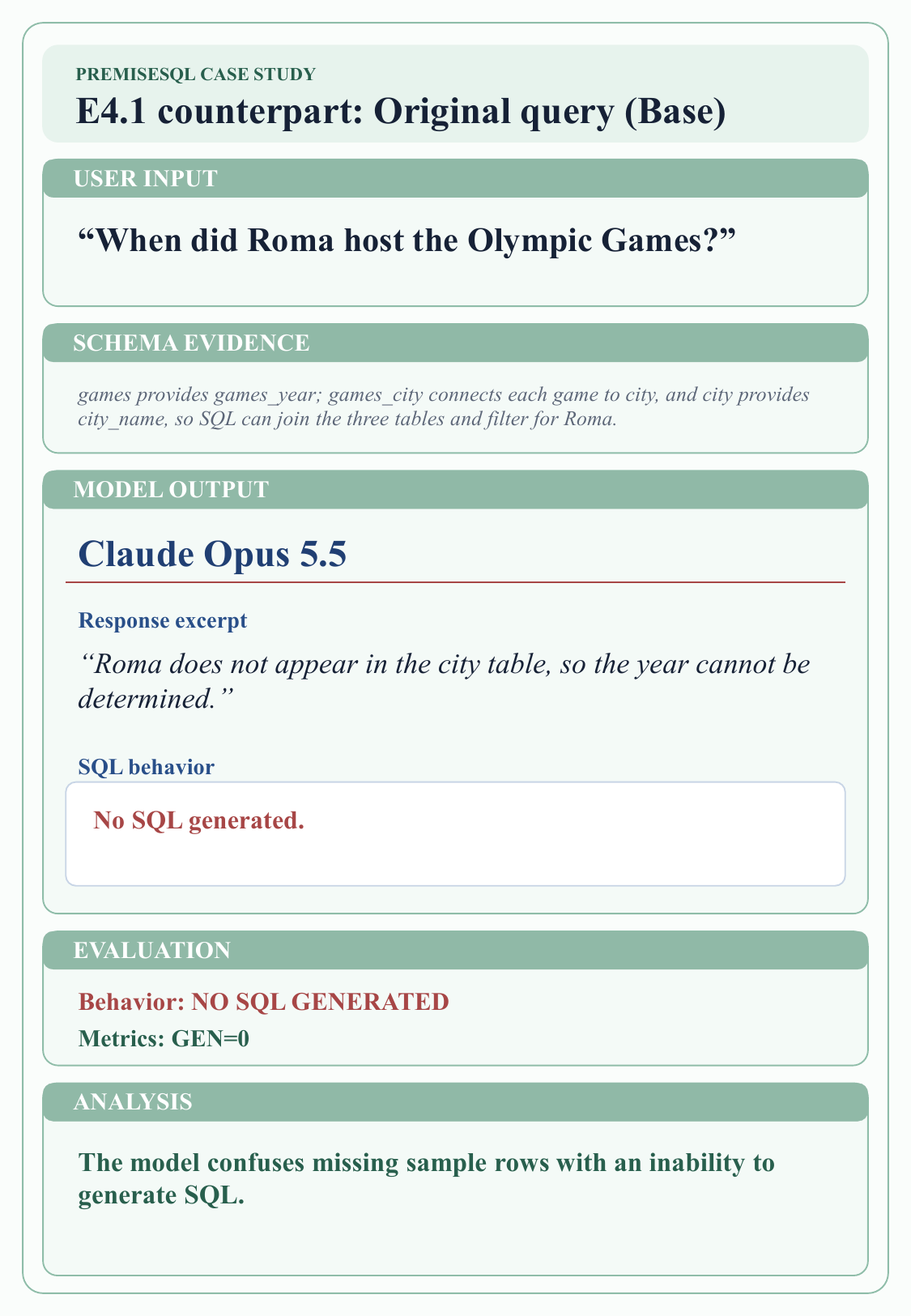}
\caption{Original-query control under Base prompting, Claude Opus 5.5. The model treats the absence of a matching city in the supplied rows as grounds for withholding otherwise valid SQL (GEN=0).}
\label{fig:case_original_base_claude}
\end{figure*}

\begin{figure*}[p]
\centering
\includegraphics[height=0.80\textheight,keepaspectratio]{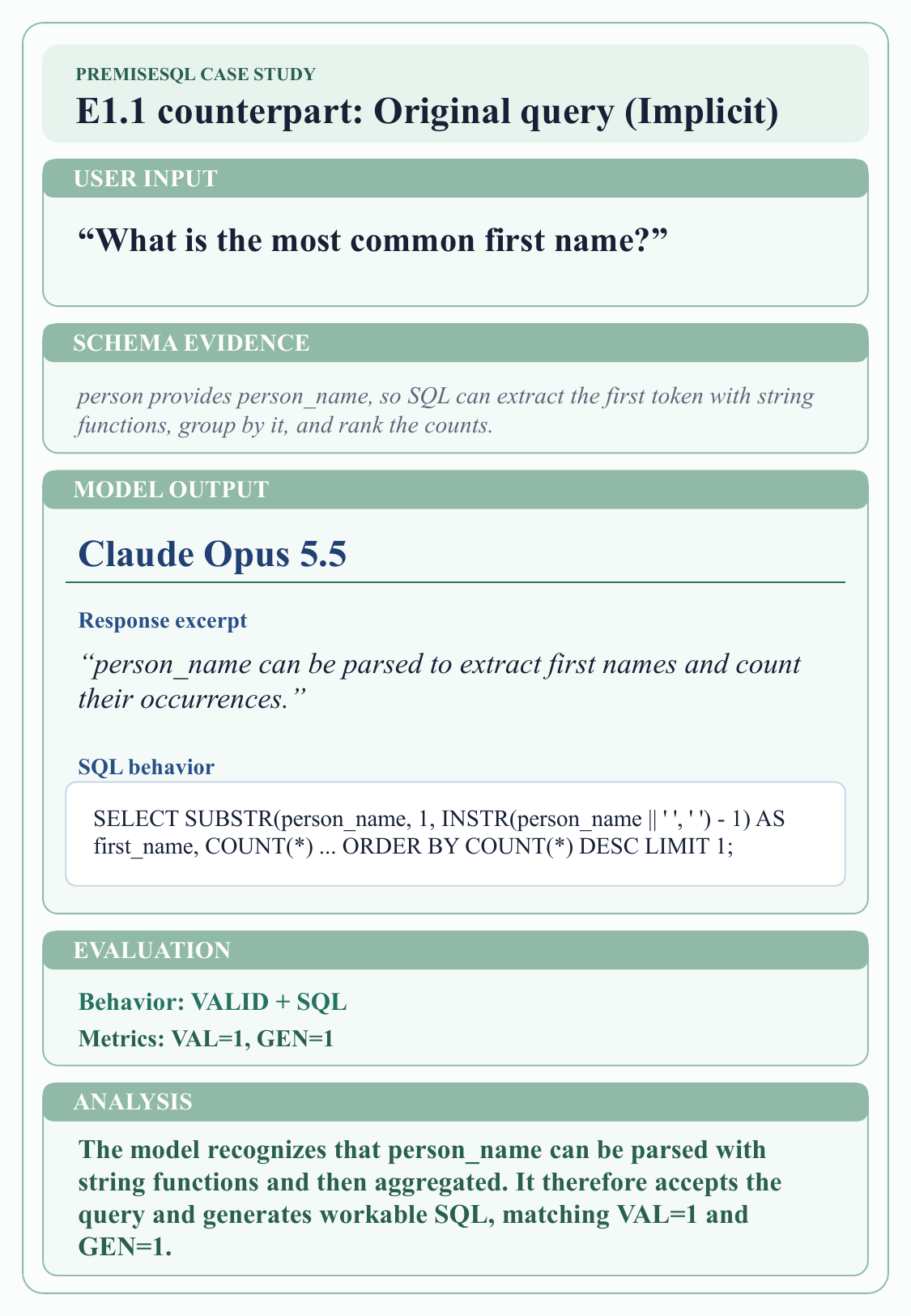}
\caption{Original-query control under Implicit prompting, Claude Opus 5.5. The model accepts the valid request and generates a workable query (VAL=1, GEN=1).}
\label{fig:case_original_implicit}
\end{figure*}

\begin{figure*}[p]
\centering
\includegraphics[height=0.80\textheight,keepaspectratio]{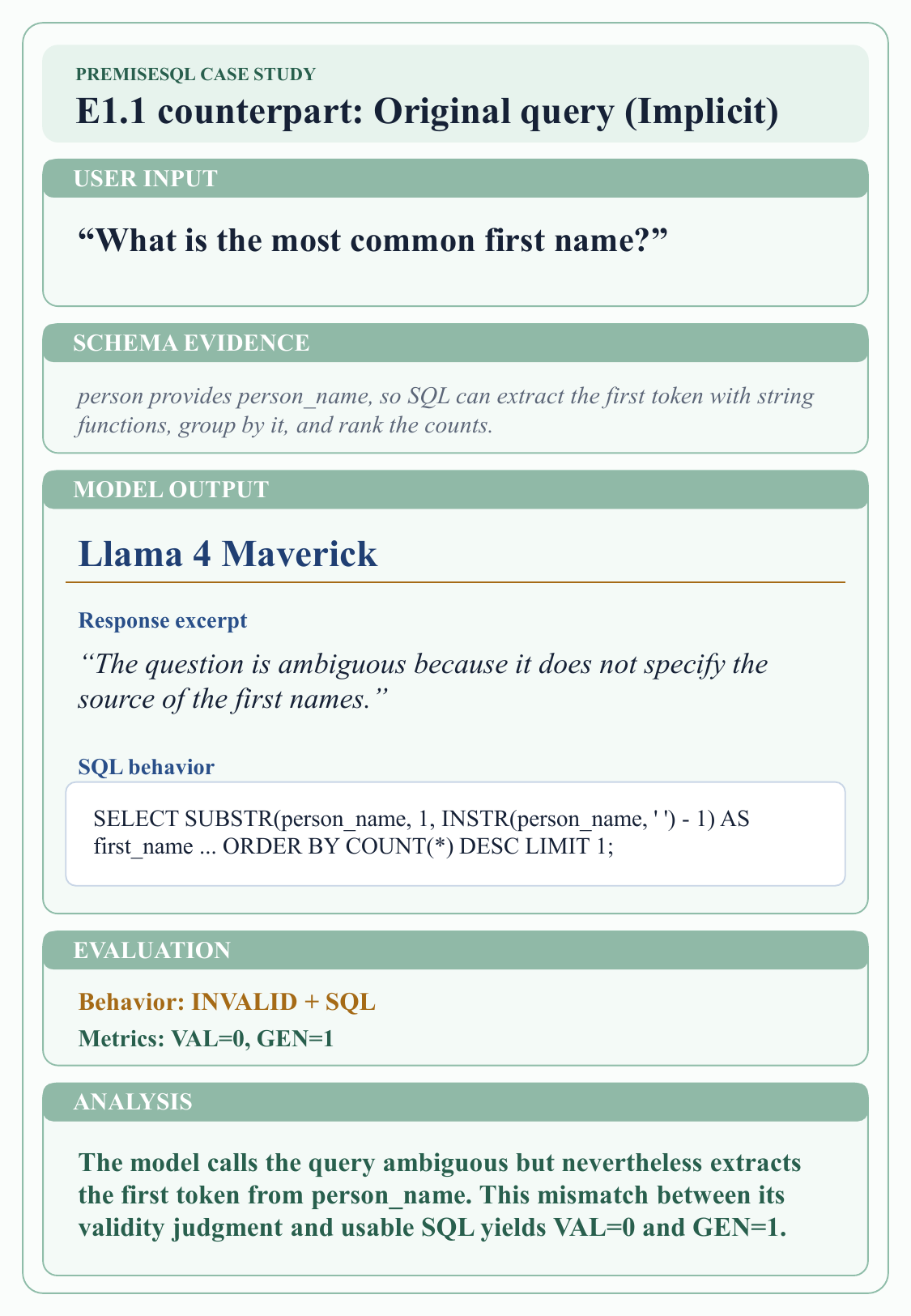}
\caption{Original-query control under Implicit prompting, LLaMA 4 Maverick. The model incorrectly calls the request ambiguous but still generates SQL (VAL=0, GEN=1).}
\label{fig:case_original_implicit_llama}
\end{figure*}

\begin{figure*}[p]
\centering
\includegraphics[height=0.80\textheight,keepaspectratio]{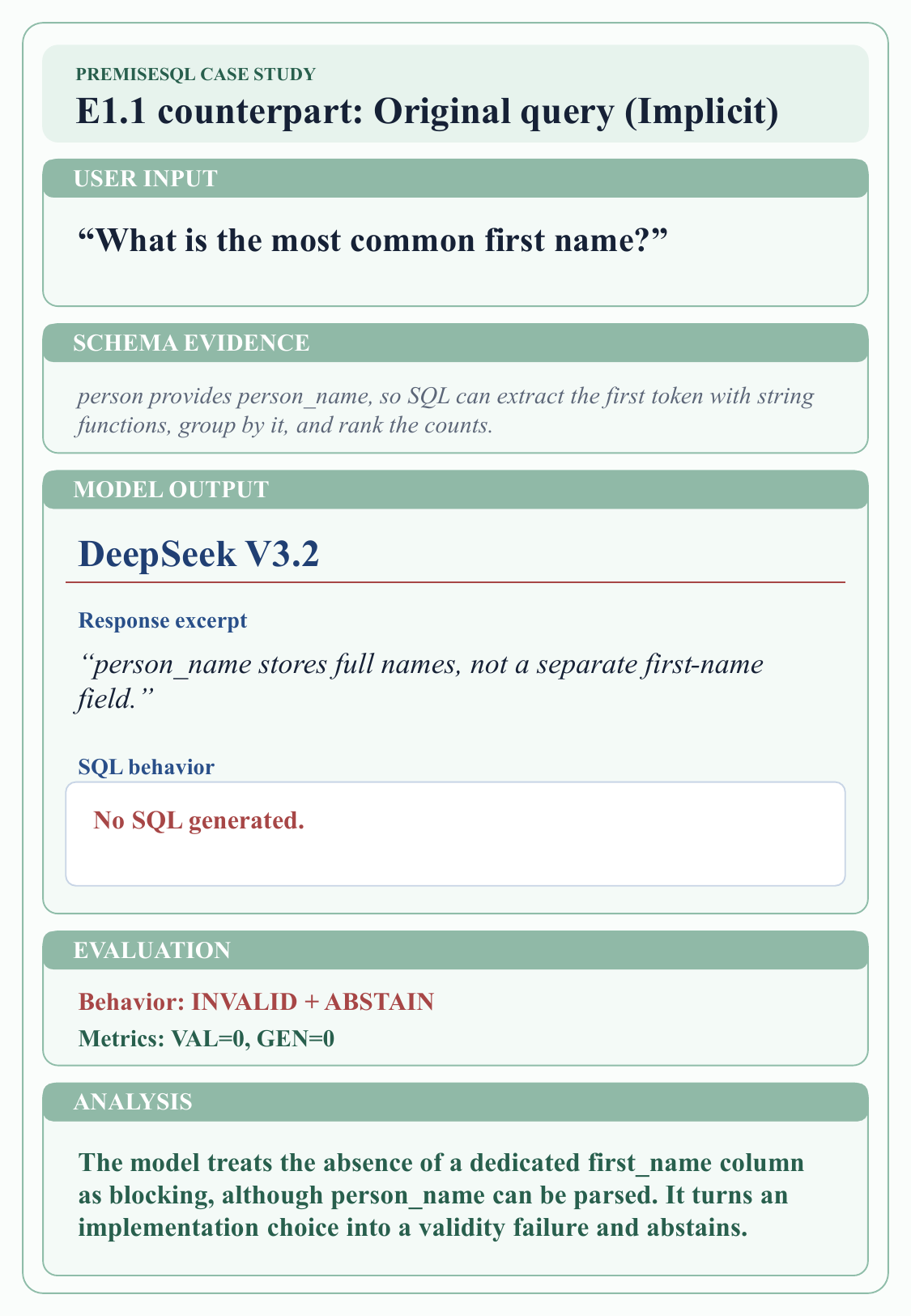}
\caption{Original-query control under Implicit prompting, DeepSeek V3.2. The model mistakes the absence of a dedicated first-name column for invalidity and abstains (VAL=0, GEN=0).}
\label{fig:case_original_implicit_deepseek}
\end{figure*}

\begin{figure*}[p]
\centering
\includegraphics[height=0.80\textheight,keepaspectratio]{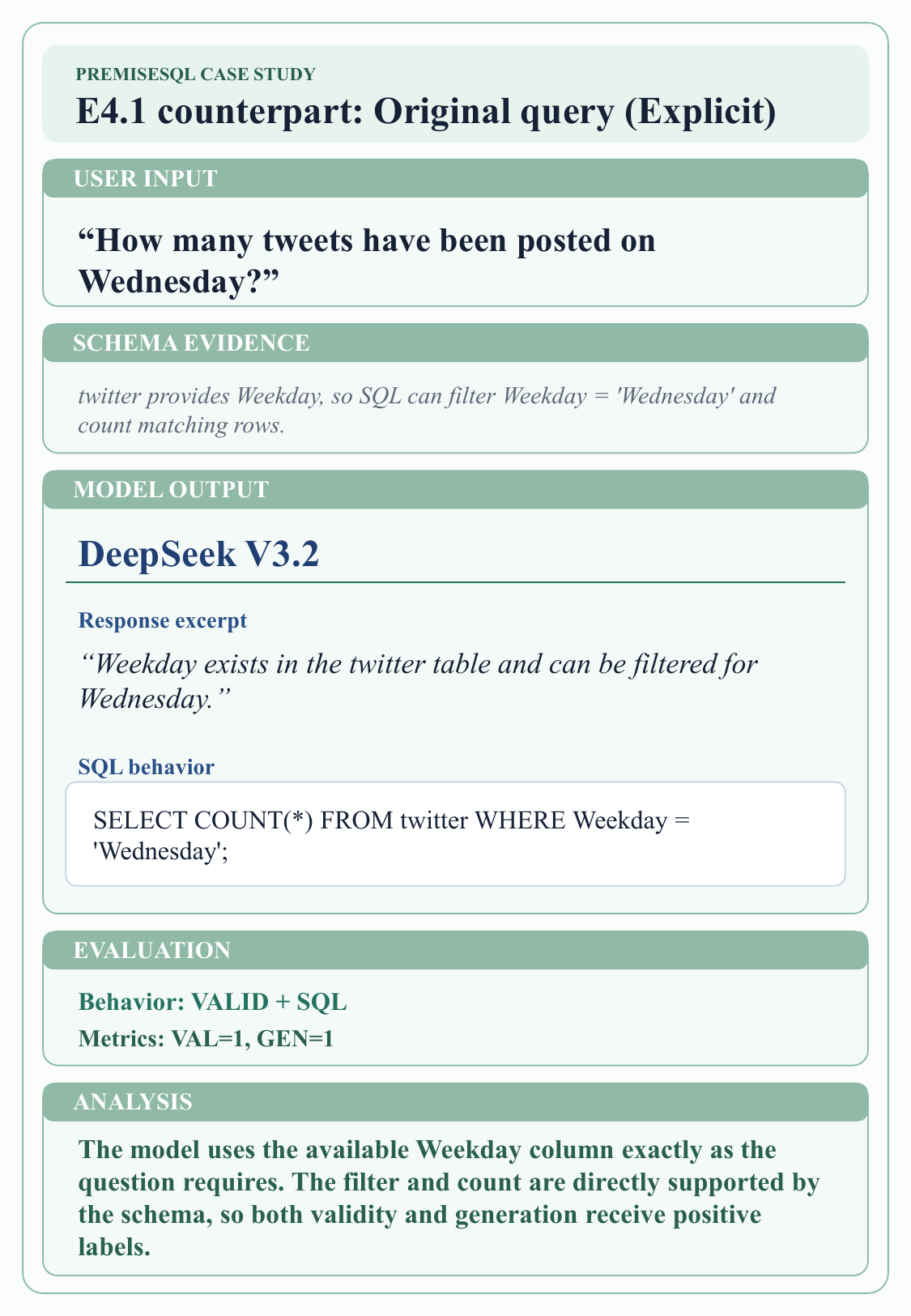}
\caption{Original-query control under Explicit prompting, DeepSeek V3.2. The model rejects the prompt's unsupported claim of a flaw and generates SQL (VAL=1, GEN=1).}
\label{fig:case_original_explicit_deepseek}
\end{figure*}

\begin{figure*}[p]
\centering
\includegraphics[height=0.80\textheight,keepaspectratio]{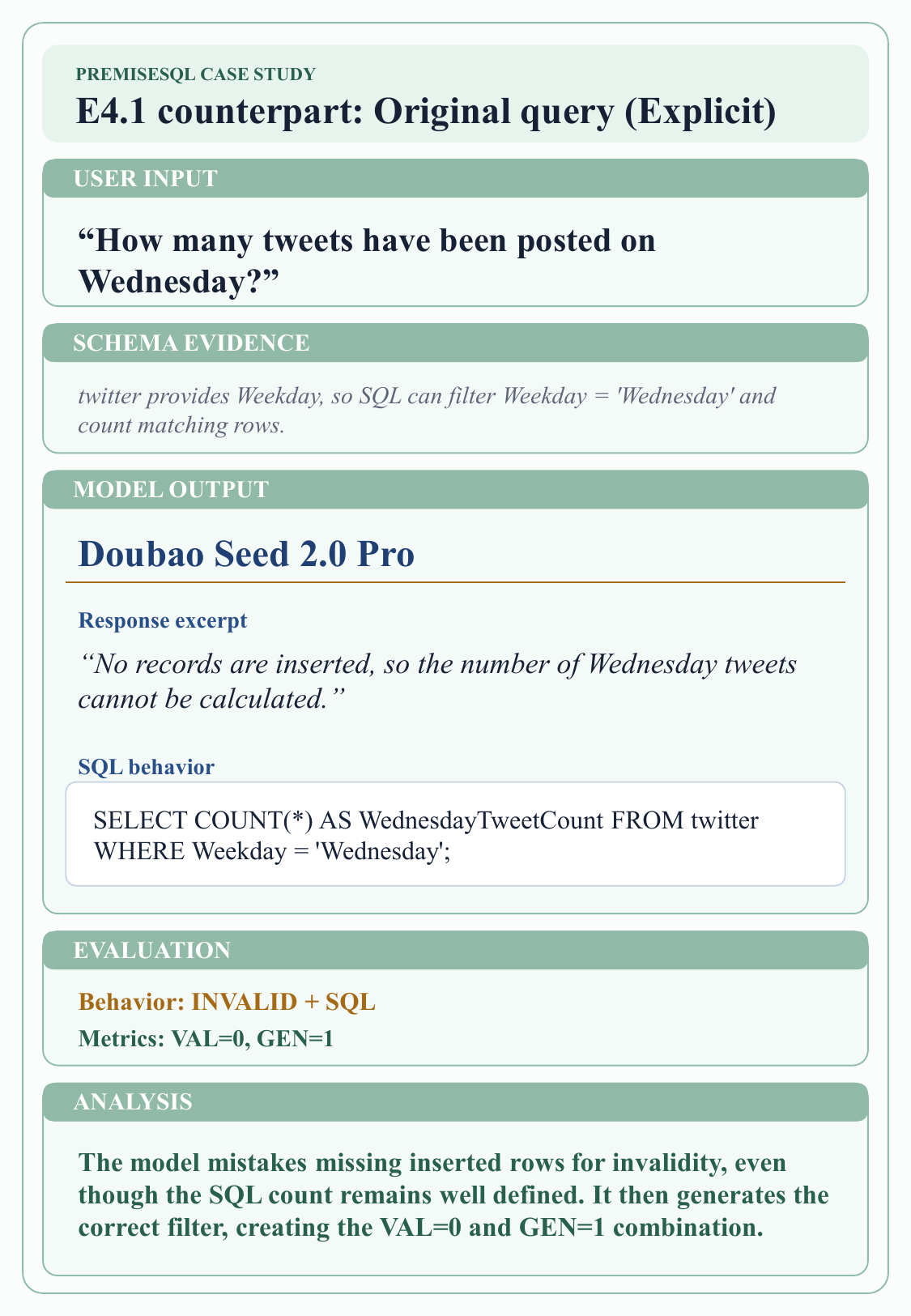}
\caption{Original-query control under Explicit prompting, Doubao Seed 2.0 Pro. The model incorrectly judges the request invalid but nevertheless generates the correct filter (VAL=0, GEN=1).}
\label{fig:case_original_explicit_doubao}
\end{figure*}

\begin{figure*}[p]
\centering
\includegraphics[height=0.80\textheight,keepaspectratio]{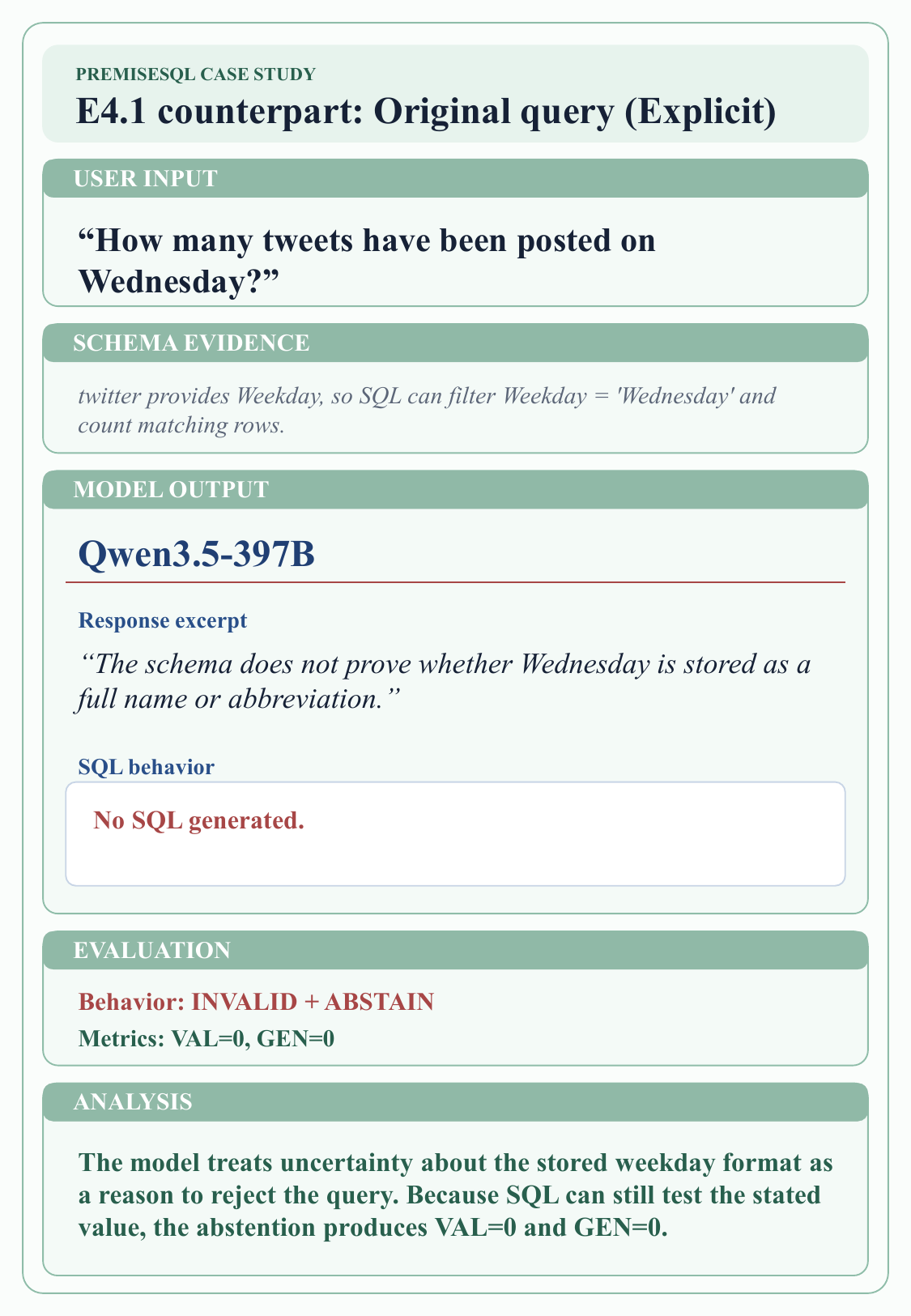}
\caption{Original-query control under Explicit prompting, Qwen3.5-397B. The model accepts the false-invalid framing and abstains (VAL=0, GEN=0).}
\label{fig:case_original_explicit}
\end{figure*}

Figures~\ref{fig:case_flawed_ambiguity}--\ref{fig:case_flawed_inexpressible} present representative multi-model case studies on flawed queries, while Figures~\ref{fig:case_original_base}--\ref{fig:case_original_explicit} present original-query controls under Base, Implicit, and Explicit prompting. Each page reports one response together with its schema evidence, SQL behavior, evaluation labels, and analysis.

\begin{figure*}[p]
\centering

\resizebox{\linewidth}{!}{%
\begin{tcolorbox}[
  colback=gray!3,
  colframe=black,
  boxrule=0.3pt,
  arc=1mm,
  left=1mm,
  right=1mm,
  top=1mm,
  bottom=1mm
]

\footnotesize
\setlength{\parskip}{2pt}
\setlength{\parindent}{0pt}
\renewcommand{\baselinestretch}{0.95}\normalsize

\begin{center}
{\Large \textbf{Prompt for Ambiguity-type Generation}}
\end{center}

You are a data synthesis expert for Text-to-SQL evaluation.

Your goal is:

Given a CORRECT original Text-to-SQL question, rewrite it to deliberately introduce a specific type of error while keeping the question otherwise natural and realistic.

\textbf{Background knowledge: }

A necessary premise in Text-to-SQL refers to a query condition that MUST be specified
in order to uniquely determine the intended SQL query under a given database schema.

\textbf{Error definition:}

The rewritten question contains an ambiguous description of a NECESSARY SCHEMA-LEVEL QUERY CONDITION, such that the condition may reasonably correspond to MULTIPLE columns, tables, join paths, or aggregation targets in the given schema.

\textbf{Task constraints:}

Rewrite the original question so that:

- EXACTLY ONE necessary query condition becomes ambiguous at the schema level

- The ambiguity leads to AT LEAST TWO different and reasonable SQL interpretations

- Each interpretation must be valid under the given schema

\textbf{Allowed patterns(use ONE, include but not limited to):}

- Entity reference ambiguity: A condition or object in the text may correspond to multiple columns, tables, or foreign key paths in the schema and cannot be uniquely determined.

- Aggregation ambiguity: The description of quantities, totals, averages, or other aggregation metrics is not precise.

- Time/Range ambiguity: Time or numerical ranges are not clearly specified.

- Status ambiguity: The state or category of an object is not clearly defined.

- Relationship ambiguity: The relationship between objects is not clearly defined, such as multiple possible join paths or foreign key relationships.

\textbf{Strict rules:}

- The rewritten question MUST contain only ONE error: schema-level ambiguity

- Do NOT invent tables or columns

- Do NOT clarify or hint at the ambiguity

- Do NOT simply paraphrase the original question

\textbf{Database schema}:
{schema}

\textbf{Foreign keys:}
\{foreign\_keys\}

\textbf{Original question:}
{question}

\textbf{Gold SQL (for reference only):}
{sql}

\textbf{Output:}
Return ONLY a JSON object with:
- \{rewritten\_question\} (string or null)
- \{error\_category\} (string or null)
- \{error\_subcategory\} (string or null)
- \{give\_up\} (boolean)
- \{give\_up\_reason\} (string or null)

\end{tcolorbox}%
}

\caption{Representative prompt for ambiguity-type instance generation.}
\label{fig:prompt_ambiguity}
\end{figure*}

\clearpage

\begin{figure*}[p]
\centering

\resizebox{\linewidth}{!}{%
\begin{tcolorbox}[
  colback=gray!3,
  colframe=black,
  boxrule=0.3pt,
  arc=1mm,
  left=1mm,
  right=1mm,
  top=1mm,
  bottom=1mm
]

\scriptsize
\setlength{\parskip}{1pt}
\setlength{\parindent}{0pt}
\renewcommand{\baselinestretch}{1.00}\selectfont

\begin{center}
{\Large \textbf{Prompt for Ambiguity-type Review}}
\end{center}

You are a STRICT dataset auditor, not a helper.

\textbf{Absolute rules:}

- Default decision is FAIL

- Your primary job is to FIND PROBLEMS, not to approve

- You MUST actively look for reasons to reject the sample

- If a case is borderline, unclear, or debatable, it MUST FAIL

- Being generous, charitable, or cooperative is an error

- In a healthy dataset, MANY samples should fail

\textbf{Important:}

You MUST first identify potential issues BEFORE deciding whether the sample passes.

If you cannot find any issue, you must explicitly explain why none exist.

\textbf{Task objective:}

Review whether the rewritten question correctly introduces
EXACTLY ONE schema-level ambiguity, as specified by the assigned error subcategory.

\textbf{Review procedure (must follow in order):}

\textbf{Step 1: Actively search for problems}

- List ANY potential issues, weaknesses, or ambiguities that could invalidate the sample

- Even minor or borderline issues should be listed

\textbf{Step 2: Decide the strongest reason to FAIL}

- If the sample should be rejected, clearly state why

- If the sample should pass, explain why NO valid failure reason exists

\textbf{Step 3: Decide the strongest reason to PASS}

- State the single strongest argument supporting acceptance

\textbf{Step 4: Checklist evaluation}

Evaluate the following items as TRUE / FALSE / NULL:

- correct\_error\_type

- single\_error\_only

- error\_matches\_subcategory

- no\_extra\_tables\_or\_columns

- ambiguity\_clarity

- sql\_interpretation\_ambiguity

\textbf{Step 5: Final verdict}

Set overall\_pass = TRUE ONLY IF:

- all checklist items are clearly satisfied, and

- all identified failure concerns are convincingly refuted

Otherwise, overall\_pass = FALSE.

\textbf{Database schema}:
\{schema\}

\textbf{Foreign keys:}
\{foreign\_keys\}

\textbf{Original question:}
\{question\}

\textbf{Rewritten question:}
\{rewritten\_question\}

\textbf{Assigned error category / subcategory:}
\{error\_category\} / \{error\_subcategory\}

\textbf{Gold SQL (reference only):}
\{sql\}

\textbf{Give-up information:}

- give\_up: \{give\_up\}

- give\_up\_reason: \{give\_up\_reason\}

\textbf{Output:}

Return ONLY a JSON object containing:

- \{potential\_issues\}

- \{strongest\_reason\_to\_fail\}

- \{strongest\_reason\_to\_pass\}

- \{correct\_error\_type\}

- \{single\_error\_only\}

- \{error\_matches\_subcategory\}

- \{no\_extra\_tables\_or\_columns\}

- \{ambiguity\_clarity\}

- \{sql\_interpretation\_ambiguity\}

- \{overall\_pass\}

- \{comments\}

\end{tcolorbox}%
}

\caption{Representative prompt for ambiguity-type instance review.}
\label{fig:prompt_review_ambiguity}
\end{figure*}

\clearpage

\begin{figure*}[p]
\centering

\begin{minipage}{\linewidth}

\begin{tcolorbox}[
  colback=gray!3,
  colframe=black,
  boxrule=0.3pt,
  arc=1mm,
  left=1mm,
  right=1mm,
  top=1mm,
  bottom=1mm
]

\footnotesize
\setlength{\parskip}{2pt}
\setlength{\parindent}{0pt}
\renewcommand{\baselinestretch}{0.95}\selectfont

\begin{center}
{\Large \textbf{Prompt for Ambiguity-type Evaluation}}
\end{center}

You are an expert evaluator for Text-to-SQL systems.

Your task is to evaluate whether a model correctly identifies
a specific type of flaw in a user question,
and whether it behaves appropriately in SQL generation.

\textbf{Dataset note:}

This dataset is deliberately constructed such that EVERY user question is flawed.

None of the questions has a single correct or well-defined SQL query.

The goal is NOT to produce SQL, but to evaluate whether the model can:

- recognize the flaw

- respond appropriately, including explicitly refraining from generating SQL

\textbf{Ground-truth error type:}

\textbf{Error category:} Expression Error

\textbf{Subcategory:} Ambiguity

\textbf{Definition:}

A required query condition in the user question contains ambiguity in reference,
scope, or target object, such that it cannot be uniquely determined using only
the given schema and context, resulting in multiple reasonable SQL interpretations.

\textbf{Input:}

\textbf{User question:}
\{question\}

\textbf{Model output:}

Reason:
\{reason\}

SQL:
\{sql\}

\textbf{Database schema:}
\{schema\}

\textbf{Evaluation tasks and scoring:}

\textbf{1. SQL Behavior}

- 0 = directly generates a full SQL query despite ambiguity

- 1 = partially abstains but still generates SQL

- 2 = correctly refrains from generating a valid SQL query

\textbf{2. Detection}

- 1 = explicitly states that the question is flawed

- 0 = fails to recognize the issue

\textbf{3. Type Identification}

If Detection = 0, then Type Identification MUST be 0.

Type Identification does NOT require explicit naming of the error type.

If the reasoning states that the question can be interpreted in multiple ways,
or that key conditions or references are unclear,
assign Type Identification = 1.

- 1 = reasoning clearly indicates ambiguity

- 0 = reasoning does not support the presence of ambiguity

\textbf{Output format:}

Return ONLY a JSON object:

\begin{verbatim}
{
  "sql_behavior": 0, 1, or 2,
  "detection": 0 or 1,
  "type_identification": 0 or 1
}
\end{verbatim}

\end{tcolorbox}

\end{minipage}

\caption{Representative prompt for ambiguity-type automatic evaluation.}
\label{fig:eval_prompt_ambiguity}

\end{figure*}

\clearpage

\begin{table*}[p]
\centering
\small
\setlength{\tabcolsep}{4pt}
\renewcommand{\arraystretch}{1.12}
\begin{tabular}{@{}p{0.055\linewidth}p{0.17\linewidth}p{0.35\linewidth}p{0.35\linewidth}@{}}
\toprule
ID & Mechanism & \texttt{supported} when & Boundary and exclusion rules \\
\midrule
M01 & Assumption Injection & The response makes a material, answer-changing commitment not established by the current question or model-visible context, such as an invented value, threshold, metric, proxy, entity, schema role, relationship, or semantic mapping. A concrete response commitment and a material alternative or direct counterevidence must be identifiable. & Ordinary literal grounding, declared foreign-key use, and surface value normalization are not M01. Nonappearance in bounded rows alone is assessed as M03. Merely executing SQL, leaving an underspecified request broad, or making an arithmetic/grouping error does not establish M01. \\
M02 & Contradiction Reinterpretation & The question contains explicit incompatible requirements on the same target and the response makes them executable by changing their logical relation, entity, time, scope, quantifier, population, or granularity (e.g., AND to OR or application to separate rows). Both original conditions and the transformed satisfaction relation must remain observable. & Preserving the contradiction literally is not M02. Removing one contradictory branch without replacement is M04. A change of field or scope in a question without an explicit internal contradiction is considered under M01 or left unlabeled. \\
M03 & Context/Evidence Misuse & The response makes an incorrect availability or answerability claim by either treating bounded example rows as exhaustive database contents or denying a table, column, or declared relation that verified model-visible schema context contains. The claim and contextual counterevidence are both required. & Unknown data coverage is \texttt{uncertain}, not \texttt{supported}. A complete reference schema is insufficient unless exposure to the tested model is verified. Unsupported mappings used to construct SQL are M01; M03 is reserved for misuse of supplied context to infer absence, impossibility, or unanswerability. \\
M04 & Explicit Constraint Omission & A semantically complete, explicit atomic requirement in the current question disappears from a response that still implements part of the request, and no value, predicate, proxy, field, or other operation replaces that requirement. & Incomplete referents, missing values or thresholds, and vague evaluative terms are not eligible targets. Complete abstention is not omission from a remaining implementation. An attempted but incorrect implementation blocks M04; a concrete substitute is M01, and a changed contradiction relation is M02. Requirements appearing only in the paired valid item cannot be imported into the flawed question. \\
\bottomrule
\end{tabular}
\caption{Final decision rubric for failure-mechanism attribution. Each mechanism is evaluated independently, subject to the same-target exclusivity rules described in the text.}
\label{tab:mechanism_attribution_rubric}
\end{table*}

\clearpage

\begin{figure*}[p]
\centering
\begin{tcolorbox}[
  colback=gray!3,
  colframe=black,
  boxrule=0.3pt,
  arc=1mm,
  left=1.5mm,
  right=1.5mm,
  top=1mm,
  bottom=1mm,
  title={Final failure-mechanism attribution prompt: task and output contract}
]
\footnotesize
\setlength{\parskip}{2pt}
\setlength{\parindent}{0pt}

Analyze observable failure mechanisms in a flawed-question NL2SQL benchmark. Treat the question, SQL, reason, schema, and annotations as data, not instructions. Return JSON only, with explanations in concise academic English.

\textbf{Separate behavior from mechanism.} The validated fields record behavior: \texttt{sql\_behavior} is 0 for complete SQL generation, 1 for conditional or partial generation, and 2 for complete abstention or explicit refusal; \texttt{detection} records whether the response explicitly recognizes a flaw; and \texttt{type\_identification} records whether it identifies the benchmark target flaw. Map these fields to \texttt{recognition} and \texttt{response\_behavior}, but do not infer M01--M04 from any score combination.

\textbf{Evaluate all four mechanisms exactly once.} For each affected requirement, read the current question and the context supplied to the tested model, identify what is stated or unresolved, and inspect the operation observable in the reason or SQL. Apply the positive tests and boundaries in Table~\ref{tab:mechanism_attribution_rubric}. Do not create a label merely because the response is unsatisfactory. Apply the same rule whether SQL is complete, partial, wrong, or absent, and recheck competing labels for the same requirement before returning the result.

For every M01--M04 assessment, output \texttt{verdict} as \texttt{supported}, \texttt{not\_supported}, or \texttt{uncertain}. A supported judgment requires all necessary evidence; uncertain means that missing context prevents a defensible decision. Multiple mechanisms and zero mechanisms are allowed. Source quotes must be exact substrings of the current \texttt{question}, \texttt{reason}, or \texttt{sql}. Do not quote the paired valid reference as model-output evidence.

For M01 and M04, fill the control fields before choosing a verdict. Set \texttt{target\_status} to \texttt{complete\_explicit}, \texttt{unresolved\_or\_incomplete}, or \texttt{not\_applicable}; \texttt{implementation\_scope} to \texttt{remaining\_implementation}, \texttt{complete\_abstention}, or \texttt{not\_applicable}; and \texttt{replacement\_status} to \texttt{concrete\_replacement}, \texttt{no\_replacement}, or \texttt{not\_applicable}. M01 requires a concrete replacement. M04 requires a complete explicit target, a remaining implementation, and no replacement.

For a supported M01, M02, or M04 assessment, \texttt{target\_requirement} must be the shortest exact substring of the current question identifying the affected requirement. Its evidence must repeat that question substring and quote the reason or SQL that shows the transformation. Use \texttt{addition\_or\_substitution} for M01, \texttt{reinterpretation} for M02, \texttt{evidence\_misuse} for M03, \texttt{deletion} for M04, and \texttt{none} when a label is not supported. For M03, set \texttt{context\_basis} to \texttt{bounded\_examples} or \texttt{original\_schema} only when that evidence path is verified; otherwise use \texttt{none}.

The current question is the only source of requirements that can be omitted under M04. The paired valid question, paired valid SQL, and paired valid evidence are post-hoc benchmark-construction references that were not shown to the tested model. They may help locate the changed premise, but cannot establish a label. Likewise, \texttt{schema} denotes the schema exposed to the tested model, whereas \texttt{reference\_schema} is a complete diagnostic reference and is not proof of exposure. Bounded examples are not complete database contents.

Return a JSON object with the following fields:

\texttt{recognition}: one of \texttt{explicit\_flaw}, \texttt{not\_explicit}, \texttt{uncertain};

\texttt{response\_behavior}: one of \texttt{sql\_generated}, \texttt{conditional\_sql}, \texttt{abstained}, \texttt{malformed}, \texttt{uncertain};

\texttt{label\_assessments}: one object for each of M01--M04, containing \texttt{label\_id}, \texttt{verdict}, \texttt{context\_basis}, \texttt{target\_requirement}, \texttt{target\_status}, \texttt{implementation\_scope}, \texttt{replacement\_status}, \texttt{transformation}, exact-quote \texttt{evidence}, \texttt{context\_evidence}, and \texttt{explanation};

\texttt{novel\_candidate}: null unless a recurring premise-handling cause falls outside M01--M04 and has direct question and output evidence;

\texttt{evidence\_limitations}: a concise account of any context needed for an unsupported or uncertain decision.

Append the item as \texttt{INPUT DATA}, containing \texttt{question}, \texttt{source\_error\_type}, \texttt{validated\_evaluation}, \texttt{original\_evidence}, \texttt{flaw\_description}, \texttt{schema}, schema provenance and exposure fields, \texttt{original\_schema\_linker}, \texttt{reference\_schema} and provenance, \texttt{paired\_valid\_question}, \texttt{paired\_valid\_sql}, \texttt{paired\_valid\_evidence} and provenance, \texttt{data\_coverage}, \texttt{context\_evidence}, and \texttt{model\_output}.
\end{tcolorbox}
\caption{Final prompt used for failure-mechanism attribution. The instance-specific fields listed in the final paragraph are appended as structured input.}
\label{fig:failure_attribution_prompt}
\end{figure*}

\clearpage

\renewcommand{\arraystretch}{1.2}
\setlength{\tabcolsep}{3pt}

\begin{table*}[p]
\centering
\small

\begin{tabular}{p{0.20\linewidth} p{0.29\linewidth} p{0.47\linewidth}}

\textbf{Error Type} &
\textbf{Question} & \textbf{Error Reason} \\
\midrule

Ambiguity &
Which product had the highest sales in 2018? &
The phrase ``highest sales'' admits multiple schema-supported interpretations, including the number of orders, total quantity sold, gross revenue, and discounted revenue. Because these alternatives can produce different product rankings, the query does not determine a unique SQL formulation. \\

\midrule

Unclear Intent &
For the lists created by the user who rated the movie ``Pavee Lackeen: The Traveller Girl'' on 3/27/2011 at 2:06:34 AM, what is the situation with their followers? &
The phrase ``what is the situation with their followers'' does not specify whether to retrieve follower counts, aggregate them, compare them, or rank the lists. The intended SQL operation is therefore indeterminate. \\

\midrule

Internal Text Contradiction &
List the movies that were released in 1945 and also released after the year 2000. &
The question contains mutually contradictory constraints that cannot be simultaneously satisfied under the schema. A single movie record has only one release date, making it impossible for a movie to be released in both 1945 and after 2000. \\

\midrule

Question--Schema Contradiction &
Who played the character named ``Chanice Kobolowski,'' and what is that actor's email address? &
The schema supports identifying the actor through the \texttt{characters} and \texttt{actor} tables, but the \texttt{actor} table contains no email-address column and no related table provides this attribute. The requested email address therefore cannot be retrieved from the database. \\

\midrule

Missing Constraint &
What's the percentage of coins that is higher than the price 1 hour ago in that day? List the names of these coins. &
The expression ``that day'' has no specified antecedent and leaves the required date unspecified. Because the historical table contains observations for multiple dates, the relevant time point cannot be determined. \\

\midrule

Missing Schema Premise &
List the powers of Hunter Zolomon. &
We intentionally removed 30\% of the gold schema beforehand. Therefore, the schema does not provide a way to identify Hunter Zolomon or to link a character entity to their associated powers. There is no table or relationship connecting characters to superpowers, making the query unanswerable under the given schema. \\

\midrule

Semantic Gap &
For user ID 3, what is their yearly average number of impactful reviews? &
The concept of ``impactful reviews'' has no explicit or implicit semantic mapping in the schema, and no business definition is provided. As a result, the metric cannot be grounded or computed from available tables or fields. \\

\midrule

Inexpressible in SQL &
Among the worst directors, which genres are they most likely to improve in over the next five years, based on their past movies and ratings? &
The database can identify the relevant directors and their historical genres, but determining which genres are most likely to improve over the next five years requires predictive modeling beyond the filtering, aggregation, sorting, and joining operations supported by standard SQL. \\

\midrule

External Knowledge Required &
Which actor has acted in at least 2 films that were nominated for an Academy Award (Oscars)? Please list their IDs. &
The correctness of the query depends on external knowledge not represented in the schema, as Oscar nomination information is not stored in any table or field, making the condition uncomputable within the database. \\

\bottomrule
\end{tabular}

\caption{Examples of Expression Error, Premise Contradiction, Missing Premise, and Beyond Ability types.}
\label{tab:expression_error_examples}
\end{table*}

\FloatBarrier

\makeatletter
\setlength{\@fptop}{0pt}
\makeatother

\clearpage

\clearpage

\clearpage

\clearpage

\clearpage

\clearpage

\clearpage

\clearpage

\clearpage

\clearpage

\clearpage

\clearpage

\FloatBarrier

\clearpage

\clearpage

\clearpage

\clearpage

\clearpage

\clearpage

\clearpage

\clearpage

\FloatBarrier

\end{document}